\documentclass[10pt,twocolumn,letterpaper]{article}

\usepackage[pagenumbers]{cvpr} 

\usepackage[accsupp]{axessibility}  
\usepackage{xspace, microtype}

\usepackage{amsfonts, mathrsfs}
\usepackage{dsfont}
\usepackage{bbm}
\usepackage{stmaryrd}
\usepackage{dirtytalk}
\usepackage{amsmath}
\usepackage{amssymb}
\usepackage{mathtools}
\usepackage{pifont}
\usepackage{ragged2e}

\usepackage{tabularx}       
\usepackage{tablefootnote}   
\usepackage{tabulary}
\usepackage{booktabs}       
\usepackage{multirow}       
\usepackage{makecell}       

\usepackage{graphicx}
\usepackage{float}
\usepackage{subcaption}
\usepackage[export]{adjustbox}
\usepackage{wrapfig}
\usepackage{tcolorbox}
\usepackage[linesnumbered,ruled,vlined]{algorithm2e}

\usepackage{tikz}
\usetikzlibrary{quotes, arrows.meta, angles, calc, 3d, shapes, intersections, plotmarks}
\usepackage{pgfplots}
\pgfplotsset{compat=1.18}  

\usepgfplotslibrary{statistics, polar, groupplots}
\pgfplotsset{every mark/.append style={solid}}
\usepackage{standalone}
\usepackage{tikzscale}
\usepackage{tikz-3dplot}

\usepackage{enumitem}

\usepackage{url}

\usepackage{array}
\usepackage{etoolbox}
\usepackage{amsthm}
\usepackage{fontawesome5}

\usepackage{placeins}

\newcolumntype{C}{>{\centering\arraybackslash}X}

\newcommand{\task}{unobserved object detection\xspace}
\newcommand{\Task}{Unobserved object detection\xspace}
\newcommand{\TASK}{Unobserved Object Detection\xspace}

\newcommand{\cells}[2]{\setlength{\baselineskip}{0.8\baselineskip}#1{\quad\quad\scriptsize$\pm$#2}}
\newcommand{\best}[2]{\setlength{\baselineskip}{0.8\baselineskip}\textbf{#1}{\quad\quad\scriptsize$\pm$#2}}

\makeatletter
\renewcommand\paragraph{
  \@startsection{paragraph}
                {4}
                {\z@}
                {2ex \@plus1ex \@minus.2ex}
                {-1.5em}
                {\normalfont\normalsize\bfseries}}
\makeatother
\definecolor{cvprblue}{rgb}{0.21,0.49,0.74}
\usepackage[pagebackref,breaklinks,colorlinks,allcolors=cvprblue]{hyperref}

\def\paperID{8100}
\def\confName{CVPR}
\def\confYear{2025}

\title{Believing is Seeing: Unobserved Object Detection using Generative Models}

\author{
    Subhransu S. Bhattacharjee,
    Dylan Campbell, 
    Rahul Shome\\[0.5em]
    School of Computing, The Australian National University \\[0.5em]
    \tt\small \{subhransu.bhattacharjee, dylan.campbell, rahul.shome\}@anu.edu.au
}

\begin{document}
\maketitle

\newcommand{\obj}{o}
\newcommand{\objects}{\mathcal{O}}
\newcommand{\scene}{S}
\newcommand{\scenes}{\mathcal{S}}
\newcommand{\rand}[1]{\mathbf{#1}}
\newcommand{\samples}{k}
\newcommand{\image}{\mathcal{I}}
\newcommand{\frust}{{F}}
\newcommand{\scenevolume}{{V}}
\newcommand{\volume}{\mathcal{V}}
\newcommand{\genai}{\mathcal{M}}
\newcommand{\objsem}{object\xspace}
\newcommand{\ssh}{\mathcal{D}}
\newcommand{\pr}{\mathbb{P}}
\newcommand{\event}{\mathcal{E}}
\newcommand{\bb}{\mathbb{B}_o}
\newcommand{\ssd}{\mathcal{D}_{\image}}
\newcommand{\sd}{\mathcal{D}_{\image\obj}}
\newcommand{\sh}{\tilde{\mathcal{D}}_{\image\obj}}
\newcommand{\inim}{\mathcal{I}}
\newcommand{\gtd}{\Gamma_{\image}}
\newcommand{\gtdo}{\Gamma_{\image\obj}}
\newcommand{\conf}{\omega_{\obj}}
\newcommand{\outim}{\mathbb{I}}
\newcommand{\outvol}{\mathbb{V}}
\newcommand{\e}{\mathbf{x}}
\newcommand{\gspan}{\mathcal{X}}
\newcommand{\ent}{\mathcal{H}}
\newcommand{\objcond}[1]{#1_\obj}
\newcommand{\cent}{\mathcal{H}^{\times}}
\newcommand{\gtsd}{\mathcal{G}_{\inim\obj}}
\newcommand{\ggrid}{\tilde{\gtsd}}
\newcommand{\dnn}{\Delta}
\newcommand{\rwa}{\mathcal{A}}
\newcommand{\ltask}{\task}
\newcommand{\expect}{\mathbb{E}}
\newcommand{\falseneg}{\text{FNR}}
\newcommand{\indi}{\mathds{1}}
\newcommand{\nyu}{NYU Depth V2\xspace}
\newcommand{\ct}{\vspace{-1em}}
\newcommand{\rk}{RealEstate10k\xspace}
\newcommand{\gtlabel}{y_{\inim\obj}}
\newcommand{\plabel}{\hat{y}_{\inim\obj}}
\newcommand{\diam}{\mathtt{diam}}
\begin{abstract}
Can objects that are not visible in an image---but are in the vicinity of the camera---be detected?
This study introduces the novel tasks of 2D, 2.5D and 3D unobserved object detection for predicting the location of nearby objects that are occluded or lie outside the image frame. 
We adapt several state-of-the-art pre-trained generative models to address this task, including 2D and 3D diffusion models and vision--language models, and show that they can be used to infer the presence of objects that are not directly observed. 
To benchmark this task, we propose a suite of metrics that capture different aspects of performance. 
Our empirical evaluation on indoor scenes from the \rk and \nyu datasets demonstrate results that motivate the use of generative models for the unobserved object detection task.
\end{abstract}
\vspace{-1em}
\section{Introduction}
\label{section:introduction}

\begin{figure}[htbp]
    \centering
    \includegraphics[width=0.95\columnwidth]{./images/splash.tex}
    \caption{\textit{\Task} aims to infer the location of objects that were not directly observed in an image. Consider this toy example of a dining table and chairs. Here we visualize a top-down view (slice) of the predicted discrete distributions $\sd^{\text{2D}}$ and $\sd^{\text{3D}}$ for the object label $\obj$ of ``chair,'' conditioned on the image $\inim$, where darker is more probable. The presence of an occluded chair (A) is predicted as relatively likely, as is the presence of an out-of-frame chair (B). Crucially, the domain $\outvol$ of the predicted 3D distribution exceeds the camera frustum (drawn in black), and the domain $\outim$ of the 2D distribution extends beyond the image plane $\inim$.}
    \label{fig:splash}
\end{figure}

Object detection~\cite{zou2023objectdetection20years} is a fundamental building block of autonomous systems that are capable of making sense of their surroundings. 
However, it is limited to visible surfaces within the camera’s field-of-view. 
While amodal object detection and segmentation~\cite{zhu2017semantic,ling2020variational} relaxes this requirement to permit partially visible objects, we propose to go further and detect those objects that are near the camera but were not observed at all.

In this paper, we propose the novel task of \task in 2D, 2.5D and 3D for inferring the presence of objects not seen in an image within a domain that extends beyond the camera frustum and behind the visible surfaces. 
We are particularly motivated by perceptual and planning tasks 
such as visual search
~\cite{zhu2017target, chad},
where a robot might be expected to use its understanding of a scene \textit{beyond} what it can directly observe.
For example, consider the setup in \cref{fig:splash}.
A robot tasked with locating someone who is seated might find it useful to look to the right, where it is likely that additional chairs would be found, or to move to view a potentially occluded chair.
Predicted spatial likelihoods of objects in a scene that was only partially observed can inform probabilistic models~\cite{kurniawati2022partially, 10496149, unaware} for planning under uncertainty. 
In this work we recover such spatio-semantic distributions over observed and unobserved parts of the scene.
We expect progress on this task to benefit downstream applications in which decisions are made in the presence of uncertainty arising from partial observations, such as active vision, navigation, visual search, adaptive planning and scene exploration~\citep{nerf-nav, nav_real, vsearch, shah2023vint, chad, jutrasdube2024adaptive, explore}.

While explicitly learning spatio-semantic distributions of scenes and objects is challenging~\citep{wen2023can}, they can be variationally inferred using generative models trained on 2D image datasets~\citep{kingma2021variational, render}. 
Novel view synthesis and semantic scene completion~\citep{Eslami2018, watson2022novel, fwd, GeNVS, bautista2022gaudi, zeronvs, lotr, li2022infinitenature} 
can offer useful spatio-semantic priors that can be used to render full 3D volumes given partial observations.
However, occlusions and incomplete or sparse data still pose significant challenges to 3D scene generation methods. 
Image diffusion models~\citep{ho2020denoising, song2021scorebased} have been shown to be useful at expanding beyond the visible frame, known as outpainting~\citep{rombach2022high, podell2024sdxl}, and some 3D diffusion models~\citep{fwd} can generate 3D representations consistent with an input image. 
These approaches typically aim for photorealism, of the completed scene, which may not be necessary or desirable for \task, where we want to infer the spatio-semantic distribution of an object situated within possible scenes.
Despite this weakness, the generative capability of such models is still highly useful. 

In this work, we investigate the ability of three categories of generative models (2D and 3D diffusion models and vision--language models) to predict spatio-semantic distributions via a sampling procedure.
We develop the associated detection pipelines, metrics, and evaluation protocol for the \task task.
Our contributions are
\begin{enumerate}[noitemsep] 
\item defining the novel task of \task;
\item extending three major classes of generative models (2D and 3D diffusion models and vision--language models) to address \task; and 
\item benchmarking these approaches under a standardized evaluation protocol with respect to the accuracy and diversity of the predictions.
\end{enumerate}
We evaluate performance under 2D, 2.5D, and 3D settings, on the \rk~\citep{stereomag} and \nyu~\citep{NYU} datasets using MS-COCO object categories~\citep{lin2014microsoft}.
The results from these pre-trained models is promising, but also reveals several challenges, motivating further research on the \task task. 
\section{Related Work}
\label{sec:related-work}

\textit{Object detection}~\cite{zou2023objectdetection20years} focuses on identifying and localizing objects within images, with recent deep learning models such as Faster R-CNN \citep{ren2016faster}, YOLO \citep{redmon2016yolo}, and transformer-based approaches like DETR~\citep{detr} significantly enhancing both accuracy and efficiency.
However, occlusions in real-world scenarios, such as autonomous driving and robotics, remains a critical challenge for traditional models.
Amodal object detection and segmentation \citep{zhu2017semantic,ling2020variational} extends traditional approaches by predicting the full projected geometry of partially occluded objects. 
\citet{li2016} introduced amodal bounding box regression including unseen parts.
Recent improvements~\citep{bayesian, back2022unseen} leverage Bayesian or hierarchical occlusion modeling. 
However, these methods cannot reason about fully occluded or out-of-frame objects.

\textit{2D diffusion models} \citep{sohl2015deep, ho2020denoising, song2021scorebased} generate high-quality, plausible images via an iterative denoising process, which can be guided by conditioning inputs such as text or images~\citep{kim2022diffusionclip, saharia2022palette}.
Classifer-free guidance~\citep{ho2022classifier}, capable of learning the conditional and unconditional models simultaneously, has been applied to scene completion and conditional synthesis~\citep{scene_completion, saharia2022photorealistic}.
Stable Diffusion \citep{rombach2022high, podell2024sdxl} is used for extending high-resolution image boundaries, conditioned on text, in image outpainting, generating high-resolution images, conditioned on text~\citep{ho2022classifier}.  
We explore spatio-semantic relationships learned by such models in the context of our task.

\textit{Novel View Synthesis (NVS)}~\citep{snavely2006photo, hartley2003multiple} is the task of generating new views of a scene from a set of posed input images.
Neural Radiance Fields (NeRF) \citep{mildenhall2021nerf} parameterize scenes as continuous volumetric radiance fields, enabling high-fidelity rendering.
However, NeRF typically requires many input images and known camera poses.
Several related methods~\citep{PixelNeRF, rockwell2021pixelsynth, xu2022sinnerf} leverage pre-trained scene priors to improve generalization in the sparse input setting, but the reconstruction is typically blurry in the unobserved regions.
More recent approaches~\citep{render, zeronvs, photonvs} integrate a diffusion model into the NVS pipeline to overcome these averaging issues.
Diffusion with Forward Models (DFM) \citep{fwd} combines a transformer encoder with PixelNeRF \citep{PixelNeRF, DIT} and uses a Denoising Diffusion Implicit Model (DDIM) \citep{DDIM} to generate diverse 3D representations conditioned on a single RGB image.
GeNVS \citep{GeNVS} follows a similar approach to sample novel views instead of full 3D scenes.
Newer models like MultiDiff~\citep{multidiff},
using depth-warped priors, show improvements in quality and speed, but code has not yet been publicly released.
Despite its computational demands and challenges with out-of-distribution data, DFM remains a strong benchmark for generating 3D-consistent views of a scene from a single image.
We investigate whether DFM-generated scenes capture plausible spatio-semantic relationships for our task.

\textit{Vision--Language Models (VLMs)}~\citep{VLM_survey} are effective for spatio-semantic reasoning, excelling at one-shot and zero-shot tasks, but struggle with context-based reasoning.
Models like CLIP~\citep{CLIP} or ALIGN~\citep{align} which align visual and textual data for object recognition and segmentation, encounter difficulties in contexts with incomplete or sparse information~\citep{spatialvlm}.
These limitations are evident in tasks requiring spatial and compositional reasoning~\citep{3dvla} and benchmarks~\citep{winoground, mllms} which highlight difficulties in ambiguous scenes.
While models like ChatGPT and LLaVA~\citep{chatgpt4, llava-1, llava-2, spatialvlm} have improved multi-modal reasoning, they have limitations in understanding 3D spatial relationships~\citep{bordes2023pug}.
\citet{cirik-etal-2022-holm} leveraged contextual semantic co-occurrences, learned by language models, from the unobserved parts of a scene and relied on semantic affinities for active prediction.
In contrast, our work evaluates whether modern VLMs~\citep{chatgpt4,llava-2,gemini,claude} can perform this passive inference, implicitly from a single observation.

\section{\TASK}
\label{sec:method}

The task of \task, as shown in \cref{fig:splash}, is to detect objects that are in the scene, but are not captured within the camera frustum. 
The approaches considered in this paper address this task by predicting a conditional distribution $\ssd$ over a bounded space and set of semantic labels given a single RGB image $\inim$, referred to as a spatio-semantic distribution (SSD).
This facilitates the inference of the probable locations of unobserved objects outside the visible region captured by the image and its associated camera frustum, using contextual cues from the image.
The spatial domain $\gspan$ defines the support of this distribution.
The SSD is an estimate of the ground-truth conditional distribution $\gtd$.
While this distribution is not observable, %
we have access to additional posed images from a realization of this conditional distribution for the purpose of evaluation. 
For each object label $\obj$, we denote the associated SSD as $\sd$.
This maps each $x \in \gspan$ to a likelihood
in $[0, 1]$.
That is, the likelihood of detecting object $\obj$ at location $x$.
For example, it assigns a likelihood to the presence of a chair one meter in front of and two meters to the right of the camera, which may not be visible in the input image.
Note that we consider all distributions to be normalized.
We consider distributions over two discretized spatial domains $\gspan$:
(1) a 2D domain 
$\outim$, 
discretized as a pixel grid, corresponding to an expansion of the input image; and
(2) a 3D domain 
$\mathbb{V}$, 
discretized as a voxel grid, extending beyond the input camera's frustum.
The 2D domain 
$\outim$ 
symmetrically extends the $H \times W$ pixel grid of the input image to $H \times W'$, where $W' > W$.
The 3D domain 
$\mathbb{V}$ 
expands the frustum to an expanded 
voxel grid aligned with the input camera pose.

\section{Generative Model-based Detection Pipelines}
\label{sec:approach}

\begin{figure}[!t]
    \vspace{-0.15in}
    \centering
    \includegraphics[width=\linewidth]{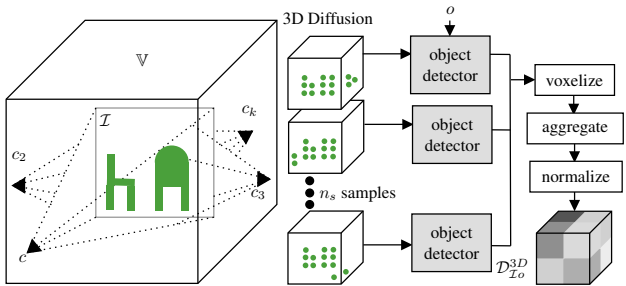}
    \vspace{-0.25in}
    \caption{The 3D diffusion-based pipeline.}
    \label{fig:3D_pipeline}
    \vspace{-0.15in}
\end{figure}

\begin{figure}[!t]
    \vspace{-0.15in}
    \centering
    \includegraphics[width=\linewidth]{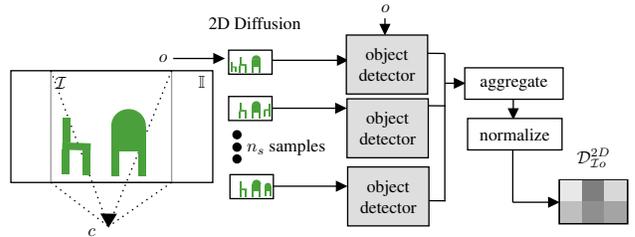}
    \vspace{-0.3in}
    \caption{The 2D diffusion-based pipeline.}
    \label{fig:SDXL_pipeline}
    \vspace{-0.15in}
\end{figure}

\begin{figure}[!b]
    \centering
    \includegraphics[width=0.8\columnwidth]{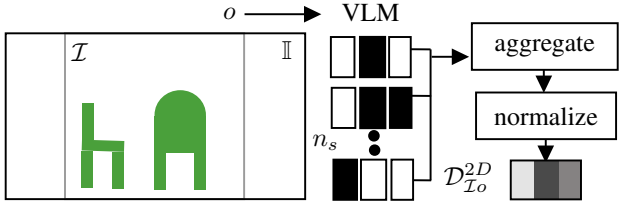}
    \caption{The VLM-based pipeline.}
    \label{fig:VLM}
\end{figure}

This section details how three pipelines based on generative models---(1) a 3D diffusion model (\cref{fig:3D_pipeline}), a 2D diffusion model (\cref{fig:SDXL_pipeline}), and (3) vision--language models (\cref{fig:VLM}), which can be used to estimate the 2D or 3D SSD $\sd$ from an RGB image $\inim$, for object $\obj$.
\paragraph{3D Diffusion-based Model.} 
We utilize Diffusion with Forward Models (DFM)~\citep{fwd} to estimate both the 2D and 3D $\sd$.
This approach generates a 3D representation conditioned on a single RGB image, from which an RGBD image can be rendered from any camera pose.
We select a set of $k = 4$ target camera poses, including the input image's camera pose, as visualized in \cref{fig:3D_pipeline}, render the corresponding RGBD images, and combine the results into a single point cloud. An off-the-shelf object detector, YOLOv8x~\citep{redmon2016yolo, ultralytics2023yolov8}, is run on these frames.
For 3D detection, the 2D bounding boxes for each object $\obj$ are back-projected onto the point cloud alongside detection confidences.
We take multiple samples from the conditional distribution by repeating the generation process a total of $n_s$ times over additional camera poses, conditioned on the same input image $\inim$, and the results are aggregated into the estimated distribution. 
We combine all the 3D samples into a common frame of reference, 
voxelize the resulting 3D confidence map into grid cells $\outvol$, average the non-zero confidences of the points across each voxel, and then apply softmax normalization to obtain the 3D distribution $\sd^{3D}$.
The DFM pipeline can expand the detection volume beyond the initial visible frame and recover occlusions in 3D using the novel camera views.
More details on the sampling process is in \cref{sec_additional_implementation_details}.
\paragraph{2D Diffusion-based Model.} 
To evaluate the capacity of a 2D diffusion model at estimating the conditional distribution, we utilize  
a 2D diffusion model,
SDXL \citep{podell2024sdxl, diffusers}, 
which expands the image canvas of $\inim$ using outpainting to perform conditional generation to fill the unobserved regions to the left and right of the input image 
($360 \times 360$ input to a $360 \times 640$ output).
The process is repeated for $n_s$ generated samples (shown in \cref{fig:SDXL_pipeline})
We then run an object detector, YOLOv8x \citep{redmon2016yolo, ultralytics2023yolov8} to assign confidence scores for object $\obj$ to each pixel, followed by softmax normalization, resulting in a 2D SSD $\sd^\text{2D}$.
To obtain the 3D distribution $\ssd^\text{3D}$, we re-projected 2D bounding boxes into 3D using metric depth values from DepthAnythingv2~\citep{depthanythingv2, depthanything} and camera parameters from the \rk dataset, with additional parameters from COLMAP reconstructions~\citep{SFM, colmap} for the \nyu dataset, then voxelized the 3D point map, averaging confidence scores within each voxel.

\paragraph{VLM-based Models.} 
We adapt several VLMs to the task of predicting the (coarsely-discretized) 2D conditional distribution: 
ChatGPT-4o \citep{chatgpt4}, Claude-3.5 Sonnet \citep{claude}, Gemini 1.5 Ultra \citep{gemini} and LLaVa-34B-v1.6 \citep{llava-1, llava-2}.
To do so, each model was provided with the $360 \times 360$ input image, padded with $140$ black pixels to the left and right, forming a $640 \times 360$ image.
For each object $\obj$, the model was queried for a constrained binary \say{Yes}/\say{No} response whether the object was within the original frame or the extended region to the left or to the right.
A fraction of \say{Yes} responses out of $n_s$ queries was produces a prediction confidence for the 
left/center/right 
regions, further normalized to generate 
$\sd^\text{2D}$ (\cref{fig:VLM}) 
for
region-wise estimates.
Prompt and experiment design details are provided in \cref{subsec:exp_setup} and \cref{sec:appendix_vlm}.

\section{Experiments}
\label{sec:exp}

We evaluate model performance on three distinct tasks---2D, 2.5D, and 3D---for
indoor scenes~\citep{stereomag,NYU} with MS-COCO objects~\citep{lin2014microsoft}.
Each experiment generates SSDs, aligned with the input image’s camera pose.
The 2D task involves estimating $\sd^\text{2D}$ within the expanded image domain $\outim$ from the viewpoint of the input camera.
The 2.5D task focuses on estimating the portion of $\sd^\text{3D}$ visible in the expanded image $\outim$, centered at the input camera location.
Finally, the 3D task extends this estimation for $\sd^\text{3D}$ over an expanded 3D region that accounts for out-of-frame detections and occlusions.

\subsection{Metrics for Unobserved Object Detection}
\label{sec:metrics}

To evaluate the \task task, we propose metrics to quantify how well the predicted conditional distribution $\sd$ for an object $\obj$ over a discrete domain $\gspan$ aligns with the true conditional distribution. However, we only have access to a single realization of this distribution, denoted by $\gtsd$. 
Significantly, the true conditional distribution $\gtdo$ is likely to have more modes than the realization $\gtsd$.
For example, given an image $\inim$ of a dining table, the ground-truth scene (the realization) happened to have several chairs clustered to the right of the camera.
However, the exact same image could equally well have corresponded to a scene with a chair to the left of the camera (another, unrealized, scene conditioned on the same image).
We expect the models to predict all plausible modes, and our metrics are informed by this asymmetry between the ground-truth and the prediction.
\paragraph{Normalized Entropy ($\ent$) and Cross-Entropy ($\cent$).}
The normalized entropy $\ent$~\citep{shannon1948} of the predicted distribution $\sd$ and the normalized cross-entropy $\cent$~\citep{cover2006} between the ground-truth distribution $\gtsd$ and $\sd$ are given by
\begin{align}
    \ent_{\inim\obj} &\coloneqq \frac{-1}{\log|\gspan|}  \sum_{x \in \gspan} \sd(x) \log \sd(x), \label{subeq:ent} \\
    \cent_{\inim\obj} &\coloneqq \frac{-1}{\log|\gspan|}  \sum_{x \in \gspan} \gtsd(x) \log \sd(x), \label{subeq:cent}
\end{align}
for input image $\inim$ and object $\obj$.
The normalization factor $\log|\gspan|$ makes the metric independent of the grid resolution and ensures that a uniform distribution evaluates to unit entropy.
This asymmetric measure penalizes the prediction of low object likelihood where the ground-truth indicates the presence of an object.
However, as required, it does not penalize the prediction of high object likelihoods where the ground-truth has not seen an object.
For the remaining metrics, we consider high-likelihood detections above a threshold $\tau$. 
The thresholded ground truth is denoted by 
$\gtsd^\tau$ for all $x\in\gspan$ where $\gtsd(x)>\tau$.
Let the detection label $\gtlabel$ be $1$ if $\gtsd^\tau$ is non-empty and $0$ otherwise.
The thresholded prediction $\sd^\tau$ and label $\plabel$ are 
defined over $\sd$.
\paragraph{Normalized Nearest Neighbor Distance ($\dnn$).}
This asymmetric metric quantifies the spatial distance between the peaks (local maxima) in the thresholded ground truth $\mathcal{P}_{\gtsd^\tau}$ to the nearest peaks in the thresholded predictions $\mathcal{P}_{\sd^\tau}$.
For a single input image $\inim$ and object $\obj$, this is given by
\begin{equation}
 \dnn_{\inim\obj} \coloneqq \frac{1}{|\mathcal{P}_{\gtsd^\tau}|} \sum_{x \in \mathcal{P}_{\gtsd^\tau}} \underset{x' \in \mathcal{P}_{\sd^\tau}}{\text{min}} \frac{\| x - x' \|}{\diam(\gspan)},
\label{eq:spatacc}
\end{equation}
where $\diam(\gspan)$ is the maximum distance within $\gspan$ (i.e., the
diagonal in the 2D or 3D grid)  that normalizes the measure with respect to the domain size.
If predicted peaks $\mathcal{P}_{\sd^\tau}$ is empty, $\dnn_{\inim\obj} \coloneqq \infty$.
Large distances indicate that the object was not predicted near where it is located in the ground truth.
\paragraph{False Negative Rate ($\falseneg$).}
Measured over the entire set of evaluations (scenes $\scenes$ and objects $\objects$),
the thresholded ground-truth and predicted labels are used in this metric to compute the false negative rate of classification, given as.
\begin{equation}
    \falseneg  \coloneqq   \frac{\sum_{\inim \in \scenes} \sum_{\obj \in \objects} \gtlabel (1-\plabel)}{\sum_{\inim \in \scenes} \sum_{\obj \in \objects} \gtlabel}.
\end{equation}
\paragraph{2D Region-wise Accuracy ($\rwa$).} 
This metric evaluates the coarse-grained classification accuracy of the predictor in 2D.
Specifically, we decompose the domain $\gspan$ into $\ell$ subdomains $\gspan^i$ and define corresponding ground-truth and prediction labels for thresholded detections over the regions as $\gtlabel^i$ and $\plabel^i$ respectively. 
For image $\inim$ and object $\obj$, this is given by
\begin{equation}
    \rwa_{\inim\obj} \coloneqq \frac{1}{\ell} 
    \sum_{i=1}^{\ell}
    \gtlabel^i \plabel^i + (1 - \gtlabel^i) (1 - \plabel^i).
    \label{eq:rwa}
\end{equation}

\subsection{Experiment Setup}
\label{subsec:exp_setup}
In this section we provide details pertaining to the datasets, baselines, implementation, and evaluation protocol.
\paragraph{Datasets.}
All models were evaluated on 10 indoor scenes from the \rk test set~\citep{stereomag} and the \nyu dataset~\citep{NYU}, automatically selected based on object detection frequency.
After filtering, each scene has an average of 190 images in \rk ($720 \times 1280$, resized to $360 \times 640$ for $\outim$, and center-cropped to $360 \times 360$ for $\inim$) and 1359 images in \nyu ($480 \times 640$, center-cropped to $360 \times 640$ for $\outim$ and center-cropped to $360 \times 360$ for $\inim$).
One randomly selected frame per scene was used as the input image $\inim$ for evaluation.
COLMAP~\citep{colmap} and GLOMAP~\citep{glomap} were used for camera pose estimation (for \nyu) and for reconstruction to obtain depth maps.
Object annotations for each expanded $360 \times 640$ image $\outim$ were generated using YOLOv8x~\citep{redmon2016yolo, ultralytics2023yolov8}, pre-trained on the MS-COCO dataset~\citep{lin2014microsoft}, retaining those with confidence greater than $\tau_\text{conf}$.
The 2D ground-truth spatial distributions $\gtsd^\text{2D}$ were computed by a softmax normalization of the detected object confidence  logits.
The 3D ground-truth spatial distributions $\gtsd^\text{3D}$ were computed by back-projecting the object confidence logits using the 3D reconstructions, voxelizing with average pooling, then applying softmax normalization.
The voxel grid has resolution $20 \times 20 \times 20$ with grid size $1.0$.
The 2.5D ground truth spatial distributions $\gtsd^\text{2.5D}$ were computed similarly, by back-projecting the confidence logits within the input camera frustum and using a voxel grid of resolution $10\times10\times10$, with a grid size of $1.0$.
Note that we only retain non-empty ground-truth distributions in the dataset: if a scene $s$ does not contain an object $\obj$, $(s, o)$ is not part of the dataset.
See \cref{sec:appendix_scene,sec:appendix_reconstruction} for details on automatic scene pre-processing, filtering, selection, and reconstruction.
\paragraph{Baselines.} 
To provide context for the benchmark, we report results for a \textit{uniform} baseline, whereby equal probabilities are assigned across the grid.
We also report results for an \textit{oracle}, which has access to the ground truth.
For the 2D task, the oracle uses the 2D ground-truth distribution corresponding to the expanded image $\outim$.
For the 2.5D task, the oracle uses the unprojected 2D ground-truth distribution.
For the 3D task, the oracle uses the 3D ground-truth distribution.
\paragraph{Implementation Details.}
The hyperparameters are set as follows:
the classification threshold $\tau  = 1.4 / |\gspan|$,
the minimum detection confidence $\tau_\text{conf} = 0.1$,
and the number of subdomains $\gspan^i$ for region-wise accuracy $\ell = 3$, corresponding to the $360 \times 360$ input image $\inim$, the $360 \times 140$ domain to the left of the image and the $360 \times 140$ domain to the right of the image.
For DFM~\citep{fwd} inference, we use autoregressive sampling with 3 timesteps per frame, generating 50 intermediate frames with default settings: a temperature of $0.85$ and a guidance scale of $2.0$. Generating each sample took $25$ minutes per target pose per sample on average; that is, 48 days to generate all samples for each dataset.

For Stable Diffusion sampling~\citep{podell2024sdxl}, the Diffusers library SDXL inpainting model was employed.
Input images were padded by 140 pixels on the left and right sides for masked inpainting.
The temperature was set to $0.25$, the prompt guidance scales were drawn from $\{0.25, 0.5, 0.75, 1.0\}$, the seeds from $\{0, 42, 123, 999, 2021\}$, and the prompts from a set of $10$, yielding $4\times5\times10 = 200$ samples per scene.
For the VLMs, we used ChatGPT-4o~\citep{chatgpt4}, Claude-Sonnet-3.5~\citep{claude}, and Gemini 1.5 Ultra~\citep{gemini} via their official APIs, and LLaVa-v1.6-34b~\citep{llava-1, llava-2} via the Replicate API.
The temperature was set to 0.25, the top-p value was set to 1.0 where applicable, and the order of the image--question pairs was randomly shuffled.
The VLMs were queried 100 times per object, region, and scene, resulting in 6000 calls each ($20 \, \text{images} \times 30 \, \text{questions} \times 10 \, \text{shuffles}$). 
Data acquisition and post-processing through these APIs took $\sim8$h.
Details are in \Cref{sec_additional_implementation_details,sec:appendix_vlm}.
Future updates or access restrictions to APIs used in this study could affect reproducibility.
All experiments were run on one NVIDIA RTX A6000 GPU. Our code and evaluation protocols are on the project page\footnote{\href{https://1ssb.github.io/UOD/}{https://1ssb.github.io/UOD/}}.
\subsection{Results}
\label{sec:exp_results}

\begin{table*}[!t]
\footnotesize
\centering
\caption{
Performance of all models at the 2D \task task on the \rk~\citep{stereomag} and \nyu~\citep{NYU} datasets.
We report the cross-entropy $\cent$, the prediction entropy $\ent$, the nearest neighbor distance $\dnn$, the false negative rate ($\falseneg$), and the region-wise classification accuracy $\rwa$.
The best non-oracle result is indicated in \textbf{bold}.
}
\label{tab:2D}
\vspace{-1em}
\renewcommand{\arraystretch}{1.18}
\setlength{\tabcolsep}{3pt}
\begin{tabularx}{\textwidth}{@{}l CCCCC CCCCC@{}}
\toprule
& \multicolumn{5}{c}{\textbf{\rk~\citep{stereomag}}} & \multicolumn{5}{c}{\textbf{\nyu~\citep{NYU}}} \\
\cmidrule(lr){2-6} \cmidrule(lr){7-11}
\textbf{Method} & $\cent \downarrow$ & $\ent \downarrow$ & $\dnn\% \downarrow$ & $\falseneg \downarrow$ & $\rwa \uparrow$ & $\cent \downarrow$ & $\ent \downarrow$ & $\dnn\% \downarrow$ & $\falseneg \downarrow$ & $\rwa \uparrow$ \\
\midrule
Uniform & \cells{1.000}{0.000} & \cells{1.000}{0.000} & $\infty$ & $1.000$ & \cells{0.553}{0.220} & \cells{1.000}{0.000} & \cells{1.000}{0.000} & $\infty$ & $1.000$ & \cells{0.558}{0.214} \\
Oracle & \cells{0.690}{0.123} & \cells{0.690}{0.123} & \cells{0.000}{0.000} & $0.000$ & \cells{1.000}{0.000} & \cells{0.714}{0.143} & \cells{0.714}{0.143} & \cells{0.000}{0.000} & $0.000$ & \cells{1.000}{0.000} \\
\midrule
DFM  & \cells{1.555}{1.460} & \best{{0.774}}{0.119} & \cells{5.990}{39.470} & $0.032$ & \cells{0.747}{0.139} & \cells{1.661}{1.336} & \best{{0.696}}{0.128} & \cells{6.217}{36.112} & $0.055$ & \cells{0.765}{0.236} \\
SDXL  & \cells{1.257}{2.033} & \cells{0.848}{0.171} & \best{{0.446}}{25.490} & $\textbf{0.000}$ & \best{{0.918}}{0.147} & \best{{1.223}}{2.107} & \cells{0.778}{0.212} & \best{{0.510}}{26.929} & $\textbf{0.000}$ & \best{{0.944}}{0.142} \\
\midrule
ChatGPT-4o & \cells{1.332}{0.265} & \cells{0.968}{0.056} & \cells{25.448}{13.504} & $0.024$ & \cells{0.816}{0.251} & \cells{1.415}{0.935}  & \cells{0.981}{0.043} & \cells{18.118}{26.556} & $0.000$ & \cells{0.810}{0.427} \\
Claude-3.5-Sonnet & \cells{1.880}{1.240} & \cells{0.935}{0.053} & \cells{22.107}{23.443} & $0.076$ & \cells{0.772}{0.148} & \cells{1.575}{1.553} & \cells{0.976}{0.040} & \cells{19.091}{22.609} & $0.024$ & \cells{0.750}{0.280} \\
Gemini-1.5-Ultra & \cells{1.968}{0.541} & \cells{0.968}{0.044} & \cells{24.389}{12.981} & $0.124$ & \cells{0.733}{0.286} & \cells{1.896}{0.909} & \cells{0.957}{0.040} & \cells{16.911}{15.458} & $0.039$ & \cells{0.705}{0.277} \\
LLaVa-v1.6-34b & \best{{1.232}}{0.602} & \cells{0.945}{0.050} & \cells{25.321}{15.532} & $0.262$ & \cells{0.770}{0.276} & \cells{1.455}{0.594} & \cells{0.970}{0.043} & \cells{24.369}{17.803} & $0.142$ & \cells{0.743}{0.283} \\
\bottomrule
\end{tabularx}
\end{table*}



\begin{table*}[!t]
\footnotesize 
\centering
\caption{
Performance of all models on the 2.5D and 3D \task tasks.
We report the cross-entropy $\cent$, prediction entropy $\ent$, nearest neighbor distance $\dnn$, and false negative rate $\falseneg$. The best non-oracle result is in \textbf{bold}.
}
\label{tab:2.5D}
\vspace{-1em}
\renewcommand{\arraystretch}{1.15} 
\setlength{\tabcolsep}{1.2pt} 
\begin{tabularx}{\textwidth}{@{}l CCCC CCCC CCCC CCCC@{}}
\toprule
& \multicolumn{8}{c}{\textbf{2.5D Task}} & \multicolumn{8}{c}{\textbf{3D Task}} \\
\cmidrule(lr){2-9} \cmidrule(lr){10-17}
& \multicolumn{4}{c}{\textbf{\rk~\citep{stereomag}}} & \multicolumn{4}{c}{\textbf{\nyu~\citep{NYU}}} & \multicolumn{4}{c}{\textbf{\rk~\citep{stereomag}}} & \multicolumn{4}{c}{\textbf{\nyu~\citep{NYU}}} \\
\cmidrule(lr){2-5} \cmidrule(lr){6-9} \cmidrule(lr){10-13} \cmidrule(lr){14-17}
\textbf{Method} & $\cent\downarrow$ & $\ent\downarrow$ & $\dnn(\%)\downarrow$ & $\falseneg\downarrow$ & $\cent\downarrow$ & $\ent\downarrow$ & $\dnn(\%)\downarrow$ & $\falseneg\downarrow$ & $\cent\downarrow$ & $\ent\downarrow$ & $\dnn(\%)\downarrow$ & $\falseneg\downarrow$ & $\cent\downarrow$ & $\ent\downarrow$ & $\dnn(\%)\downarrow$ & $\falseneg\downarrow$ \\
\midrule
Uniform & \cells{1.000}{0.000} & \cells{1.000}{0.000} & $\infty$ & $1.000$ & \cells{1.000}{0.000} & \cells{1.000}{0.000} & \cells{0.000}{0.000} & $1.000$ & \cells{1.000}{0.000} & \cells{1.000}{0.000} & $\infty$ & $1.000$ & \cells{1.000}{0.000} & \cells{1.000}{0.000} & $\infty$ & $1.000$ \\
Oracle & \cells{0.238}{0.014} & \cells{0.238}{0.014} & \cells{0.000}{0.000} & $0.000$ & \cells{0.297}{0.006} & \cells{0.297}{0.006} & \cells{0.000}{0.000} & $0.000$ & \cells{0.288}{0.136} & \cells{0.288}{0.136} & \cells{0.000}{0.000} & $0.000$ & \cells{0.168}{0.176} & \cells{0.168}{0.176} & \cells{0.000}{0.000} & $0.000$ \\
\midrule
DFM & \cells{1.955}{2.033} & \best{0.530}{0.441} & \best{5.128}{3.042} & $0.032$ & \cells{1.785}{2.101} & \best{0.610}{2.077} & \best{4.214}{2.803} & $0.055$ & \best{2.471}{4.415} & \best{0.303}{0.202} & \best{9.190}{6.223} & \textbf{0.017} & \best{2.062}{3.915} & \best{0.541}{0.374} & \best{3.948}{5.801} & \textbf{0.000} \\
SDXL & \best{1.752}{2.043} & \cells{0.617}{0.305} & \cells{6.277}{2.229} & \textbf{0.000} & \best{1.533}{2.916} & \cells{0.655}{0.323} & \cells{5.245}{2.351} & \textbf{0.000} & -- & -- & -- & -- & -- & -- & -- & -- \\
\bottomrule
\end{tabularx}
\end{table*}

\begin{table*}[!t]
\renewcommand{\cells}[2]{#1}
\renewcommand{\best}[2]{\textbf{#1}}
\footnotesize
\centering
\caption{
Ablation and analysis on the combined \rk and \nyu dataset. The {best variant} of each method is in \textbf{bold}.
}
\vspace{-1em}
\label{tab:ablations}
\renewcommand{\arraystretch}{1.15}
\setlength{\tabcolsep}{1.7pt}
\begin{tabularx}{\textwidth}{@{}l CCCCC CCCC CCCC@{}}
\toprule
& \multicolumn{5}{c}{\textbf{2D Task}} & \multicolumn{4}{c}{\textbf{2.5D Task}} & \multicolumn{4}{c}{\textbf{3D Task}}\\
\cmidrule(lr){2-6} \cmidrule(lr){7-10}  \cmidrule(lr){11-14}
\textbf{Method} & $\cent\downarrow$ & $\ent\downarrow$ & $\dnn(\%)\downarrow$ & $\falseneg\downarrow$ & $\rwa \uparrow$ & $\cent\downarrow$ & $\ent\downarrow$ & $\dnn(\%)\downarrow$ & $\falseneg\downarrow$ & $\cent\downarrow$ & $\ent\downarrow$ & $\dnn(\%)\downarrow$ & $\falseneg\downarrow$ \\
\midrule


DFM (25 samples, 4 poses) & \best{1.595}{1.414} & \best{0.744}{0.122} & \best{6.076}{38.229} & \best{0.04}{N/A} & \best{0.754}{0.182} & \best{1.890}{1.529} & \best{0.560}{0.370} & \best{4.781}{2.953} & \best{0.04}{N/A} & \best{2.303}{4.217} & \best{0.401}{0.285} & \best{7.041}{6.054} & \best{0.01}{N/A} \\

DFM (25 samples, 3 poses) & \cells{1.944}{1.067} & \cells{0.860}{0.126} & \cells{19.019}{33.547} & \cells{0.14}{N/A} & \cells{0.612}{0.226} & \cells{2.037}{0.071} & \cells{0.808}{0.084} & \cells{12.339}{1.163} & \cells{0.13}{N/A} & \cells{3.226}{6.827} & \cells{0.558}{0.313} & \cells{8.797}{6.703} & \cells{0.08}{N/A} \\

DFM (25 samples, 2 poses) & \cells{2.466}{1.099} & \cells{0.931}{0.069} & \cells{14.734}{43.179} & \cells{0.88}{N/A} & \cells{0.567}{0.185} & \cells{2.562}{2.523} & \cells{0.994}{0.004} & \cells{13.205}{1.643} & \cells{0.86}{N/A} & \cells{4.931}{5.973} & \cells{0.703}{0.354} & \cells{11.105}{7.375} & \cells{0.38}{N/A} \\

DFM (15 samples, 4 poses) & \cells{1.804}{0.447} & \cells{0.925}{0.094} & \cells{16.059}{24.313} & \cells{0.58}{N/A} & \cells{0.592}{0.268} & \cells{2.140}{0.733} & \cells{0.873}{0.072} & \cells{8.759}{6.272} & \cells{0.67}{N/A} & \cells{4.047}{5.321} & \cells{0.858}{0.431} & \cells{9.980}{6.794} & \cells{0.07}{N/A} \\

DFM (10 samples, 4 poses) & \cells{2.342}{1.264} & \cells{0.952}{0.013} & \cells{14.787}{29.830} & \cells{0.72}{N/A} & \cells{0.574}{0.344} & \cells{2.480}{1.309} & \cells{0.996}{0.002} & \cells{17.678}{9.530} & \cells{0.75}{N/A} & \cells{4.695}{3.633} & \cells{0.965}{0.451} & \cells{12.481}{7.388} & \cells{0.22}{N/A} \\

\midrule

SDXL (w/ object cues) & \best{1.244}{2.061} & \best{0.870}{0.176} & \best{0.470}{26.046} & \best{0.00}{N/A} & \best{0.928}{0.145} & \best{1.668}{2.412} & \best{0.631}{0.311} & \best{5.884}{2.276} & \best{0.00}{--} & -- & -- & -- & --  \\

SDXL (w/o object cues) & \cells{1.977}{--} & \cells{0.885}{--} & \cells{6.037}{14.951} & \cells{0.49}{N/A} & \cells{0.716}{0.277} & \cells{2.129}{1.803} & \cells{0.860}{0.103} & \cells{16.051}{7.449} & \cells{0.48}{--} & -- & -- & -- & --  \\

SDXL (w/o prompts) & \cells{2.963}{--} & \cells{0.926}{0.084} & \cells{6.482}{13.643} & \cells{0.51}{N/A} & \cells{0.767}{0.279} & \cells{2.229}{1.855} & \cells{0.952}{0.014} & \cells{17.949}{8.531} & \cells{0.51}{--} & -- & -- & -- & -- \\

\bottomrule
\end{tabularx}
\end{table*}
In \cref{tab:2D,tab:2.5D}, we report the performance of the generative models at the 2D, 2.5D and 3D \task tasks. 
The metrics are reported as averages and standard deviations over scene--object combinations.
The distance average excludes $\dnn=\infty$ entries; these are captured by the false negative rate. 
Methods were only evaluated on tasks for which they are applicable: we lift the VLM outputs from their binary labels to a 2D distribution, and lift the 2D diffusion outputs to a 2.5D distribution, but cannot obtain the 3D distribution from either.
For the 2D task (\cref{tab:2D}), the 2D diffusion model shows the best performance, while the 3D diffusion model remains competitive. The coarse-grained VLMs struggle in most metrics with large spatial errors, $\dnn$, as expected.
For the 2.5D task (\cref{tab:2.5D}, left), the 3D diffusion model consistently outperforms the 2D diffusion model (SDXL)~\citep{podell2024sdxl, rombach2022high}, augmented with monocular depth estimates \cite{depthanythingv2}, with the exception of cross-entropy and the false negative rate when the outpainting model is prompted with the target object.
Finally, for the 3D task (\cref{tab:2.5D}, right), we see that the 3D diffusion model remains competitive.
Lower values of $\cent$ and $\dnn$ indicate better agreement with  $\gtsd$.
For example, a low $\dnn$ indicates that the predicted modes that overlap with those in $\gtsd$.

\begin{figure*}[!t]
    \centering
    \rotatebox[origin=l]{90}{{\footnotesize TV}}
    \begin{subfigure}[]{0.27\linewidth}
        \centering
        \begin{tikzpicture}
            \node [anchor=south west,inner sep=0] at (0,0) {\includegraphics[width=\linewidth]{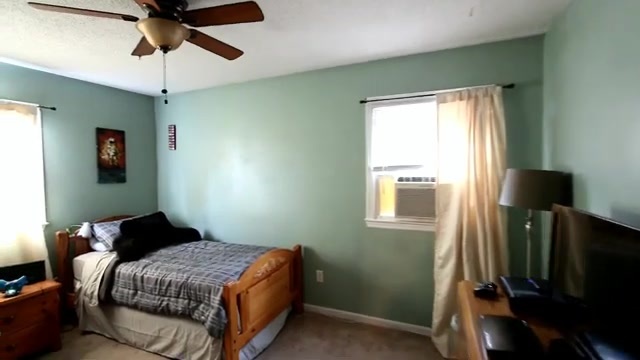}};
            \draw[white, thick, dashed] (0.22*\linewidth, 0.01) -- (0.22*\linewidth, 0.5648*\linewidth);
            \draw[white, thick, dashed] (0.78*\linewidth, 0.01) -- (0.78*\linewidth, 0.5648*\linewidth);
        \end{tikzpicture}
        \label{fig:qual_row1_col1}
    \end{subfigure}\hfill
    \begin{subfigure}[]{0.152\linewidth}
        \centering
        \includegraphics[width=\linewidth]{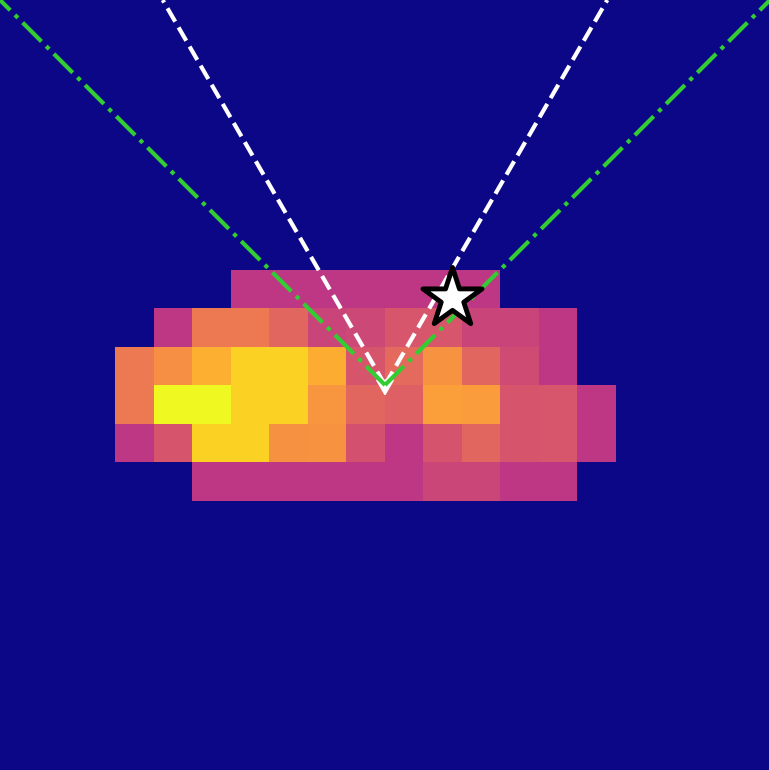}
        \label{fig:qual_row1_col2}
    \end{subfigure}\hfill
    \begin{subfigure}[]{0.27\linewidth}
        \centering
        \includegraphics[width=\linewidth]{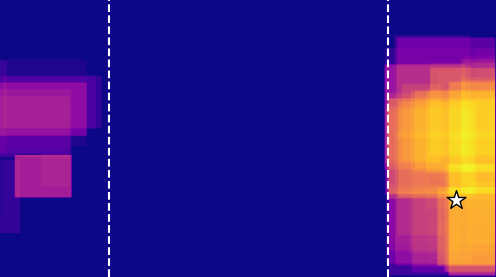}
        \label{fig:qual_row1_col3}
    \end{subfigure}\hfill
    \begin{subfigure}[]{0.27\linewidth}
        \centering
        \includegraphics[width=\linewidth]{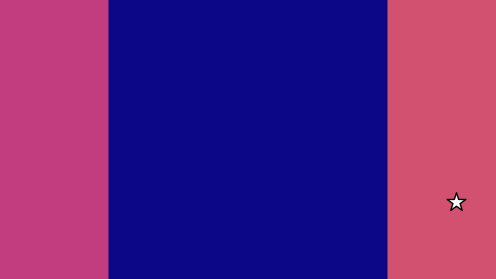}
        \label{fig:qual_row1_col4}
    \end{subfigure}\\
    \vspace{-1em}
    \rotatebox[origin=l]{90}{{\footnotesize \hspace{-3em} REFRIGERATOR}}
    \begin{subfigure}[]{0.27\linewidth}
        \centering
        \begin{tikzpicture}
            \node [anchor=south west,inner sep=0] at (0,0) {\includegraphics[width=\linewidth]{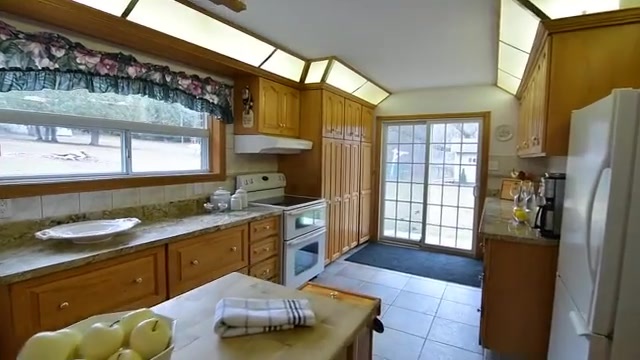}};
            \draw[white, thick, dashed] (0.22*\linewidth, 0.01) -- (0.22*\linewidth, 0.5648*\linewidth);
            \draw[white, thick, dashed] (0.78*\linewidth, 0.01) -- (0.78*\linewidth, 0.5648*\linewidth);
        \end{tikzpicture}
        \label{fig:qual_row2_col1}
    \end{subfigure}\hfill
    \begin{subfigure}[]{0.152\linewidth}
        \centering
        \includegraphics[width=\linewidth]{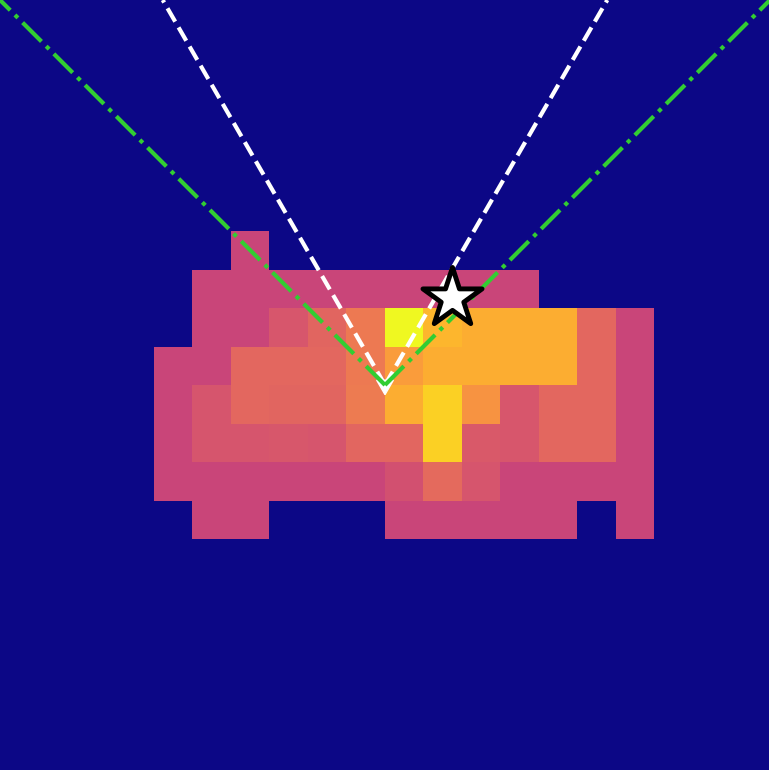}
        \label{fig:qual_row2_col2}
    \end{subfigure}\hfill
    \begin{subfigure}[]{0.27\linewidth}
        \centering
        \includegraphics[width=\linewidth]{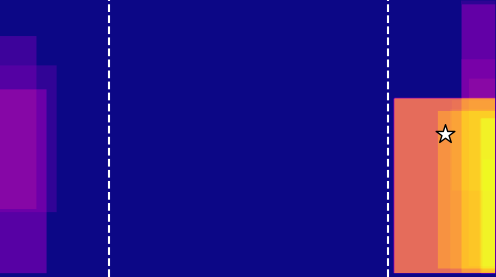}
        \label{fig:qual_row2_col3}
    \end{subfigure}\hfill
    \begin{subfigure}[]{0.27\linewidth}
        \centering
        \includegraphics[width=\linewidth]{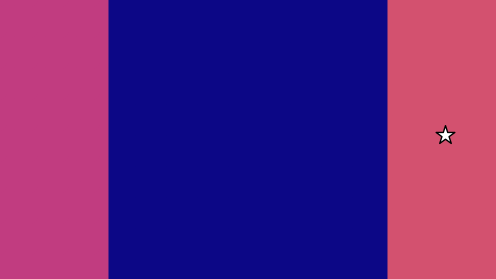}
        \label{fig:qual_row2_col4}
    \end{subfigure}\\
    \vspace{-1em}
    \rotatebox[origin=l]{90}{{\footnotesize SINK}}
    \begin{subfigure}[]{0.27\linewidth}
        \centering
        \begin{tikzpicture}
            \node [anchor=south west,inner sep=0] at (0,0) {\includegraphics[width=\linewidth]{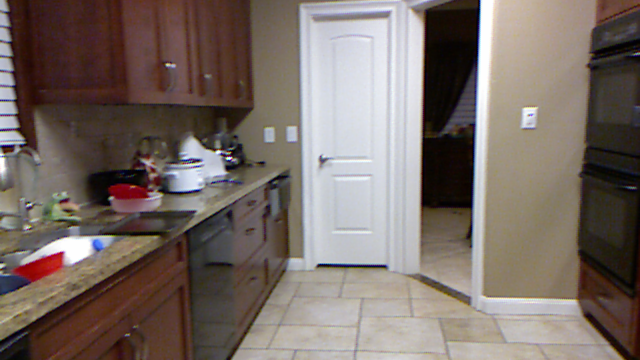}};
            \draw[white, thick, dashed] (0.22*\linewidth, 0.01) -- (0.22*\linewidth, 0.5648*\linewidth);
            \draw[white, thick, dashed] (0.78*\linewidth, 0.01) -- (0.78*\linewidth, 0.5648*\linewidth);
        \end{tikzpicture}
        \label{fig:qual_row3_col1}
    \end{subfigure}\hfill
    \begin{subfigure}[]{0.152\linewidth}
        \centering
        \includegraphics[width=\linewidth]{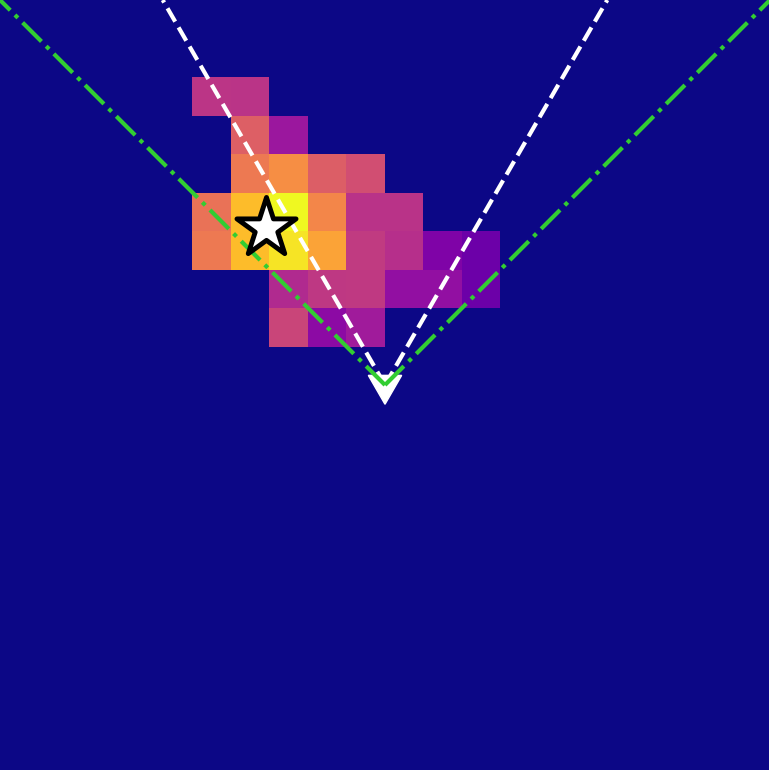}
        \label{fig:qual_row3_col2}
    \end{subfigure}\hfill
    \begin{subfigure}[]{0.27\linewidth}
        \centering
        \includegraphics[width=\linewidth]{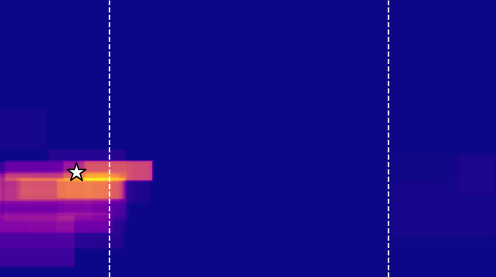}
        \label{fig:qual_row3_col3}
    \end{subfigure}\hfill
    \begin{subfigure}[]{0.27\linewidth}
        \centering
        \includegraphics[width=\linewidth]{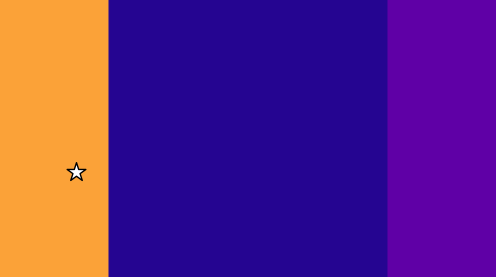}
        \label{fig:qual_row3_col4}
    \end{subfigure}\\%
    \vspace{-1em}
    \rotatebox[origin=l]{90}{{\footnotesize \hspace{-1em} LAPTOP}}
    \begin{subfigure}[]{0.27\linewidth}
        \centering
        \begin{tikzpicture}
            \node [anchor=south west,inner sep=0] at (0,0) {\includegraphics[width=\linewidth]{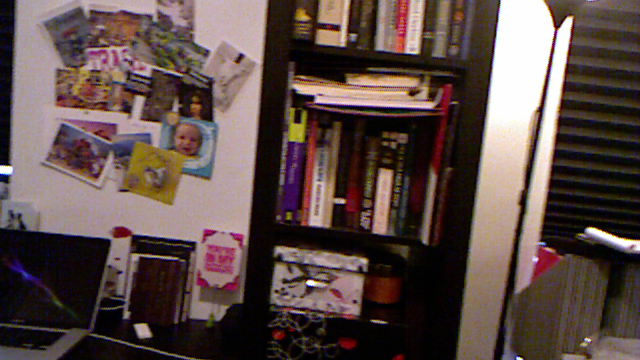}};
            \draw[white, thick, dashed] (0.22*\linewidth, 0.01) -- (0.22*\linewidth, 0.5648*\linewidth);
            \draw[white, thick, dashed] (0.78*\linewidth, 0.01) -- (0.78*\linewidth, 0.5648*\linewidth);
        \end{tikzpicture}
        \label{fig:qual_row4_col1}
    \end{subfigure}\hfill
    \begin{subfigure}[]{0.152\linewidth}
        \centering
        \includegraphics[width=\linewidth]{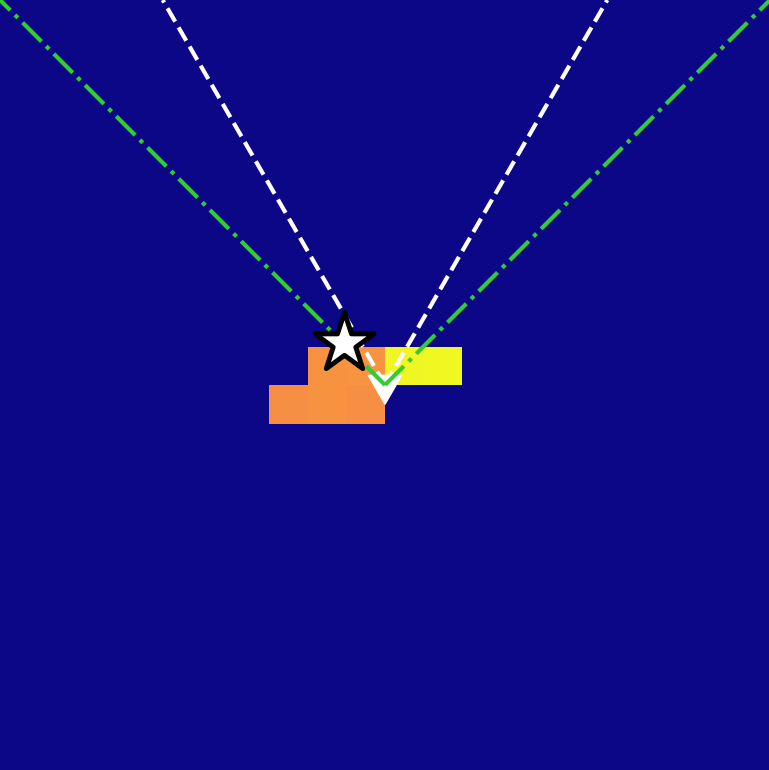}
        \label{fig:qual_row4_col2}
    \end{subfigure}\hfill
    \begin{subfigure}[]{0.27\linewidth}
        \centering
        \includegraphics[width=\linewidth]{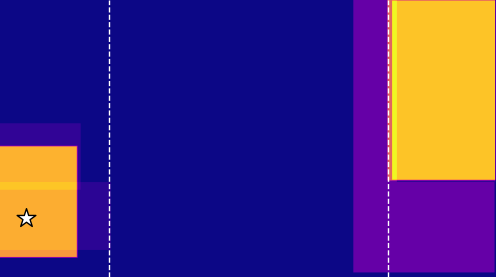}
        \label{fig:qual_row4_col3}
    \end{subfigure}\hfill
    \begin{subfigure}[]{0.27\linewidth}
        \centering
        \includegraphics[width=\linewidth]{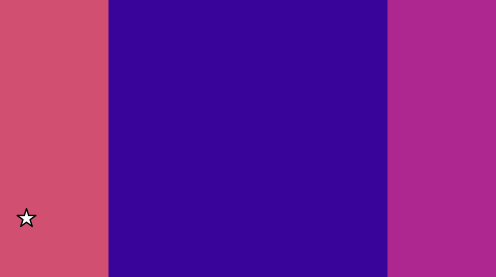}
        \label{fig:qual_row4_col4}
    \end{subfigure}\\%
    \vspace{-1em}
    \rotatebox[origin=l]{90}{\footnotesize SINK}
    \begin{subfigure}[]{0.27\linewidth}
        \centering
        \begin{tikzpicture}
            \node [anchor=south west,inner sep=0] at (0,0) {\includegraphics[width=\linewidth]{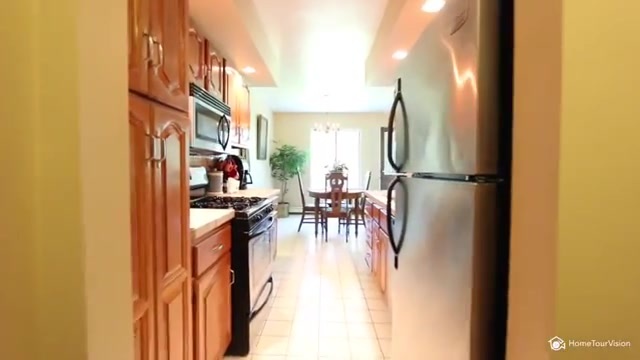}};
            \draw[white, thick, dashed] (0.22*\linewidth, 0.01) -- (0.22*\linewidth, 0.5648*\linewidth);
            \draw[white, thick, dashed] (0.78*\linewidth, 0.01) -- (0.78*\linewidth, 0.5648*\linewidth);
        \end{tikzpicture}
        \caption{$\inim$ (center) and $\outim$ (full)}
        \label{fig:qual_row6_col1}
    \end{subfigure}\hfill
    \begin{subfigure}[]{0.152\linewidth}
        \centering
        \includegraphics[width=\linewidth]{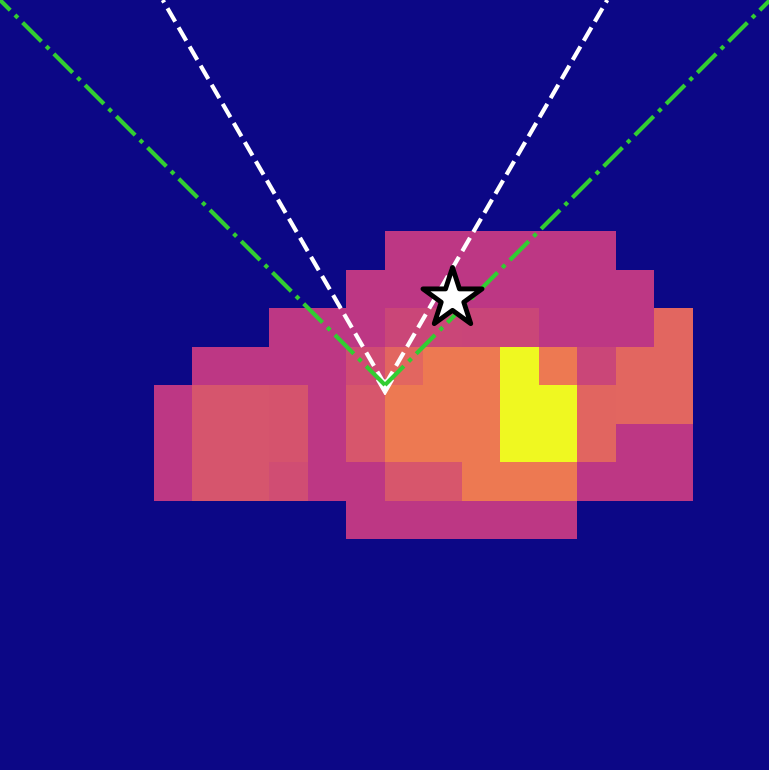}
        \caption{DFM $\sd^\text{3D}$}
        \label{fig:qual_row6_col2}
    \end{subfigure}\hfill
    \begin{subfigure}[]{0.27\linewidth}
        \centering
        \includegraphics[width=\linewidth]{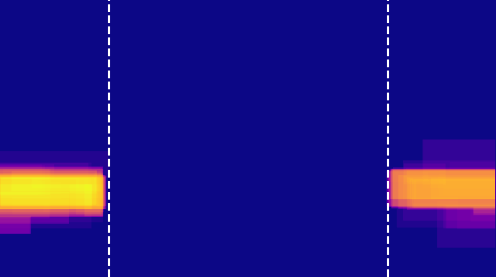}
        \caption{SDXL $\sd^\text{2D}$}
        \label{fig:qual_row6_col3}
    \end{subfigure}\hfill
    \begin{subfigure}[]{0.27\linewidth}
        \centering
        \includegraphics[width=\linewidth]{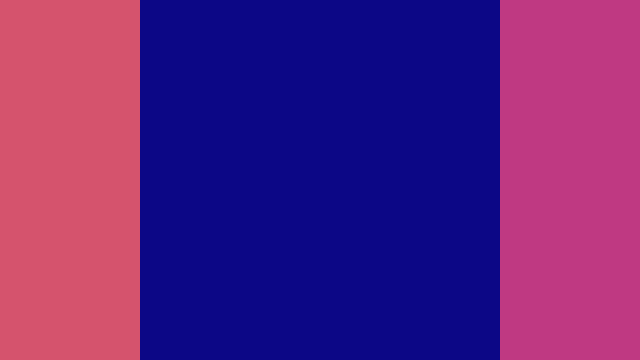}
        \caption{ChatGPT-4o $\sd^\text{2D}$}
        \label{fig:qual_row6_col4}
    \end{subfigure}\vfill
    \vspace{3pt}
    {\tt 0}\ \includegraphics[width=0.5\linewidth, trim={12pt 8pt 10pt 8pt}, clip]{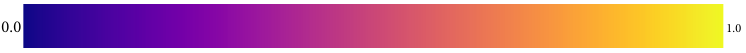}\ {\tt 1}
    \caption{\textbf{Qualitative results.}
    Each row shows the predicted 2D and top-down 3D spatial distributions generated by each method for various object categories: TV (first row), refrigerator (second row), sink (third row), laptop (fourth row), and sink (fifth row). Notably, in the bottom row, the DFM-based model infers the likely presence of a sink, occluded by the refrigerator, albeit not with a high likelihood.
    A white triangle marks the camera position, while dashed and dot-dashed lines depict the camera frustums for $\inim$ and $\outim$.
    The white star indicates the ground-truth position of the object, when visible in 2D.
    Heatmap colors indicate object likelihood, with warmer tones representing higher probabilities.
    Since these are spatially-normalized distributions, we use a log-scale for visualization.
    }
    \label{fig:qual}
\end{figure*}

\label{sec:fail}
\begin{figure*}[ht!]
    \centering
    \begin{subfigure}[]{0.27\linewidth}
        \centering
        \begin{tikzpicture}
            \node [anchor=south west,inner sep=0] at (0,0) {\includegraphics[width=\linewidth]{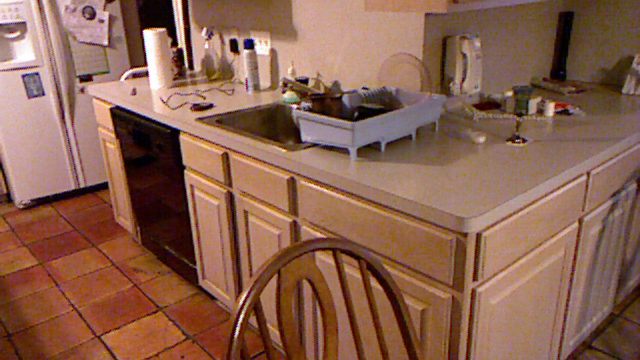}};
            \draw[white, thick, dashed] (0.22*\linewidth, 0.01) -- (0.22*\linewidth, 0.5648*\linewidth); %
            \draw[white, thick, dashed] (0.78*\linewidth, 0.01) -- (0.78*\linewidth, 0.5648*\linewidth); %
        \end{tikzpicture}
        \label{fig:f11}
    \end{subfigure}
    \begin{subfigure}[]{0.15\linewidth}
        \centering
        \includegraphics[width=\linewidth]{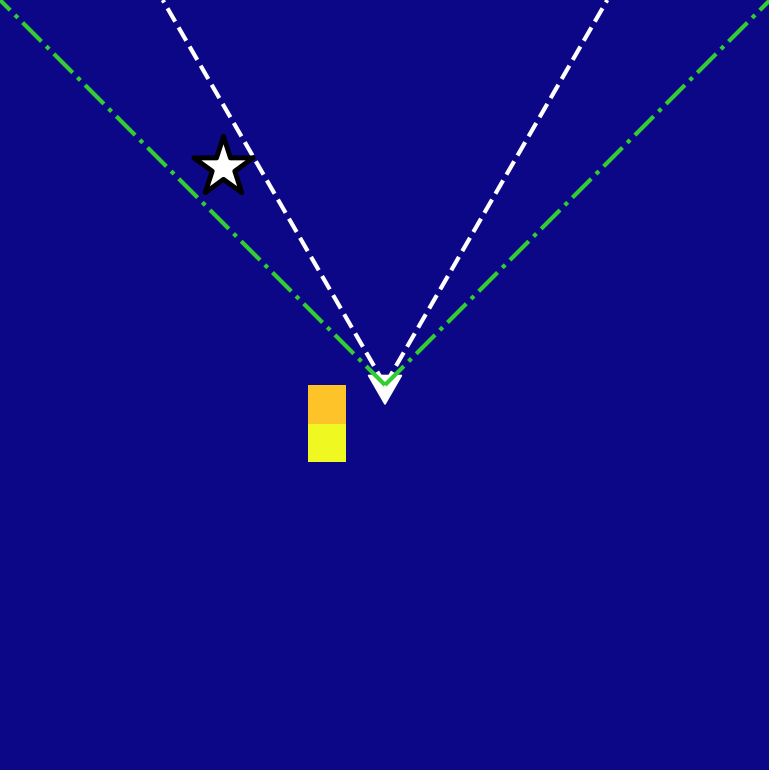}
        \label{fig:f12}
    \end{subfigure}
    \begin{subfigure}[]{0.27\linewidth}
        \centering
        \includegraphics[width=\linewidth]{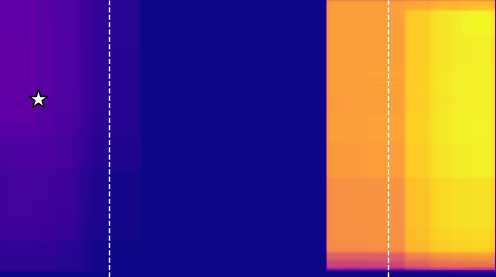}
        \label{fig:f10}
    \end{subfigure}
    \begin{subfigure}[]{0.27\linewidth}
        \centering        \includegraphics[width=\linewidth]{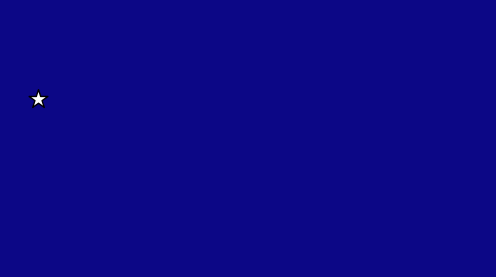}
        \label{fig:f14}
    \end{subfigure}%
    \vspace{-10pt}
    
    \begin{subfigure}[]{0.27\linewidth}
        \centering
        \begin{tikzpicture}
            \node [anchor=south west,inner sep=0] at (0,0) {\includegraphics[width=\linewidth]{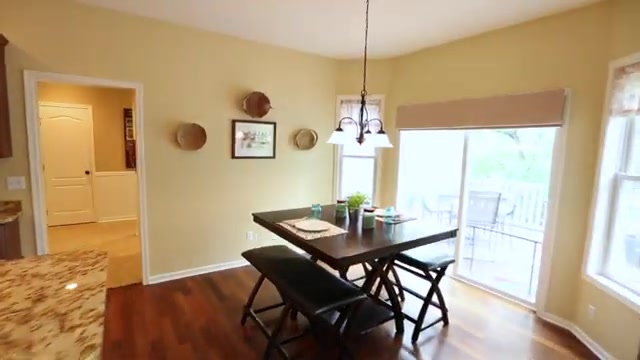}};
            \draw[white, thick, dashed] (0.22*\linewidth, 0.01) -- (0.22*\linewidth, 0.5648*\linewidth); %
            \draw[white, thick, dashed] (0.78*\linewidth, 0.01) -- (0.78*\linewidth, 0.5648*\linewidth); %
        \end{tikzpicture}
        \caption{$\inim$ (center) and $\outim$ (full)}
        \label{fig:failure_11}
    \end{subfigure}
    \begin{subfigure}[]{0.15\linewidth}
        \centering
        \includegraphics[width=\linewidth]{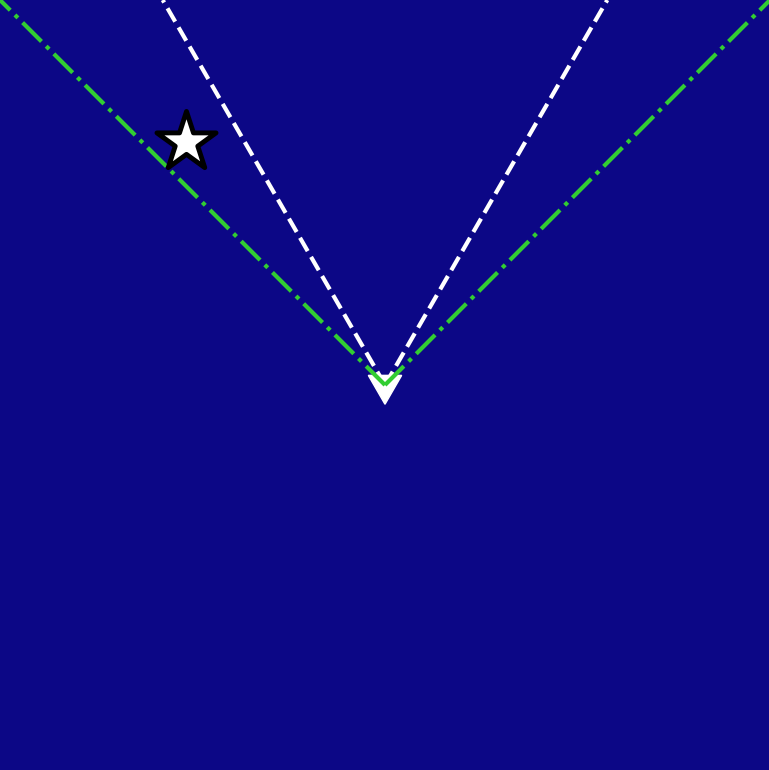}
        \caption{DFM-based $\sd^\text{3D}$}
        \label{fig:failure_12}
    \end{subfigure}
    \begin{subfigure}[]{0.27\linewidth}
        \centering
        \includegraphics[width=\linewidth]{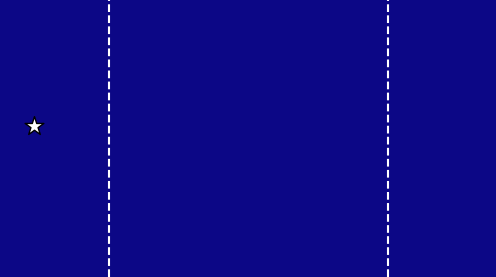}
        \caption{SDXL-based $\sd^\text{2D}$}
        \label{fig:failure_14}
    \end{subfigure}
    \begin{subfigure}[]{0.27\linewidth}
        \centering
        \includegraphics[width=\linewidth]{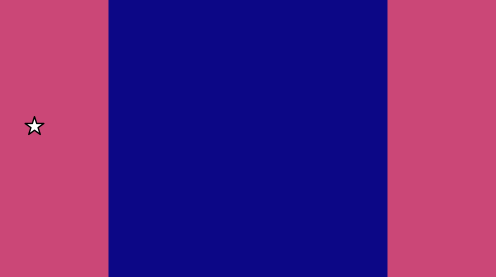}
        \caption{ChatGPT-based $\sd^\text{2D}$}
        \label{fig:failure_15}
    \end{subfigure}%
    \vspace{0.35em}
    \begin{subfigure}[]{0.322\linewidth}\centering
        \includegraphics[width=\linewidth]{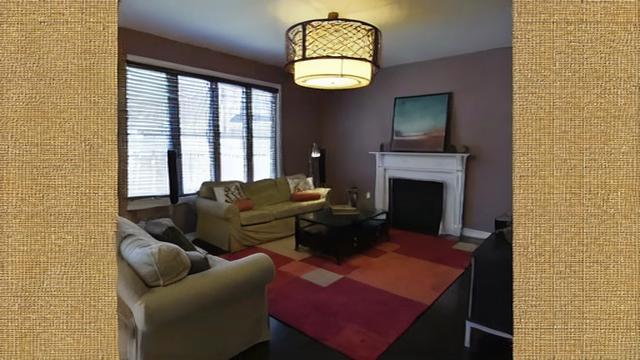}
        \caption{SDXL w/o object cue prompt failure}
        \label{fig:sd11}
    \end{subfigure}
    \begin{subfigure}[]{0.322\linewidth}\centering
        \includegraphics[width=\linewidth]{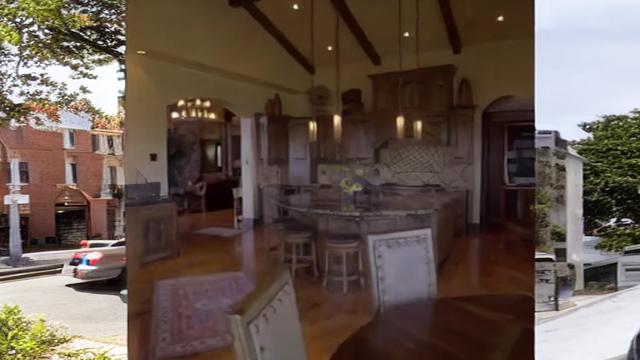}
        \caption{SDXL w/o object cue prompt failure}
        \label{fig:sd12}
    \end{subfigure}
    \begin{subfigure}[]{0.322\linewidth}\centering
        \includegraphics[width=\linewidth]{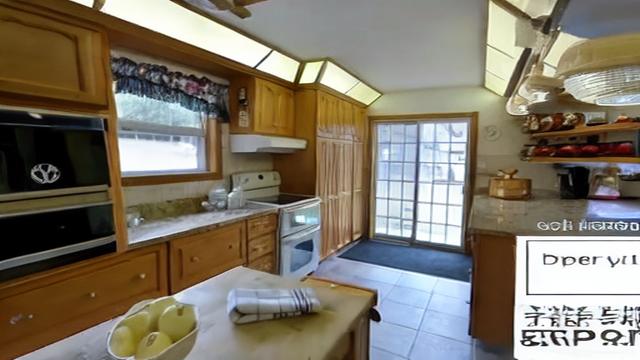}
        \caption{Failure with object cues}
        \label{fig:sd13}
    \end{subfigure}%
    \vspace{-6px}
    \caption{\textbf{Failure cases.} 
    In the first row, the DFM-based model  misses plausible locations where a refrigerator might be and the ChatGPT-4o-based model fails to report any detections.
    In the second row, the presence of the door on the left is not inferred by the DFM or SDXL-based models, primarily due to the object detector's false negatives on synthetic images for that class.
    The last row shows outpainting failure cases.
    The SDXL outpainting model, when prompted without object cues, often extends images with textures \subref{fig:sd11} or unrelated imagery \subref{fig:sd12}.
    Even when object cues (here, ``refrigerator'') are provided \subref{fig:sd13}, results may ignore the prompt and generate unnaturally extended images.
    }
    \label{fig:failure_cases}
\end{figure*}

Qualitative results are shown in \cref{fig:qual}, with the $360 \times 360$ center-cropped input image $\inim$ shown within the dashed lines in the first column, a top-down view of the DFM-predicted 3D distribution in the second column, the SDXL-predicted 2D distribution in the third column, and the ChatGPT-predicted 2D distribution in the last column.
In particular, the DFM-based model usually predicts high likelihoods near the ground-truth object's 3D location, including when the object is occluded or out-of-frame, which is a significantly harder task than the 2D prediction task; see, for example, the final row. 
More qualitative results can be found in~\cref{subsec:qual}.
Failure cases are shown in \cref{fig:failure_cases}, where all models perform poorly and an object category (door) that elicits many false negatives in the object detector.
We also observe that the diffusion-based outpainting approach has its own modes of failure, also shown in \cref{fig:failure_cases},
where the results are not informative in a scene if the model is not prompted with the label of the object to be detected.

\subsection{Ablation Studies \& Analysis}
\label{sec:exp_ablation}
In \cref{tab:ablations}, we report the ablation study and analysis for the 2D and 3D diffusion-based detectors.
First, we analyze the effect of reducing the number of camera poses $k$ at which the DFM model is queried, from $4$ to $2$, which affects the coverage of the scene.
To do so, we randomly drop one or two camera poses and report the performance.
We find that the performance decreases considerably as the number of camera poses decreases, across all tasks.
Second, we analyze the effect of reducing the number of conditionally-independent samples per additional camera pose from $25$ to $10$, which reduces the diversity of the predicted conditional distribution which negatively affects performance.
We also ablate the 2D diffusion-based approach, investigating the effect of object cues in the text prompts on the results, evaluating the pipeline
(a) without  prompts; 
(b) without object cues, where the model is provided with text guidance but without referring to objects (\eg, \say{extend this indoor scene naturally to the left and right of the given frame}); and
(c) with object cues in the prompts (\eg, \say{extend this scene naturally on both sides, with objects like [OBJECT]}).
Prompts with object cues appear to significantly impact performance of the SDXL-based method in the task, indicating that these methods make heavy use of textual guidance in the generative process, as demonstrated in the 2D and 2.5D ablation study.
Details of the complete analysis and ablation study can be found in the \cref{appendix:ablations}. We used default hyperparameters without any dataset-specific changes for all our experiments in the generation process (details in \cref{sec:appendix_scene,sec_additional_implementation_details,sec:appendix_vlm}).
\section{Conclusion}
\label{sec:conclusion}
This work has introduced the \task task in 2D, 2.5D and 3D, has presented three detection pipelines using state-of-the-art generative models, and has evaluated their performance on a proposed set of metrics in a dataset of indoor scenes and common objects.
We find that \task presents different characteristics in 2D and 3D.
Expanding the camera frame can be used in both 2D and 3D, with detection pipelines showing the ability to predict objects that may be fully outside the visible frustum.
In contrast, predicting objects within 3D regions beyond the camera frustum suggests that 3D pipelines have the capability to detect fully occluded objects. While the 2D SDXL-based pipeline and VLMs cannot detect occluded objects, the 3D pipeline leveraging DFM achieves this.

This work has several significant limitations. 
In particular, the inference time of 3D diffusion models, is currently very large and are costly to re-train.
Since our approaches require many samples to approximate the conditional distribution, this is a significant bottleneck and stymies their use in real-time applications like in robotics.
Another limitation is the need for object prompts in outpainting-based methods, since this is likely to bias the obtained empirical conditional distribution.
Also, VLM-based evaluations may not be fully reliable, since it is possible these models could have seen and memorized the test images during training.
Since the task requires that the model not over-fit to the ground truth realization---which is an instance of the true conditional distribution---there are inherent challenges in measuring performance that reflects
accurate modeling of the underlying true distribution that is unavailable in real-world settings.

Future applications will apply the proposed pipeline to outdoor and large-scale scenes~\cite{dl3dv}.  Synthetic datasets~\citep{minigrid} can let us access known conditional distributions in simulation~\citep{robothor}. Beyond object detection, different spatio-semantic features like occupancy or affordances~\citep{saycan} are promising next steps.
Improvements in solving this task could pave the way for robotics applications in reactive control~\citep{react}, visual search~\citep{vsearch}, planning~\citep{jutrasdube2024adaptive}, navigation~\citep{nav_real}, and exploration~\citep{explore}.
This motivates the current work as a significant step toward improving machine vision capabilities for unobserved object detection.
\FloatBarrier
\clearpage
\newpage
\section*{Acknowledgments}
Subhransu is supported by the University Research Scholarship at the Australian National University. This research was partially funded by the U.S.\ Government under DARPA TIAMAT HR00112490421. The views and conclusions expressed in this document are solely those of the authors and do not represent the official policies or endorsements, either expressed or implied, of the U.S.\ Government. This research was also funded by the Australian Research Council under ITRH IH210100030.

{
    \small
    \bibliographystyle{ieeenat_fullname}
    \bibliography{main}

\begin{thebibliography}{90}
\providecommand{\natexlab}[1]{#1}
\providecommand{\url}[1]{\texttt{#1}}
\expandafter\ifx\csname urlstyle\endcsname\relax
  \providecommand{\doi}[1]{doi: #1}\else
  \providecommand{\doi}{doi: \begingroup \urlstyle{rm}\Url}\fi

\bibitem[Adamkiewicz et~al.(2022)Adamkiewicz, Chen, Caccavale, Gardner, Culbertson, Bohg, and Schwager]{nerf-nav}
Michal Adamkiewicz, Timothy Chen, Adam Caccavale, Rachel Gardner, Preston Culbertson, Jeannette Bohg, and Mac Schwager.
\newblock Vision-only robot navigation in a neural radiance world.
\newblock \emph{IEEE Robotics and Automation Letters}, 7\penalty0 (2), 2022.

\bibitem[Anciukevičius et~al.(2023)Anciukevičius, Xu, Fisher, Henderson, Bilen, Mitra, and Guerrero]{render}
Titas Anciukevičius, Zexiang Xu, Matthew Fisher, Paul Henderson, Hakan Bilen, Niloy~J. Mitra, and Paul Guerrero.
\newblock {RenderDiffusion}: Image diffusion for {3D} reconstruction, inpainting and generation.
\newblock In \emph{Proceedings of the IEEE/CVF Conference on Computer Vision and Pattern Recognition}, 2023.

\bibitem[Anthropic(2024)]{claude}
Anthropic.
\newblock {Claude 3.5 Sonnet}.
\newblock \url{https://www.anthropic.com/news/claude-3-5-sonnet}, 2024.

\bibitem[Aydemir et~al.(2013)Aydemir, Pronobis, Göbelbecker, and Jensfelt]{vsearch}
Alper Aydemir, Andrzej Pronobis, Moritz Göbelbecker, and Patric Jensfelt.
\newblock Active visual object search in unknown environments using uncertain semantics.
\newblock \emph{IEEE Transactions on Robotics}, 29\penalty0 (4), 2013.

\bibitem[Back et~al.(2022)Back, Lee, Kim, Noh, Kang, Bak, and Lee]{back2022unseen}
Seunghyeok Back, Joosoon Lee, Taewon Kim, Sangjun Noh, Raeyoung Kang, Seongho Bak, and Kyoobin Lee.
\newblock Unseen object amodal instance segmentation via hierarchical occlusion modeling.
\newblock In \emph{Proceedings of the International Conference on Robotics and Automation}, 2022.

\bibitem[Bautista et~al.(2022)Bautista, Guo, Abnar, Talbott, Toshev, Chen, Dinh, Zhai, Goh, Ulbricht, Dehghan, and Susskind]{bautista2022gaudi}
Miguel~{\'A}ngel Bautista, Pengsheng Guo, Samira Abnar, Walter Talbott, Alexander~T Toshev, Zhuoyuan Chen, Laurent Dinh, Shuangfei Zhai, Hanlin Goh, Daniel Ulbricht, Afshin Dehghan, and Joshua~M. Susskind.
\newblock {GAUDI}: A neural architect for immersive 3{D} scene generation.
\newblock In \emph{Proceedings of the Conference on Neural Information Processing Systems}, 2022.

\bibitem[Bordes et~al.(2023)Bordes, Shekhar, Ibrahim, Bouchacourt, Vincent, and Morcos]{bordes2023pug}
Florian Bordes, Shashank Shekhar, Mark Ibrahim, Diane Bouchacourt, Pascal Vincent, and Ari Morcos.
\newblock {PUG:} {P}hotorealistic and semantically controllable synthetic data for representation learning.
\newblock In \emph{Proceedings of the Conference on Neural Information Processing Systems}, 2023.

\bibitem[Carion et~al.(2020)Carion, Massa, Synnaeve, Usunier, Kirillov, and Zagoruyko]{detr}
Nicolas Carion, Francisco Massa, Gabriel Synnaeve, Nicolas Usunier, Alexander Kirillov, and Sergey Zagoruyko.
\newblock End-to-end object detection with transformers.
\newblock In \emph{Proceedings of the European Conference on Computer Vision}, 2020.

\bibitem[Carlone et~al.(2015)Carlone, Tron, Daniilidis, and Dellaert]{carlone}
Luca Carlone, Roberto Tron, Kostas Daniilidis, and Frank Dellaert.
\newblock Initialization techniques for {3D SLAM}: A survey on rotation estimation and its use in pose graph optimization.
\newblock In \emph{Proceedings of the IEEE International Conference on Robotics and Automation}, 2015.

\bibitem[Chan et~al.(2023)Chan, Nagano, Chan, Bergman, Park, Levy, Aittala, Mello, Karras, and Wetzstein]{GeNVS}
Eric~R. Chan, Koki Nagano, Matthew~A. Chan, Alexander~W. Bergman, Jeong~Joon Park, Axel Levy, Miika Aittala, Shalini~De Mello, Tero Karras, and Gordon Wetzstein.
\newblock Generative novel view synthesis with {{3D}}-aware diffusion models.
\newblock In \emph{Proceedings of the IEEE/CVF International Conference on Computer Vision}, 2023.

\bibitem[Chen et~al.(2024)Chen, Xu, Kirmani, Ichter, Driess, Florence, Sadigh, Guibas, and Xia]{spatialvlm}
Boyuan Chen, Zhuo Xu, Sean Kirmani, Brian Ichter, Danny Driess, Pete Florence, Dorsa Sadigh, Leonidas Guibas, and Fei Xia.
\newblock {SpatialVLM}: Endowing vision-language models with spatial reasoning capabilities.
\newblock In \emph{Proceedings of the IEEE/CVF Conference on Computer Vision and Pattern Recognition}, 2024.

\bibitem[Chevalier{-}Boisvert et~al.(2023)Chevalier{-}Boisvert, Dai, Towers, Perez{-}Vicente, Willems, Lahlou, Pal, Castro, and Terry]{minigrid}
Maxime Chevalier{-}Boisvert, Bolun Dai, Mark Towers, Rodrigo Perez{-}Vicente, Lucas Willems, Salem Lahlou, Suman Pal, Pablo~Samuel Castro, and Jordan Terry.
\newblock Minigrid {\&} {M}iniworld: Modular {\&} customizable reinforcement learning environments for goal-oriented tasks.
\newblock In \emph{Proceedings of the Conference on Neural Information Processing Systems}, 2023.

\bibitem[Cirik et~al.(2022)Cirik, Morency, and Berg-Kirkpatrick]{cirik-etal-2022-holm}
Volkan Cirik, Louis-Philippe Morency, and Taylor Berg-Kirkpatrick.
\newblock {HOLM}: Hallucinating objects with language models for referring expression recognition in partially-observed scenes.
\newblock In \emph{Proceedings of the Annual Meeting of the Association for Computational Linguistics}, 2022.

\bibitem[Cover and Thomas(2006)]{cover2006}
Thomas~M. Cover and Joy~A. Thomas.
\newblock Differential entropy.
\newblock In \emph{Elements of Information Theory}. Wiley-Interscience, 2nd edition, 2006.

\bibitem[Deitke et~al.(2020)Deitke, Han, Herrasti, Kembhavi, Kolve, Mottaghi, Salvador, Schwenk, VanderBilt, Wallingford, Weihs, Yatskar, and Farhadi]{robothor}
Matt Deitke, Winson Han, Alvaro Herrasti, Aniruddha Kembhavi, Eric Kolve, Roozbeh Mottaghi, Jordi Salvador, Dustin Schwenk, Eli VanderBilt, Matthew Wallingford, Luca Weihs, Mark Yatskar, and Ali Farhadi.
\newblock {RoboTHOR}: An open simulation-to-real embodied {AI} platform.
\newblock In \emph{Proceedings of the IEEE/CVF Conference on Computer Vision and Pattern Recognition}, 2020.

\bibitem[Eslami et~al.(2018)Eslami, Rezende, Besse, Viola, Morcos, Garnelo, Ruderman, Rusu, Danihelka, Gregor, et~al.]{Eslami2018}
S.~M.~Ali Eslami, Danilo~Jimenez Rezende, Frederic Besse, Fabio Viola, Ari~S. Morcos, Marta Garnelo, Avraham Ruderman, Andrei~A. Rusu, Ivo Danihelka, Karol Gregor, et~al.
\newblock Neural scene representation and rendering.
\newblock \emph{Science}, 360\penalty0 (6394), 2018.

\bibitem[Gervet et~al.(2023)Gervet, Chintala, Batra, Malik, and Chaplot]{nav_real}
Theophile Gervet, Soumith Chintala, Dhruv Batra, Jitendra Malik, and Devendra~Singh Chaplot.
\newblock Navigating to objects in the real world.
\newblock \emph{Science Robotics}, 8\penalty0 (79), 2023.

\bibitem[Gonzalez and Woods(2008)]{gonzalez2008digital}
Rafael~C. Gonzalez and Richard~E. Woods.
\newblock Intensity transformations and spatial filtering.
\newblock In \emph{Digital Image Processing}, chapter~3. Pearson Education, 3rd edition, 2008.

\bibitem[(Google~Research)(2024)]{gemini}
Gemini~Team (Google~Research).
\newblock Gemini: A family of highly capable multimodal models.
\newblock \emph{arXiv preprint arXiv:2312.11805}, 2024.

\bibitem[Hartley and Zisserman(2004)]{hartley2003multiple}
Richard Hartley and Andrew Zisserman.
\newblock N-view computational methods.
\newblock In \emph{Multiple View Geometry in Computer Vision}. Cambridge {U}niversity {P}ress, 2004.

\bibitem[Ho and Salimans(2021)]{ho2022classifier}
Jonathan Ho and Tim Salimans.
\newblock Classifier-free diffusion guidance.
\newblock In \emph{NeurIPS 2021 Workshop on Deep Generative Models and Downstream Applications}, 2021.

\bibitem[Ho et~al.(2020)Ho, Jain, and Abbeel]{ho2020denoising}
Jonathan Ho, Ajay Jain, and Pieter Abbeel.
\newblock Denoising diffusion probabilistic models.
\newblock In \emph{Proceedings of the Conference on Neural Information Processing Systems}, 2020.

\bibitem[Ichter et~al.(2022)Ichter, Brohan, Chebotar, Finn, Hausman, Herzog, Ho, Ibarz, Irpan, Jang, Julian, Kalashnikov, Levine, Lu, Parada, Rao, Sermanet, Toshev, Vanhoucke, Xia, Xiao, Xu, Yan, Brown, Ahn, Cortes, Sievers, Tan, Xu, Reyes, Rettinghouse, Quiambao, Pastor, Luu, Lee, Kuang, Jesmonth, Jeffrey, Ruano, Hsu, Gopalakrishnan, David, Zeng, and Fu]{saycan}
Brian Ichter, Anthony Brohan, Yevgen Chebotar, Chelsea Finn, Karol Hausman, Alexander Herzog, Daniel Ho, Julian Ibarz, Alex Irpan, Eric Jang, Ryan Julian, Dmitry Kalashnikov, Sergey Levine, Yao Lu, Carolina Parada, Kanishka Rao, Pierre Sermanet, Alexander~T Toshev, Vincent Vanhoucke, Fei Xia, Ted Xiao, Peng Xu, Mengyuan Yan, Noah Brown, Michael Ahn, Omar Cortes, Nicolas Sievers, Clayton Tan, Sichun Xu, Diego Reyes, Jarek Rettinghouse, Jornell Quiambao, Peter Pastor, Linda Luu, Kuang-Huei Lee, Yuheng Kuang, Sally Jesmonth, Kyle Jeffrey, Rosario~Jauregui Ruano, Jasmine Hsu, Keerthana Gopalakrishnan, Byron David, Andy Zeng, and Chuyuan~Kelly Fu.
\newblock {Do As I Can, Not As I Say}: Grounding language in robotic affordances.
\newblock In \emph{Proceedings of the Annual Conference on Robot Learning}, 2022.

\bibitem[Jayaraman and Grauman(2018)]{explore}
Dinesh Jayaraman and Kristen Grauman.
\newblock Learning to look around: Intelligently exploring unseen environments for unknown tasks.
\newblock In \emph{Proceedings of the IEEE/CVF Conference on Computer Vision and Pattern Recognition}, 2018.

\bibitem[Jia et~al.(2021)Jia, Yang, Xia, Chen, Parekh, Pham, Le, Sung, Li, and Duerig]{align}
Chao Jia, Yinfei Yang, Ye Xia, Yi-Ting Chen, Zarana Parekh, Hieu Pham, Quoc Le, Yun-Hsuan Sung, Zhen Li, and Tom Duerig.
\newblock Scaling up visual and vision-language representation learning with noisy text supervision.
\newblock In \emph{Proceedings of the 38th International Conference on Machine Learning}, 2021.

\bibitem[Jocher et~al.(2023)Jocher, Qiu, and Chaurasia]{ultralytics2023yolov8}
Glenn Jocher, Jing Qiu, and Ayush Chaurasia.
\newblock Ultralytics {YOLO}.
\newblock \url{https://ultralytics.com}, 2023.

\bibitem[Jutras-Dubé et~al.(2024)Jutras-Dubé, Zhang, and Bera]{jutrasdube2024adaptive}
Pascal Jutras-Dubé, Ruqi Zhang, and Aniket Bera.
\newblock Adaptive planning with generative models under uncertainty.
\newblock In \emph{Proceedings of the International Conference on Intelligent Robots and Systems}, 2024.

\bibitem[Katyal et~al.(2019)Katyal, Popek, Paxton, Burlina, and Hager]{unaware}
Kapil Katyal, Katie Popek, Chris Paxton, Phil Burlina, and Gregory~D. Hager.
\newblock Uncertainty-aware occupancy map prediction using generative networks for robot navigation.
\newblock In \emph{Proceedings of the International Conference on Robotics and Automation}, 2019.

\bibitem[Kim et~al.(2022)Kim, Kwon, and Ye]{kim2022diffusionclip}
Gwanghyun Kim, Taesung Kwon, and Jong~Chul Ye.
\newblock Diffusionclip: Text-guided diffusion models for robust image manipulation.
\newblock In \emph{Proceedings of the IEEE/CVF Conference on Computer Vision and Pattern Recognition}, 2022.

\bibitem[Kingma et~al.(2021)Kingma, Salimans, Poole, and Ho]{kingma2021variational}
Diederik~P. Kingma, Tim Salimans, Ben Poole, and Jonathan Ho.
\newblock Variational diffusion models.
\newblock In \emph{Proceedings of the Conference on Neural Information Processing Systems}, 2021.

\bibitem[Kurniawati(2022)]{kurniawati2022partially}
Hanna Kurniawati.
\newblock Partially observable markov decision processes and robotics.
\newblock \emph{Annual Review of Control, Robotics, and Autonomous Systems}, 5\penalty0 (1), 2022.

\bibitem[Li and Malik(2016)]{li2016}
Ke Li and Jitendra Malik.
\newblock Amodal instance segmentation.
\newblock In \emph{Proceedings of the European Conference on Computer Vision}, 2016.

\bibitem[Li et~al.(2022)Li, Wang, Snavely, and Kanazawa]{li2022infinitenature}
Zhengqi Li, Qianqian Wang, Noah Snavely, and Angjoo Kanazawa.
\newblock {InfiniteNature-Zero}: Learning perpetual view generation of natural scenes from single images.
\newblock In \emph{Proceedings of the European Conference on Computer Vision}. Springer, 2022.

\bibitem[Lin et~al.(2014)Lin, Maire, Belongie, Hays, Perona, Ramanan, Doll{\'a}r, and Zitnick]{lin2014microsoft}
Tsung-Yi Lin, Michael Maire, Serge Belongie, James Hays, Pietro Perona, Deva Ramanan, Piotr Doll{\'a}r, and C~Lawrence Zitnick.
\newblock Microsoft {COCO}: Common objects in context.
\newblock In \emph{Proceedings of the European Conference on Computer Vision}, 2014.

\bibitem[Ling et~al.(2020)Ling, Acuna, Kreis, Kim, and Fidler]{ling2020variational}
Huan Ling, David Acuna, Karsten Kreis, Seung~Wook Kim, and Sanja Fidler.
\newblock Variational amodal object completion.
\newblock In \emph{Proceedings of the Conference on Neural Information Processing Systems}, 2020.

\bibitem[Ling et~al.(2024)Ling, Sheng, Tu, Zhao, Xin, Wan, Yu, Guo, Yu, Lu, et~al.]{dl3dv}
Lu Ling, Yichen Sheng, Zhi Tu, Wentian Zhao, Cheng Xin, Kun Wan, Lantao Yu, Qianyu Guo, Zixun Yu, Yawen Lu, et~al.
\newblock {Dl3DV-10k}: A large-scale scene dataset for deep learning-based 3{D} vision.
\newblock In \emph{Proceedings of the IEEE/CVF Conference on Computer Vision and Pattern Recognition}, 2024.

\bibitem[Liu et~al.(2023)Liu, Li, Wu, and Lee]{llava-1}
Haotian Liu, Chunyuan Li, Qingyang Wu, and Yong~Jae Lee.
\newblock Visual instruction tuning.
\newblock In \emph{Proceedings of the Conference on Neural Information Processing Systems}, 2023.

\bibitem[Liu et~al.(2024)Liu, Li, Li, and Lee]{llava-2}
Haotian Liu, Chunyuan Li, Yuheng Li, and Yong~Jae Lee.
\newblock Improved baselines with visual instruction tuning.
\newblock In \emph{Proceedings of the IEEE/CVF Conference on Computer Vision and Pattern Recognition}, 2024.

\bibitem[Lowe(1999)]{sift}
David~G Lowe.
\newblock Object recognition from local scale-invariant features.
\newblock In \emph{Proceedings of the IEEE/CVF International Conference on Computer Vision}, 1999.

\bibitem[Ma et~al.(2023)Ma, Hong, Gul, Gandhi, Gao, and Krishna]{mllms}
Zixian Ma, Jerry Hong, Mustafa~Omer Gul, Mona Gandhi, Irena Gao, and Ranjay Krishna.
\newblock @ {CREPE}: Can vision-language foundation models reason compositionally?
\newblock In \emph{Proceedings of the IEEE/CVF Conference on Computer Vision and Pattern Recognition}, 2023.

\bibitem[Mildenhall et~al.(2020)Mildenhall, Srinivasan, Tancik, Barron, Ramamoorthi, and Ng]{mildenhall2021nerf}
Ben Mildenhall, Pratul~P. Srinivasan, Matthew Tancik, Jonathan~T. Barron, Ravi Ramamoorthi, and Ren Ng.
\newblock {NeRF}: Representing scenes as neural radiance fields for view synthesis.
\newblock In \emph{Proceedings of the European Conference on Computer Vision}, 2020.

\bibitem[Mur-Artal and Tardós(2017)]{mur2017orb}
Raúl Mur-Artal and Juan~D Tardós.
\newblock {ORB-SLAM2}: An open-source {SLAM} system for monocular, stereo, and {RGB-D} cameras.
\newblock In \emph{IEEE Transactions on Robotics}. IEEE, 2017.

\bibitem[Mur-Artal et~al.(2015)Mur-Artal, Montiel, and Tardós]{mur2015orb}
Raúl Mur-Artal, Jose Maria~Martinez Montiel, and Juan~D Tardós.
\newblock {ORB-SLAM}: A versatile and accurate monocular {SLAM} system.
\newblock \emph{IEEE Transactions on Robotics}, 31\penalty0 (5), 2015.

\bibitem[Müller et~al.(2024)Müller, Schwarz, Roessle, Porzi, Bulò, Nießner, and Kontschieder]{multidiff}
Norman Müller, Katja Schwarz, Barbara Roessle, Lorenzo Porzi, Samuel~Rota Bulò, Matthias Nießner, and Peter Kontschieder.
\newblock {MultiDiff}: Consistent novel view synthesis from a single image.
\newblock In \emph{Proceedings of the IEEE/CVF Conference on Computer Vision and Pattern Recognition}, 2024.

\bibitem[Nagami and Schwager(2024)]{10496149}
Keiko Nagami and Mac Schwager.
\newblock State estimation and belief space planning under epistemic uncertainty for learning-based perception systems.
\newblock \emph{IEEE Robotics and Automation Letters}, 9\penalty0 (6), 2024.

\bibitem[OpenAI(2023)]{chatgpt4}
OpenAI.
\newblock {GPT-4} {T}echnical report.
\newblock \emph{arXiv preprint arXiv:2303.08774}, 2023.

\bibitem[Ota et~al.(2021)Ota, Jha, Onishi, Kanezaki, Yoshiyasu, Sasaki, Mariyama, and Nikovski]{react}
Kei Ota, Devesh Jha, Tadashi Onishi, Asako Kanezaki, Yusuke Yoshiyasu, Yoko Sasaki, Toshisada Mariyama, and Daniel Nikovski.
\newblock Deep reactive planning in dynamic environments.
\newblock In \emph{Proceedings of the Conference on Robot Learning}, 2021.

\bibitem[Pan et~al.(2024)Pan, Barath, Pollefeys, and Sch\"{o}nberger]{glomap}
Linfei Pan, Daniel Barath, Marc Pollefeys, and Johannes~Lutz Sch\"{o}nberger.
\newblock Global structure-from-motion revisited.
\newblock In \emph{Proceedings of the European Conference on Computer Vision}, 2024.

\bibitem[Peebles and Xie(2023)]{DIT}
William Peebles and Saining Xie.
\newblock Scalable diffusion models with transformers.
\newblock In \emph{Proceedings of the IEEE/CVF International Conference on Computer Vision}, 2023.

\bibitem[Pertuz et~al.(2013)Pertuz, Puig, and Garcia]{Pertuz2013}
Said Pertuz, Domenec Puig, and Miguel~Angel Garcia.
\newblock Analysis of focus measure operators for shape-from-focus.
\newblock \emph{Pattern Recognition}, 46\penalty0 (5):\penalty0 1415--1432, 2013.

\bibitem[Podell et~al.(2024)Podell, English, Lacey, Blattmann, Dockhorn, M{\"u}ller, Penna, and Rombach]{podell2024sdxl}
Dustin Podell, Zion English, Kyle Lacey, Andreas Blattmann, Tim Dockhorn, Jonas M{\"u}ller, Joe Penna, and Robin Rombach.
\newblock {SDXL:} {I}mproving latent diffusion models for high-resolution image synthesis.
\newblock In \emph{Proceedings of the International Conference on Learning Representations}, 2024.

\bibitem[Radford et~al.(2021)Radford, Kim, Hallacy, Ramesh, Goh, Agarwal, Sastry, Askell, Mishkin, Clark, Krueger, and Sutskever]{CLIP}
Alec Radford, Jong~Wook Kim, Chris Hallacy, Aditya Ramesh, Gabriel Goh, Sandhini Agarwal, Girish Sastry, Amanda Askell, Pamela Mishkin, Jack Clark, Gretchen Krueger, and Ilya Sutskever.
\newblock Learning transferable visual models from natural language supervision.
\newblock In \emph{Proceedings of the International Conference on Machine Learning}, pages 8748--8763. PMLR, 2021.

\bibitem[Redmon et~al.(2016)Redmon, Divvala, Girshick, and Farhadi]{redmon2016yolo}
Joseph Redmon, Santosh Divvala, Ross Girshick, and Ali Farhadi.
\newblock {You Only Look Once}: Unified, real-time object detection.
\newblock In \emph{Proceedings of the IEEE/CVF Conference on Computer Vision and Pattern Recognition}, 2016.

\bibitem[Ren et~al.(2016)Ren, He, Girshick, and Sun]{ren2016faster}
Shaoqing Ren, Kaiming He, Ross Girshick, and Jian Sun.
\newblock {Faster R-CNN}: Towards real-time object detection with region proposal networks.
\newblock \emph{IEEE Transactions on Pattern Analysis and Machine Intelligence}, 39\penalty0 (6), 2016.

\bibitem[Ren and Wang(2022)]{lotr}
Xuanchi Ren and Xiaolong Wang.
\newblock Look outside the room: Synthesizing a consistent long-term {3D} scene video from a single image.
\newblock In \emph{Proceedings of the IEEE/CVF Conference on Computer Vision and Pattern Recognition}, 2022.

\bibitem[Rockwell et~al.(2021)Rockwell, Fouhey, and Johnson]{rockwell2021pixelsynth}
Chris Rockwell, David~F Fouhey, and Justin Johnson.
\newblock Pixelsynth: Generating a {{3D}}-consistent experience from a single image.
\newblock In \emph{Proceedings of the IEEE/CVF International Conference on Computer Vision}, 2021.

\bibitem[Rombach et~al.(2022)Rombach, Blattmann, Lorenz, Esser, and Ommer]{rombach2022high}
Robin Rombach, Andreas Blattmann, Dominik Lorenz, Patrick Esser, and Bj{\"o}rn Ommer.
\newblock High-resolution image synthesis with latent diffusion models.
\newblock In \emph{Proceedings of the IEEE/CVF Conference on Computer Vision and Pattern Recognition}, 2022.

\bibitem[Saharia et~al.(2022{\natexlab{a}})Saharia, Chan, Chang, Lee, Ho, Salimans, Fleet, and Norouzi]{saharia2022palette}
Chitwan Saharia, William Chan, Huiwen Chang, Chris Lee, Jonathan Ho, Tim Salimans, David Fleet, and Mohammad Norouzi.
\newblock Palette: Image-to-image diffusion models.
\newblock \emph{ACM Transactions on Graphics}, 2022{\natexlab{a}}.

\bibitem[Saharia et~al.(2022{\natexlab{b}})Saharia, Chan, Saxena, Li, Whang, Denton, Ghasemipour, Gontijo-Lopes, Ayan, Salimans, Ho, Fleet, and Norouzi]{saharia2022photorealistic}
Chitwan Saharia, William Chan, Saurabh Saxena, Lala Li, Jay Whang, Emily Denton, Seyed Kamyar~Seyed Ghasemipour, Raphael Gontijo-Lopes, Burcu~Karagol Ayan, Tim Salimans, Jonathan Ho, David~J. Fleet, and Mohammad Norouzi.
\newblock Photorealistic text-to-image diffusion models with deep language understanding.
\newblock In \emph{Proceedings of the Conference on Neural Information Processing Systems}, 2022{\natexlab{b}}.

\bibitem[Sargent et~al.(2024)Sargent, Li, Shah, Herrmann, Yu, Zhang, Chan, Lagun, Fei-Fei, Sun, and Wu]{zeronvs}
Kyle Sargent, Zizhang Li, Tanmay Shah, Charles Herrmann, Hong-Xing Yu, Yunzhi Zhang, Eric~Ryan Chan, Dmitry Lagun, Li Fei-Fei, Deqing Sun, and Jiajun Wu.
\newblock {ZeroNVS}: Zero-shot 360-degree view synthesis from a single image.
\newblock In \emph{Proceedings of the IEEE/CVF Conference on Computer Vision and Pattern Recognition}, 2024.

\bibitem[Sch\"{o}nberger et~al.(2016)Sch\"{o}nberger, Zheng, Pollefeys, and Frahm]{colmap}
Johannes~Lutz Sch\"{o}nberger, Enliang Zheng, Marc Pollefeys, and Jan-Michael Frahm.
\newblock Pixelwise view selection for unstructured multi-view stereo.
\newblock In \emph{Proceedings of the European Conference on Computer Vision}, 2016.

\bibitem[Schönberger and Frahm(2016)]{SFM}
Johannes~Lutz Schönberger and Jan-Michael Frahm.
\newblock Structure-from-motion revisited.
\newblock In \emph{Proceedings of the IEEE/CVF Conference on Computer Vision and Pattern Recognition}, 2016.

\bibitem[Shah et~al.(2023)Shah, Sridhar, Dashora, Stachowicz, Black, Hirose, and Levine]{shah2023vint}
Dhruv Shah, Ajay Sridhar, Nitish Dashora, Kyle Stachowicz, Kevin Black, Noriaki Hirose, and Sergey Levine.
\newblock Vi{NT}: A foundation model for visual navigation.
\newblock In \emph{Proceedings of the Annual Conference on Robot Learning}, 2023.

\bibitem[Shannon(1948)]{shannon1948}
Claude~E. Shannon.
\newblock A mathematical theory of communication.
\newblock \emph{Bell System Technical Journal}, 27\penalty0 (3), 1948.

\bibitem[Silberman et~al.(2012)Silberman, Hoiem, Kohli, and Fergus]{NYU}
Nathan Silberman, Derek Hoiem, Pushmeet Kohli, and Rob Fergus.
\newblock Indoor segmentation and support inference from {RGB-D} images.
\newblock In \emph{Proceedings of the European Conference on Computer Vision}. Springer, 2012.

\bibitem[Sivic et~al.(2007)Sivic, Isard, Zisserman, Philbin, and Chum]{sequential}
Josef Sivic, Michael Isard, Andrew Zisserman, James Philbin, and Ondrej Chum.
\newblock Object retrieval with large vocabularies and fast spatial matching.
\newblock In \emph{Proceedings of the IEEE/CVF Conference on Computer Vision and Pattern Recognition}, 2007.

\bibitem[Snavely et~al.(2006)Snavely, Seitz, and Szeliski]{snavely2006photo}
Noah Snavely, Steven~M Seitz, and Richard Szeliski.
\newblock Photo tourism: Exploring photo collections in {3D}.
\newblock In \emph{ACM Transactions on Graphics}, 2006.

\bibitem[Sohl-Dickstein et~al.(2015)Sohl-Dickstein, Weiss, Maheswaranathan, and Ganguli]{sohl2015deep}
Jascha Sohl-Dickstein, Eric Weiss, Niru Maheswaranathan, and Surya Ganguli.
\newblock Deep unsupervised learning using nonequilibrium thermodynamics.
\newblock In \emph{Proceedings of the International Conference on Machine Learning}, 2015.

\bibitem[Song et~al.(2021)Song, Meng, and Ermon]{DDIM}
Jiaming Song, Chenlin Meng, and Stefano Ermon.
\newblock Denoising diffusion implicit models.
\newblock In \emph{Proceedings of the International Conference on Learning Representations}, 2021.

\bibitem[Song and Ermon(2021)]{song2021scorebased}
Yang Song and Stefan Ermon.
\newblock Score-based generative modeling through stochastic differential equations.
\newblock In \emph{Proceedings of the International Conference on Learning Representations}, 2021.

\bibitem[Sun et~al.(2022)Sun, Kortylewski, and Yuille]{bayesian}
Yihong Sun, Adam Kortylewski, and Alan Yuille.
\newblock Amodal segmentation through out-of-task and out-of-distribution generalization with a {B}ayesian model.
\newblock In \emph{Proceedings of the IEEE/CVF Conference on Computer Vision and Pattern Recognition}, 2022.

\bibitem[Tang et~al.(2024)Tang, Nie, Markhasin, Dai, Thies, and Niessner]{scene_completion}
Jiapeng Tang, Yinyu Nie, Lev Markhasin, Angela Dai, Justus Thies, and Matthias Niessner.
\newblock {DiffuScene}: Denoising diffusion models for generative indoor scene synthesis.
\newblock In \emph{Proceedings of the IEEE/CVF Conference on Computer Vision and Pattern Recognition}, 2024.

\bibitem[Tewari et~al.(2023)Tewari, Yin, Cazenavette, Rezchikov, Tenenbaum, Durand, Freeman, and Sitzmann]{fwd}
Ayush Tewari, Tianwei Yin, George Cazenavette, Semon Rezchikov, Joshua~B. Tenenbaum, Fredo Durand, William~T. Freeman, and Vincent Sitzmann.
\newblock Diffusion with forward models: Solving stochastic inverse problems without direct supervision.
\newblock In \emph{Proceedings of the Conference on Neural Information Processing Systems}, 2023.

\bibitem[Thrush et~al.(2022)Thrush, Jiang, Bartolo, Singh, Williams, Kiela, and Ross]{winoground}
Tristan Thrush, Ryan Jiang, Max Bartolo, Amanpreet Singh, Adina Williams, Douwe Kiela, and Candace Ross.
\newblock Winoground: Probing vision and language models for visio-linguistic compositionality.
\newblock In \emph{Proceedings of the IEEE/CVF Conference on Computer Vision and Pattern Recognition}, 2022.

\bibitem[von Platen et~al.(2022)von Platen, Patil, Lozhkov, Cuenca, Lambert, Rasul, Davaadorj, Nair, Paul, Berman, Xu, Liu, and Wolf]{diffusers}
Patrick von Platen, Suraj Patil, Anton Lozhkov, Pedro Cuenca, Nathan Lambert, Kashif Rasul, Mishig Davaadorj, Dhruv Nair, Sayak Paul, William Berman, Yiyi Xu, Steven Liu, and Thomas Wolf.
\newblock Diffusers: State-of-the-art diffusion models.
\newblock \url{https://github.com/huggingface/diffusers}, 2022.

\bibitem[Wang et~al.(2004)Wang, Bovik, Sheikh, and Simoncelli]{Wang2004}
Zhou Wang, Alan~C. Bovik, Hamid~R. Sheikh, and Eero~P. Simoncelli.
\newblock Image quality assessment: From error visibility to structural similarity.
\newblock \emph{IEEE Transactions on Image Processing}, 13\penalty0 (4), 2004.

\bibitem[Watson et~al.(2023)Watson, Chan, Martin-Brualla, Ho, Tagliasacchi, and Norouzi]{watson2022novel}
Daniel Watson, William Chan, Ricardo Martin-Brualla, Jonathan Ho, Andrea Tagliasacchi, and Mohammad Norouzi.
\newblock Novel view synthesis with diffusion models.
\newblock In \emph{Proceedings of the International Conference on Learning Representations}, 2023.

\bibitem[Wen et~al.(2023)Wen, Jayaraman, and Gao]{wen2023can}
Chuan Wen, Dinesh Jayaraman, and Yang Gao.
\newblock Can transformers capture spatial relations between objects?
\newblock In \emph{Proceedings of the International Conference on Learning Representations}, 2023.

\bibitem[Xu et~al.(2022)Xu, Jiang, Wang, Fan, Shi, and Wang]{xu2022sinnerf}
Dejia Xu, Yifan Jiang, Peihao Wang, Zhiwen Fan, Humphrey Shi, and Zhangyang Wang.
\newblock {SinNeRF}: Training neural radiance fields on complex scenes from a single image.
\newblock In \emph{Proceedings of the European Conference on Computer Vision}, 2022.

\bibitem[Yang et~al.(2024{\natexlab{a}})Yang, Kang, Huang, Xu, Feng, and Zhao]{depthanything}
Lihe Yang, Bingyi Kang, Zilong Huang, Xiaogang Xu, Jiashi Feng, and Hengshuang Zhao.
\newblock {Depth Anything}: Unleashing the power of large-scale unlabeled data.
\newblock In \emph{Proceedings of the IEEE/CVF Conference on Computer Vision and Pattern Recognition}, 2024{\natexlab{a}}.

\bibitem[Yang et~al.(2024{\natexlab{b}})Yang, Kang, Huang, Zhao, Xu, Feng, and Zhao]{depthanythingv2}
Lihe Yang, Bingyi Kang, Zilong Huang, Zhen Zhao, Xiaogang Xu, Jiashi Feng, and Hengshuang Zhao.
\newblock {Depth Anything V2}.
\newblock In \emph{Proceedings of the Conference on Neural Information Processing Systems}, 2024{\natexlab{b}}.

\bibitem[Yu et~al.(2021)Yu, Ye, Tancik, and Kanazawa]{PixelNeRF}
Alex Yu, Vickie Ye, Matthew Tancik, and Angjoo Kanazawa.
\newblock {pixelNeRF}: Neural radiance fields from one or few images.
\newblock In \emph{Proceedings of the IEEE/CVF Conference on Computer Vision and Pattern Recognition}, 2021.

\bibitem[Yu et~al.(2023)Yu, Forghani, Derpanis, and Brubaker]{photonvs}
Jason~J. Yu, Fereshteh Forghani, Konstantinos~G. Derpanis, and Marcus~A. Brubaker.
\newblock Long-term photometric consistent novel view synthesis with diffusion models.
\newblock In \emph{Proceedings of the IEEE/CVF International Conference on Computer Vision}, 2023.

\bibitem[Zeng et~al.(2021)Zeng, Röfer, and Jenkins]{chad}
Zhen Zeng, Adrian Röfer, and Odest~Chadwicke Jenkins.
\newblock Semantic linking maps for active visual object search.
\newblock In \emph{Proceedings of the International Joint Conference on Artificial Intelligence}, 2021.

\bibitem[Zhang et~al.(2024)Zhang, Huang, Jin, and Lu]{VLM_survey}
Jingyi Zhang, Jiaxing Huang, Sheng Jin, and Shijian Lu.
\newblock Vision-language models for vision tasks: A survey.
\newblock \emph{IEEE Transactions on Pattern Analysis and Machine Intelligence}, 2024.

\bibitem[Zhen et~al.(2024)Zhen, Qiu, Chen, Yang, Yan, Du, Hong, and Gan]{3dvla}
Haoyu Zhen, Xiaowen Qiu, Peihao Chen, Jincheng Yang, Xin Yan, Yilun Du, Yining Hong, and Chuang Gan.
\newblock {3D-VLA}: A {3D} vision-language-action generative world model.
\newblock In \emph{Proceedings of the International Conference on Machine Learning}, 2024.

\bibitem[Zhou et~al.(2018)Zhou, Tucker, Flynn, Fyffe, and Snavely]{stereomag}
Tinghui Zhou, Richard Tucker, John Flynn, Graham Fyffe, and Noah Snavely.
\newblock Stereo magnification: Learning view synthesis using multiplane images.
\newblock In \emph{Proceedings of ACM SIGGRAPH}, 2018.

\bibitem[Zhu et~al.(2017{\natexlab{a}})Zhu, Mottaghi, Kolve, Lim, Gupta, Fei-Fei, and Farhadi]{zhu2017target}
Yuke Zhu, Roozbeh Mottaghi, Eric Kolve, Joseph~J Lim, Abhinav Gupta, Li Fei-Fei, and Ali Farhadi.
\newblock Target-driven visual navigation in indoor scenes using deep reinforcement learning.
\newblock In \emph{Proceedings of the IEEE International Conference on Robotics and Automation}, 2017{\natexlab{a}}.

\bibitem[Zhu et~al.(2017{\natexlab{b}})Zhu, Tian, Metaxas, and Doll{\'a}r]{zhu2017semantic}
Yan Zhu, Yuandong Tian, Dimitris Metaxas, and Piotr Doll{\'a}r.
\newblock Semantic amodal segmentation.
\newblock In \emph{Proceedings of the IEEE/CVF Conference on Computer Vision and Pattern Recognition}, 2017{\natexlab{b}}.

\bibitem[Zou et~al.(2023)Zou, Chen, Shi, Guo, and Ye]{zou2023objectdetection20years}
Zhengxia Zou, Keyan Chen, Zhenwei Shi, Yuhong Guo, and Jieping Ye.
\newblock Object detection in 20 years: A survey.
\newblock \emph{Proceedings of the IEEE}, 2023.

\end{thebibliography}
}

\appendix
\clearpage
\maketitlesupplementary
\section{Object, Scene and Parameter Selection Methodology}
\label{sec:appendix_scene}
In this Appendix, we detail the methodologies for scene selection, the filtering process for scenes analyzed, the criteria for selecting images and objects, and the approach for parameter calibration.

\paragraph{\rk random scene selection methodology.} To systematically select diverse scenes, we sampled video frames in the \rk test dataset, which contains world-to-camera poses and pinhole camera intrinsics~\citep{stereomag}, built using the ORBSLAM pipeline (with an ambiguous scale)~\citep{mur2017orb, mur2015orb}. Due to memory limitations, we downloaded 250 random scenes for analysis. \rk contains indoor and outdoor scenes but lacks metadata or labels, thus to interpret and filter each scene, our objective was to select representative viewpoints in each scene. 
We selected these frames, for every scene, at approximately 
equidistant~\citep{carlone} intervals in $\mathrm{SE}(3)$, between the two farthest camera-to-world poses for each scene.
Using these selected images, per scene, we used the ChatGPT-4o API~\citep{chatgpt4} to classify scenes as indoor or outdoor based on these 10 images per scene. For the indoor scenes, we further utilized the API to categorize them into four types: kitchen, study, bedroom, and living room. This automated labeling ensured that our subsequent analysis remained focused and consistent across the relevant indoor scenes.

\paragraph{\nyu random scene selection methodology.}For the \nyu dataset~\citep{NYU}, which contains meta-labels, this process was not necessary. From the entire dataset, we selected 50 random scenes that met the following criteria: living rooms, bedrooms, kitchens, and studies. These categories ensured that the selected scenes closely matched the diversity found in \rk, making the study more homogeneous compared to other available meta-labels in the dataset.

\paragraph{Filtering and selection of randomly selected scenes.} We refined the selection of indoor scenes from both datasets based on the following criteria:  
\begin{itemize}[noitemsep]
    \item \textit{Scene Size}: Scenes containing fewer than 50 images were excluded to mitigate the risk of reconstruction sparsity.  
    \item \textit{Camera Movement}: To ensure sufficient variation in movement, a minimum threshold was applied to the geodesic distance between the two furthest camera poses, requiring it to exceed 0.01 units.  
    \item \textit{Semantic Content}: Scenes were filtered based on object detection using YOLOv8x~\citep{ultralytics2023yolov8, redmon2016yolo} on COCO object classes~\citep{lin2014microsoft}. Specifically, scenes were excluded if fewer than 50\% of frames contained objects detected with a confidence score $\geq 0.5$.  
\end{itemize}
Out of the multiple scenes that met these criteria in each dataset, we randomly selected 10 scenes from both datasets for our study. For the two datasets the selected scenes, target objects, and the input images are tabulated in \cref{tab:scenes}.

\begin{table*}
    \centering
    \caption{Scene selected from \rk~\citep{stereomag} and \nyu~\citep{NYU} datasets.}
    \renewcommand{\arraystretch}{1.2}
    \begin{tabularx}{\linewidth}{X X | X l}
        \toprule
        \multicolumn{2}{c}{\rk Dataset~\citep{stereomag}} & \multicolumn{2}{| c}{\nyu Dataset~\citep{NYU}} \\
        \cmidrule(lr){1-2} \cmidrule(lr){3-4}
        Scene ID & Image ID & Scene ID & Image ID \\
        \midrule
        2e4013ea92d04301 & 119586133   & Living Room 0004  & 1295148543.251260-1026494144 \\
        2bec33eeeab0bb9d & 34768067    & Kitchen 0040      & 1315269892.882236-1150326380 \\
        2e64a2d17f9a76f7 & 162629000   & Living Room 0016  & 1300200232.988284-1300278508 \\
        2b625e92f2cf9de4 & 51384667    & Living Room 0010  & 1295836465.564725-1670107084 \\
        2cb9869cb05a9a01 & 77786042    & Living Room 0002  & 1294890229.045795-2653268294 \\
        3c64a373bc1c53bd & 199767000   & Bedroom 0025      & 1315330245.479316-1684325155 \\
        ff6d8ab35e042db5 & 142142000   & Bedroom 0029      & 1315423943.586243-93796617   \\
        2bd7cee1fa9c8996 & 51133333    & Kitchen 0024      & 1315441158.531288-3169924603 \\
        3de41ace235a3a13 & 49616000    & Kitchen 0031      & 1315165725.285327-3895871610 \\
        2d6d5e82bda0611c & 153253000   & Study 0003        & 1300708629.505940-4057834691 \\
        \bottomrule
    \end{tabularx}
    \label{tab:scenes}
\end{table*}

\paragraph{End-to-end object selection pipeline.} We deployed an end-to-end automated object selection pipeline on the randomly acquired scenes, logging the occurrence frequency of detected objects across the subsets of both datasets. From the combined scenes of \nyu and \rk scenes, we selected the 20 most frequently detected object classes among the 80 classes in the COCO dataset~\citep{lin2014microsoft}. For our analysis, we focused on 10 random object categories from these 20 classes that were present in at least one of the chosen scenes in the combined choices of scenes from both datasets. The COCO object categories analyzed include: \textit{refrigerator, TV, bed, chair, sink, oven, book, laptop, couch, and door}. We used Yolov8x~\citep{redmon2016yolo, ultralytics2023yolov8} for its strong adaptability for images of various resolutions and superior generalizability. Note that all 2D/3D cases analyzed were manually inspected and verified against YOLOv8x annotations to ensure high quality ground truth annotations.

\paragraph{Context image-object pair selection for the study.} For each scene, we select a single image based on specific criteria to ensure the target object is effectively represented within the scene. The selection process involves two criteria. First, an image is selected if the target object is absent in the $360 \times 360$ center-crop but present in the $360 \times 640$ full image. Alternatively, an image is chosen if the target object is not visible in the $360 \times 640$ frame but appears in more than 50\% of the scene's frames at a confidence $\geq0.5$. This criterion accounts for objects that may be occluded or located entirely outside the camera's frustum while still present within the scene.
These criteria result in a large set of image-object pairs across scenes, meeting at least one of the specified conditions. For the \nyu dataset, the fraction of available cases is relatively higher, which is expected due to slower average sequential camera movements compared to the \rk dataset. From this filtered set, we select 10 random test image-object pairs per dataset.
To avoid cross-fading or sudden movement artifacts, images are not chosen from the first or last 10 timestamps, where such transitions were qualitatively observed in some camera trajectories.
For the \rk dataset, there are 31 scene-object pairs where the ground truth observation captures a relevant COCO object in both 2D and 2.5D representations, with 37 cases observed in 3D. In contrast, the \nyu dataset contains 19 scene-object pairs captured within the frustum in 2D and 2.5D, with 26 cases in 3D.
\begin{figure}
    \centering
    \includegraphics[width=1\columnwidth]{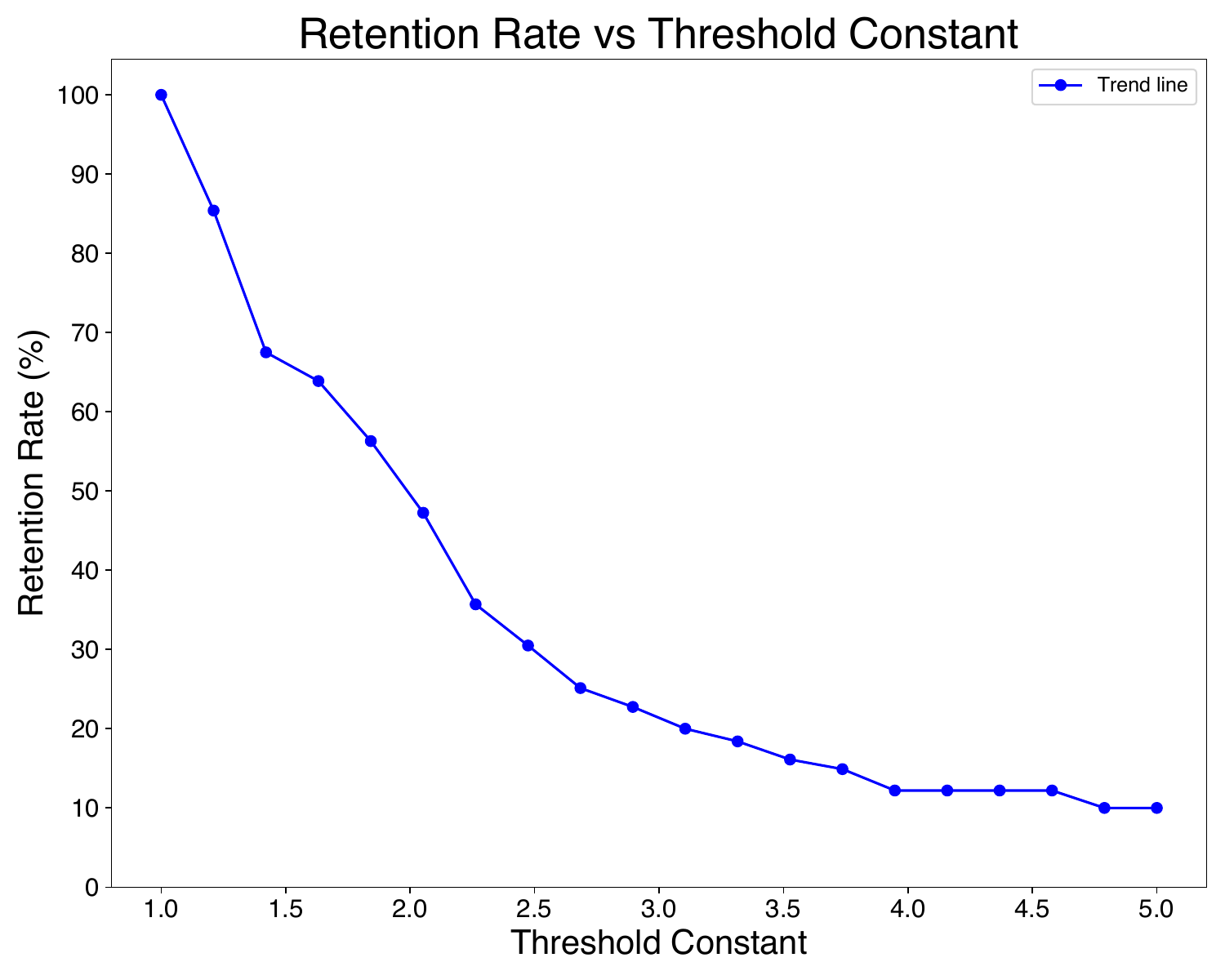}
    \caption{Threshold Multiplier $\lambda$ vs the Retention Rate of points averaged across all predicted grids in 2D, 2.5D and 3D.}
    \label{fig:multiplier}
\end{figure}

\begin{figure}
    \centering
    \includegraphics[width=1\columnwidth]{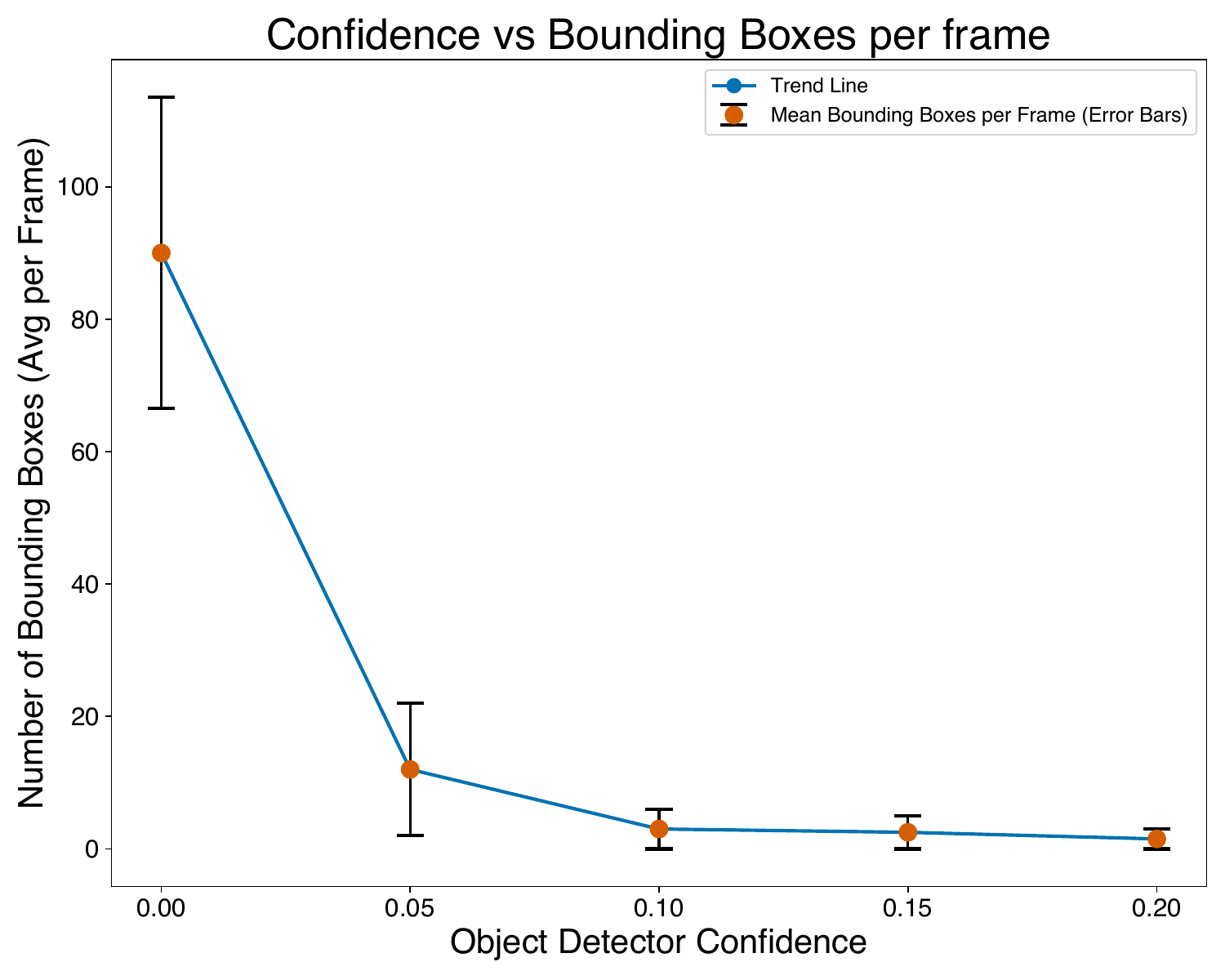}
    \caption{Calibration of Object Detector Confidence using Bounding Boxes per Frame.}
    \label{fig:conf-thresh}
\end{figure}

\paragraph{Parameter calibration methodology.} As shown in \cref{fig:conf-thresh}, we calibrated the detector confidence by analyzing the average bounding boxes per frame across 100 randomly selected videos from the indoor \rk test set out of the 250 downloaded videos. We utilize the same heuristics across both the datasets. A threshold of 0.1 achieves a balanced trade-off between sensitivity and precision with stable variance. All other parameters, followed the YOLOv8x default parameters~\citep{ultralytics2023yolov8}. 
For the threshold of the metric, we use a multiplier as $\tau\coloneqq\lambda/|\gspan| > 1/|\gspan|$. To calibrate this value, we use the average retention rates across all points in all predicted grids pre-normalization to post-normalization across the \rk dataset (2D and 3D inclusive) and plot it as a percentage. We see a breaking point at $\lambda \approx 1.4$, as shown in \cref{fig:multiplier}, which informs our choice of the heuristic.

\section{3D scene reconstruction details}
\label{sec:appendix_reconstruction}
In this appendix we outline the 3D scene reconstruction methodology for the two datasets.

\paragraph{\rk 3D reconstruction.} We used \textit{SIFT}~\citep{sift} to extract up to 50,000 features per image across multiple GPUs. These features were used for sequential matching~\citep{sequential}, followed by point triangulation using the camera parameters provided by the \rk dataset~\citep{stereomag}. This process yielded sufficient matches for patch-match stereo reconstruction\citep{colmap, SFM, hartley2003multiple}, though the resulting 3D model remained relatively sparse~\citep{stereomag}.

\paragraph{\nyu 3D reconstruction.} The \nyu dataset posed significant challenges for reconstruction due to the absence of ground-truth poses. Using COLMAP~\citep{colmap} for scene mapping was computationally intensive, so we opted for GLOMAP~\citep{glomap}, which significantly expedited the process. We calculated the camera intrinsics in normalized device coordinates (NDC), \ie, clip space, for each reconstructed scene, using the COLMAP pinhole camera model, with the principal points positioned at the image midpoints. Camera extrinsics obtained from the reconstruction were subsequently utilized for downstream tasks. Next, we undistorted the images and applied patch-match stereo reconstruction to generate a dense 3D reconstruction.

\paragraph{Post-reconstruction processing.} After dense reconstruction, statistical outlier filtering was applied to reduce noise and improve the quality of the reconstructed 3D point cloud. This process evaluates each point based on distances to its 20 nearest neighbors. 
Points deviating by more than twice the standard deviation from the mean distance are classified as outliers and are removed.
We applied a depth clipping threshold of 10.0 distance units with respect to the input camera pose.
Finally, each 3D reconstruction was transformed to align with the camera pose of a selected reference image in each scene. For every point cloud, given the bounding boxes, for the target objects, from the 2D images, we back-project to obtain the target 3D points with their confidences. 
We voxelize these points as required for the ground truth representation assigning the average non-zero confidence of the points within the voxel, to each voxel center. 
Details of the other COLMAP reconstruction parameters used will be released with the code.

\section{Additional implementation details}
\label{sec_additional_implementation_details}
In this appendix, we provide additional implementation details, especially on the sampling process and post-processing, filtering, pose selection process, and depth handling techniques employed in our study. 

\paragraph{System Requirements.} The experiments were conducted on a single machine equipped with an NVIDIA A6000 GPU, an AMD Ryzen Threadripper Pro 5995WX CPU, and 64GB of DDR5 RAM, running Ubuntu 22.04 LTS. From sampling to the final metrics summary, our end-to-end pipeline takes up to 38 minutes utilizing multiprocessing. 

\paragraph{Implementation details for $\dnn$.} To identify the distribution peaks or modes, we applied spatial filtering within the Moore neighborhood for each grid element \citep{gonzalez2008digital}, using constant zero padding at the edges to detect local maxima. Multiple kernel sizes (3, 5, 7, and 10) were used to ensure robust peak detection across varying scales. To compute the nearest distances between these detected peaks, we employed a \texttt{KDTree} data structure.

\paragraph{DFM sample detection implementation.} 
The outputs of the DFM~\citep{fwd} are of resolution $128 \times 128$. A key limitation of using an off-the-shelf detection model like Yolov8x is its reduced ability to detect all instances at lower resolutions than its training resolution. To address this, we resize the images from $128 \times 128$ to $512 \times 512$ for detection, compute the bounding boxes at the higher resolution, and then rescale the bounding box extents back to the original $128 \times 128$ resolution (to the nearest integer pixel value) for further processing.

\paragraph{DFM sampling and filtering details.} DFM~\citep{fwd} inference was performed using autoregressive sampling with 3 time-steps per sample, generating 30 intermediate frames, with a temperature of $0.85$ and a guidance scale of $2.5$.
To filter out intermediate frames potentially exhibiting noisy or featureless outputs due to unconditioned sampling, we implemented a three-phase filtering pipeline. 
First, noise was quantified using the Laplacian variance $V_L$, and samples were discarded if $V_L < 100$ in more than 80\% of the frames~\citep{Pertuz2013}. 
Next, Structural Similarity Index (SSIM) was computed between consecutive frames, with videos rejected if $\text{SSIM} > 0.9$ in over 50\% of the frames, indicating insufficient variation~\citep{Wang2004}.
Finally, per-pixel average intensity $L_i$ was tracked, and samples were discarded if the variation between consecutive frames remained below 50 units for over 50\% of the frames, indicating static content or poor lighting.
The inference time required for each sample per target camera pose constitutes a significant limitation of the DFM. Moreover, the filtration step, conducted after sample generation, often extended the overall process to several days to collect 75 samples per input image, with the duration varying between different input images. This computational overhead poses a significant constraint on scaling the sample size.

\begin{figure}[!htb]
    \centering
    \includegraphics[width = \columnwidth]{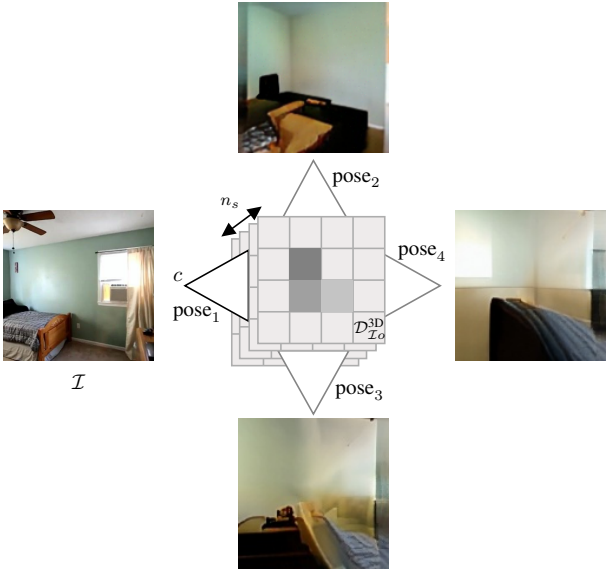}
    \caption{The 3D diffusion pipeline leverages four camera poses to generate up to $n_s$ samples, given the  input image $\inim$.} 
    \label{fig:dfm-poses}
\end{figure}

\paragraph{Choice of poses in $\mathrm{SE}(2)$ for DFM sampling.} To balance scene diversity and computational efficiency for the DFM experiments, three target poses were selected for each context image. These poses, as illustrated in \cref{fig:dfm-poses}, (Pose 1 being the input frame, \cref{fig:3D_pipeline}) were chosen based on their ability to cover the scene without blind spots, defined in terms of $(x,y,\theta)$ in \cref{tab:poses}.
\begin{table*}[!htbp]
\centering
\caption{Details of DFM test poses, their coordinates, and spatial coverage in the scene.}
\begin{tabular}{lcl}
\toprule
\textbf{Pose} & \textbf{Coordinates (x, y, $\theta$)} & \textbf{Description} \\ 
\midrule
\textbf{Pose 1} & $(0, 0, 0^\circ)$ & Input pose set at the origin relative to the scene movement. \\ 
\textbf{Pose 2} & $(-2, 2, 90^\circ)$ & Covers the western (right) boundary and central zones of the scene. \\ 
\textbf{Pose 3} & $(2, 2, -90^\circ)$ & Focuses on the eastern (left) boundary, ensuring coverage of the left and central areas. \\ 
\textbf{Pose 4} & $(0, 5, 180^\circ)$ & Captures depth and provides a longitudinal perspective from the south. \\ 
\bottomrule
\end{tabular}
\label{tab:poses}
\end{table*}
\begin{figure}
    \centering
    \begin{subfigure}{0.11\textwidth}
        \includegraphics[width=\linewidth]{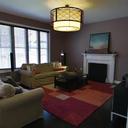}
    \end{subfigure}
    \hfill
    \begin{subfigure}{0.11\textwidth}
        \includegraphics[width=\linewidth]{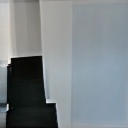}
    \end{subfigure}
    \hfill
    \begin{subfigure}{0.11\textwidth}
        \includegraphics[width=\linewidth]{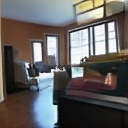}
    \end{subfigure}
    \hfill
    \begin{subfigure}{0.11\textwidth}
        \includegraphics[width=\linewidth]{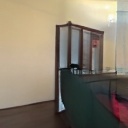}
    \end{subfigure}
    
    \vspace{1ex}
    \begin{subfigure}{0.11\textwidth}
        \includegraphics[width=\linewidth]{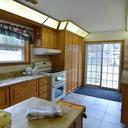}
    \end{subfigure}
    \hfill
    \begin{subfigure}{0.11\textwidth}
        \includegraphics[width=\linewidth]{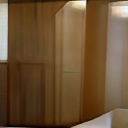}
    \end{subfigure}
    \hfill
    \begin{subfigure}{0.11\textwidth}
        \includegraphics[width=\linewidth]{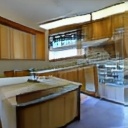}
    \end{subfigure}
    \hfill
    \begin{subfigure}{0.11\textwidth}
        \includegraphics[width=\linewidth]{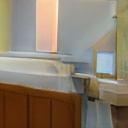}
    \end{subfigure}

    \vspace{1ex}
    \begin{subfigure}{0.11\textwidth}
        \includegraphics[width=\linewidth]{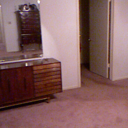}
    \end{subfigure}
    \hfill
    \begin{subfigure}{0.11\textwidth}
        \includegraphics[width=\linewidth]{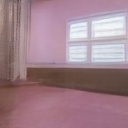}
    \end{subfigure}
    \hfill
    \begin{subfigure}{0.11\textwidth}
        \includegraphics[width=\linewidth]{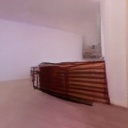}
    \end{subfigure}
    \hfill
    \begin{subfigure}{0.11\textwidth}
        \includegraphics[width=\linewidth]{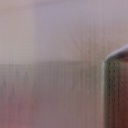}
    \end{subfigure}
    
    \vspace{1ex}
    \begin{subfigure}{0.11\textwidth}
        \includegraphics[width=\linewidth]{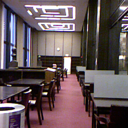}
        \caption{Pose 1}
    \end{subfigure}
    \hfill
    \begin{subfigure}{0.11\textwidth}
        \includegraphics[width=\linewidth]{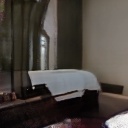}
        \caption{Pose 2}
    \end{subfigure}
    \hfill
    \begin{subfigure}{0.11\textwidth}
        \includegraphics[width=\linewidth]{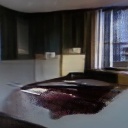}
        \caption{Pose 3}
    \end{subfigure}
    \hfill
    \begin{subfigure}{0.11\textwidth}
        \includegraphics[width=\linewidth]{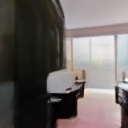}
        \caption{Pose 4}
    \end{subfigure}

    \caption{DFM samples images from four different camera poses, including the input pose, with a resolution of \(128 \times 128\) for each image. DFM~\citep{fwd} internally rescaled the original input image of \(360 \times 360\).}
    \label{fig:DFM_samples}
\end{figure}

\paragraph{Handling scene depth of the generated samples.} For the 3D study, we utilized the aggregated point cloud from the samples from the corresponding poses, and correspondingly all valid 3D points in the ground-truth reconstruction, including occluded parts of the scene, voxelized into a grid of size $20 \times 20 \times 20$, such that the camera position is at the center of the grid. For the SDXL-based study, taking a similar approach as \citet{spatialvlm} to lift our representation to 2.5D, we generated metric depths up to the far plane of $10$ units using DepthAnythingv2~\citep{depthanything, depthanythingv2}, selected for its superior cross-domain generalizability and compatibility with our known camera model. This approach provided depths within the frustum, ensuring scale consistency across the analyses, making sure that the camera centers are aligned to the voxel grid center for the 3D metric computations. For the \rk dataset experiments, we used the NDC camera intrinsics which is provided, and for the \nyu dataset experiments, we used the NDC camera intrinsics that are obtained for each scene, derived from the reconstruction process.
In our 2.5D study, ground truth representations and DFM-aggregated point clouds (with associated confidences) were culled within the camera's frustum determined by the camera's pose and intrinsic parameters. Occluded points were removed using Z-culling with a depth buffer, ensuring only the closest points to the camera's projection were retained~\footnote{\href{https://developer.nvidia.com/gpugems/gpugems/part-v-performance-and-practicalities/chapter-29-efficient-occlusion-culling}{Efficient occlusion culling methods}}. These points were scaled and voxelized into a grid of size $10 \times 10 \times 10$ from the 3D grid. 
The voxel grid was indexed with the input camera positioned at a 5-unit offset from the grid center along the $+z$-axis~\footnote{\href{https://docs.opencv.org/2.4/modules/calib3d/doc/camera_calibration_and_3d_reconstruction.html}{Camera Calibration and 3D reconstruction documentation}}. 
The projected pixels and the associated confidences were used for 2D analysis, while the corresponding 3D points were incorporated into the 2.5D study for voxelization. Pixels without any projections are replaced with zero confidence.
\begin{table*}[ht!]
\centering
\caption{VLM Prompts for Region-Wise Queries}
\begin{tabularx}{\textwidth}{l X}
\toprule
\textbf{Region} & \textbf{Prompt} \\
\midrule
\textbf{Left}   & \say{If the image frame is extended by \textbf{140 pixels} to the \textbf{left}, is it likely that there would be a/an \textbf{[OBJECT]} there? \textbf{Answer strictly in: Yes/No.}} \\
\midrule
\textbf{Right}  & \say{If the image frame is extended by \textbf{140 pixels} to the \textbf{right}, is it likely that there would be a/an \textbf{[OBJECT]} there? \textbf{Answer strictly in: Yes/No.}} \\
\midrule
\textbf{Central} & \say{Is it likely that there is a/an \textbf{[OBJECT]} within the frame of this image? \textbf{Answer strictly in: Yes/No.}} \\
\bottomrule
\end{tabularx}
\label{tab:VLM_prompts}
\end{table*}

\paragraph{SDXL generation details.} 

For our experiments, we used the publicly available Stable Diffusion XL Inpainting pipeline from the Hugging Face Diffusers library~\citep{diffusers}. Future changes to APIs and access to this model may affect reproducibility.
During the 2D generation process with SDXL, we used a fixed set of seeds for all experiments. The input image is placed at the center of a $360 \times 640$ canvas with 140 pixels of masking on each side of the center $360 \times 360$ crop, by the image dimension of $360 \times 140$. An additional filter was applied to remove cases where \textit{person(s)} were detected with a confidence score of $\geq 0.5$.
Both the image and mask are resized to $512\times 512$, and was resized back to $360 \times 640$, to match the target dimension. In certain cases, the combination of seed and prompt led to the generation of NSFW content, which the model automatically rejected, returning a blank image for the outpainted regions. These samples were excluded from our analysis. To ensure fairness in aggregation, we resampled till we hit the target number for all sets in every prompt regime. We used 10 different prompts in three different regimes, (1) with object cues in text prompts, (2) without object cues in text prompts and (3) no text prompts. The prompts used were:

\begin{enumerate}
\item Extend this indoor scene naturally to the left and right of the given frame (with objects like [OBJECT]).
\item Extend this indoor scene on both sides (with the object: [OBJECT]).
\item Outpaint 140 pixels to the left and right of the room ensuring continuity (using the object: [OBJECT]).
\item Extend the indoor scene horizontally to reveal more of the indoor scene (including a/an [OBJECT]).
\item Add 140 pixels to both sides of the frame to expand the indoor scene (incorporating a/an [OBJECT]).
\item Widen the indoor scene on the left and right (ensuring the object: [OBJECT]).
\item Outpaint 140 pixels to the left and right of the given image horizontally to create a larger view of the indoor scene (featuring the object: [OBJECT]).
\item Expand the boundaries of the frame to the left and right (adding a/an [OBJECT]) to maintain continuity of this indoor scene image.
\item Extend this scene on both sides to display a broader perspective of the indoor scene (including a/an [OBJECT]).
\item Add to the left and right of the image, showcasing more of the indoor setting (with a/an [OBJECT]).
\end{enumerate}

Parentheses indicate object cues, which are included in prompts with cues and omitted in prompts without cues. In the no-text-prompt regime, the prompt field is left empty. To ensure consistency in sampling for the ablation study, we generate 200 such prompts in total.
\begin{figure*}[ht]
    \centering
    \rotatebox[origin=l]{90}{\footnotesize CAR}
    \begin{subfigure}[]{0.27\linewidth}
        \centering
        \begin{tikzpicture}
            \node [anchor=south west,inner sep=0] at (0,0) {\includegraphics[width=\linewidth]{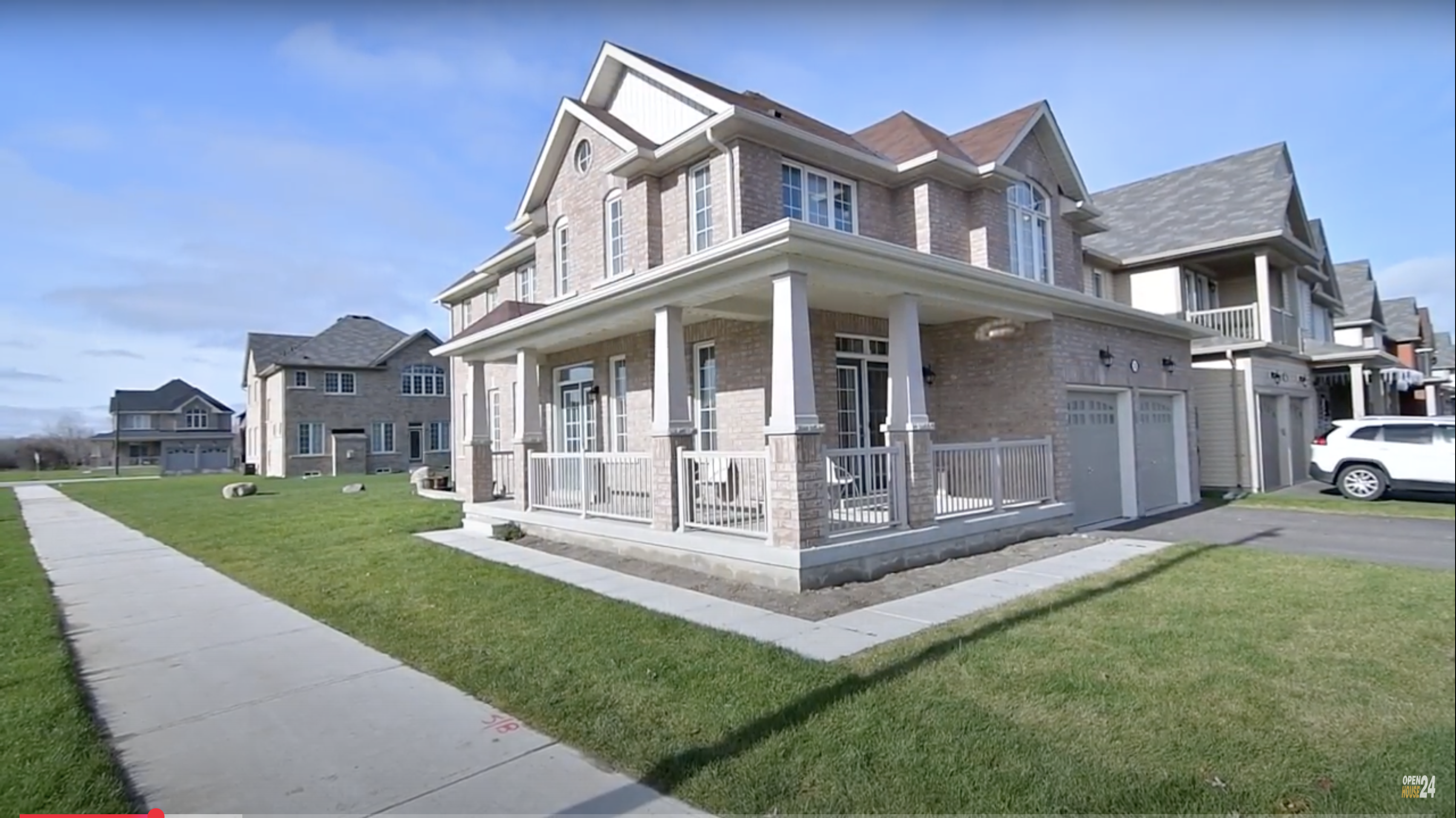}};
            \draw[white, thick, dashed] (0.22*\linewidth, 0.01) -- (0.22*\linewidth, 0.5648*\linewidth);
            \draw[white, thick, dashed] (0.78*\linewidth, 0.01) -- (0.78*\linewidth, 0.5648*\linewidth);
        \end{tikzpicture}
        \caption{$\inim$ (center) and $\outim$ (full)}
    \end{subfigure}
    \begin{subfigure}[]{0.15\linewidth}
        \centering
        \includegraphics[width=\linewidth]{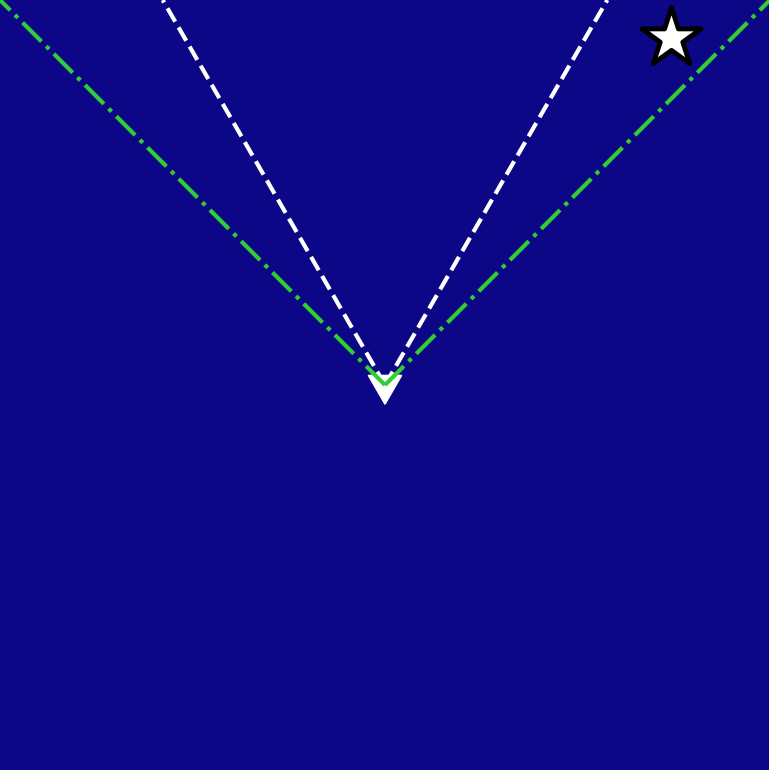}
        \caption{DFM-based $\sd^\text{3D}$}
    \end{subfigure}
    \begin{subfigure}[]{0.27\linewidth}
        \centering
        \includegraphics[width=\linewidth]{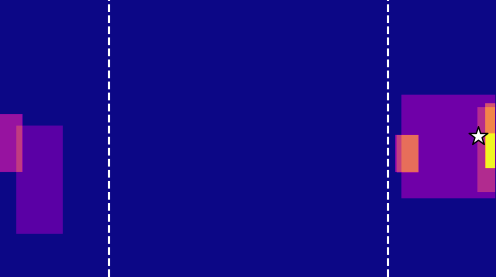}
        \caption{SDXL-based $\sd^\text{2D}$}
    \end{subfigure}
    \begin{subfigure}[]{0.27\linewidth}
        \centering
        \includegraphics[width=\linewidth]{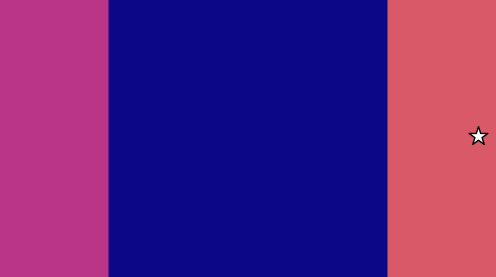}
        \caption{ChatGPT-4o-based $\sd^\text{2D}$}
    \end{subfigure}
    {\tt 0}\ 
    \includegraphics[width=0.5\linewidth, trim={12pt 8pt 10pt 8pt}, clip]{images/plasma_scale.pdf}\ {\tt 1}
    \caption{Results for an outdoor scene--object pair. The GT position is denoted by a white star.}
    \label{fig:outdoor}
    \vspace{-1em}
\end{figure*}
\section{Detailed methodology for VLM sampling}
\label{sec:appendix_vlm}
This appendix  provides additional details on how we implemented VLM query structuring and response processing for the study. 

\paragraph{Query format for VLMs.} To reduce the likelihood of incoherent outputs caused by uncontrolled generation for larger tokens, we restricted responses to binary \say{Yes} or \say{No}, limiting all answers to a single token. This simplification mitigated errors stemming from the unpredictability of free-form responses, especially in tasks requiring spatial reasoning. 
Trials with spatial queries showed significant variability and low accuracy against ground truth, largely due to per-voxel granularity required for 2.5D and 3D studies. Hence, we defer these experiments to future work. We standardized queries to focus on object presence or relative positions in 2D, tabulated in \cref{tab:VLM_prompts}.

\paragraph{Answer retrieval pipeline.} To generate confidence values for each region-object pair, we calculated a granular score for each region (left, right, and within the frame) based on the fraction of \say{Yes} answers provided by the model, for that region. 
These scores were softmax normalized across each pixel to ensure consistency, with their sum equal to 1, and treated as model's normalized confidence.
The normalized scores formed the basis of the 2D spatio-semantic distribution, $\ssd^{\text{2D}}$, with dimensions $360 \times 640$, enabling a uniform comparison across models.
We employed regular expressions (regex) parsing to automate the extraction of \say{Yes} or \say{No} answers from the model's responses. 
The binary response format minimized parsing errors during answer retrieval. 
For qualitative analysis, the normalized scores were further re-scaled using a log-scale adjustment to emphasize variations effectively, while maintaining parity with other analyses.

\paragraph{Access dates and configurations.} Visual-language models (VLMs) were queried using an automated pipeline between November 1\textsuperscript{st} and 2\textsuperscript{nd}, 2024. The latest official APIs were used for Claude 3.5 Sonnet~\citep{claude} (\texttt{claude-3-5-sonnet-20241022}), ChatGPT-4o~\citep{chatgpt4} (\texttt{chatgpt-4o-latest}), and Gemini 1.5 Pro~\citep{gemini} (\texttt{gemini-1.5-pro-latest}). LLaVa-34B-v1.6 (Nous-Hermes-2-34B)~\citep{llava-1, llava-2}, accessed via the Replicate API\footnote{\href{https://replicate.com/yorickvp/llava-v1.6-34b}{Replicate LLaVa-v1.6-34B model API documentation}}, is open source (unlike the other proprietary models) but was evaluated through APIs, as we could not perform local inference owing to model size.

\section{Additional Results}
\label{appendix:ablations}
\paragraph{DFM sample and pose ablations.} In this section we provide all the complete ablation tables for individual datasets---\nyu~\citep{NYU} and \rk~\citep{stereomag}. We tabulate the results of the ablation of Diffusion with Forward Model samples~\citep{fwd} in 2D~\cref{table:ablations_Re10k_2D,table:ablations_nyu_2D}, in 2.5D~\cref{table:ablations_Re10k_2.5D,table:ablations_NYU_2.5D} and in 3D~\cref{table:DFM_3D_ablation_Re10k,table:DFM_3D_ablation_NYU}. 
We apply the same random seed to select 10 or 15 samples for random drop-outs, ensuring consistency in the sample selection process, retaining the points and the confidences from the input poses in all cases.

\paragraph{SDXL qualitative analysis.} We also provide the breakdown for the analysis of the SDXL-based analysis in 2D~\cref{table:SDXL_ablations_2D_Re10k,table:SDXL_ablations_2D_NYU} and the analysis in 2.5D~\cref{table:SDXL_NYU_2.5D,table:SDXL_Re10k_2.5D}. In \cref{tab:ablations}, we tabulated the combined ablation study using the total number of valid samples in both datasets.
We present the following metrics along with their means and standard deviations: normalized entropy ($\ent$), normalized cross-entropy ($\cent$), normalized nearest neighbor distance as a percentage ($\dnn$), 2D region-wise accuracy ($\rwa$), and false negative rates ($\falseneg$).

\begin{figure*}[htbp]
    \centering
    \begin{subfigure}[b]{0.32\linewidth}
        \centering
        \begin{tikzpicture}
            \node [anchor=south west, inner sep=0] at (0,0) {
                \includegraphics[width=\linewidth]{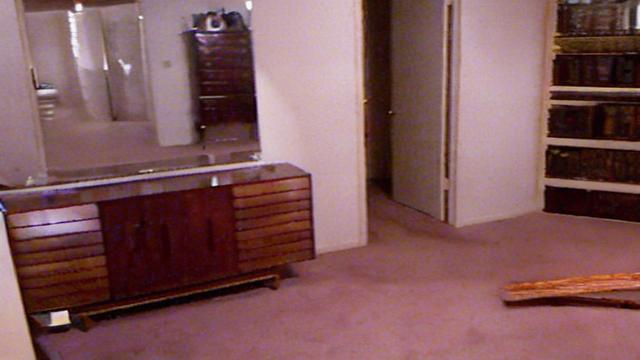}
            };
            \draw[white, thick, dashed] (0.22*\linewidth, 0.01) -- (0.22*\linewidth, 0.5648*\linewidth);
            \draw[white, thick, dashed] (0.78*\linewidth, 0.01) -- (0.78*\linewidth, 0.5648*\linewidth);
        \end{tikzpicture}
        \caption{No object prompt}
        \label{fig:extend_natural}
    \end{subfigure}
    \hfill
    \begin{subfigure}[b]{0.32\linewidth}
        \centering
        \begin{tikzpicture}
            \node [anchor=south west, inner sep=0] at (0,0) {
                \includegraphics[width=\linewidth]{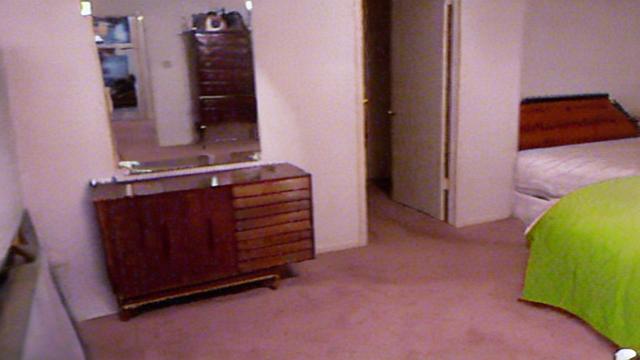}
            };
            \draw[white, thick, dashed] (0.22*\linewidth, 0.01) -- (0.22*\linewidth, 0.5648*\linewidth);
            \draw[white, thick, dashed] (0.78*\linewidth, 0.01) -- (0.78*\linewidth, 0.5648*\linewidth);
        \end{tikzpicture}
        \caption{With object prompt: bed}
        \label{fig:refrigerator_scene}
    \end{subfigure}
    \hfill
    \begin{subfigure}[b]{0.32\linewidth}
        \centering
        \begin{tikzpicture}
            \node [anchor=south west, inner sep=0] at (0,0) {
                \includegraphics[width=\linewidth]{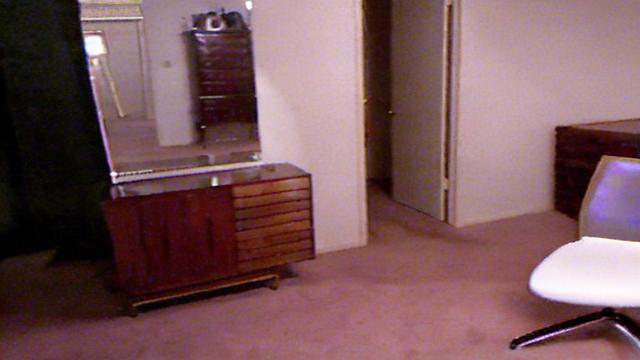}
            };
            \draw[white, thick, dashed] (0.22*\linewidth, 0.01) -- (0.22*\linewidth, 0.5648*\linewidth);
            \draw[white, thick, dashed] (0.78*\linewidth, 0.01) -- (0.78*\linewidth, 0.5648*\linewidth);
        \end{tikzpicture}
        \caption{With object prompt: chair}
        \label{fig:bedroom_chair}
    \end{subfigure}

    \vspace{1em}

    \begin{subfigure}[b]{0.32\linewidth}
        \centering
        \begin{tikzpicture}
            \node [anchor=south west, inner sep=0] at (0,0) {
                \includegraphics[width=\linewidth]{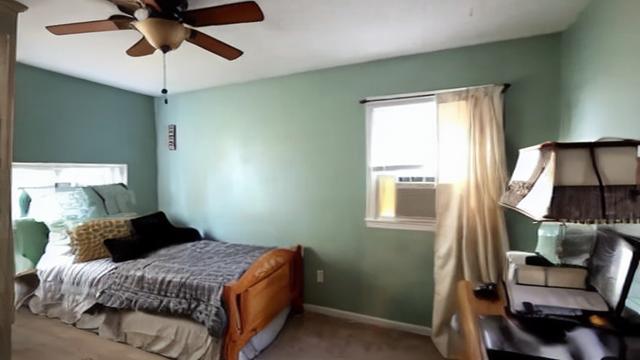}
            };
            \draw[white, thick, dashed] (0.22*\linewidth, 0.01) -- (0.22*\linewidth, 0.5648*\linewidth);
            \draw[white, thick, dashed] (0.78*\linewidth, 0.01) -- (0.78*\linewidth, 0.5648*\linewidth);
        \end{tikzpicture}
        \caption{No object prompt}
        \label{fig:scene_extend}
    \end{subfigure}
    \hfill
    \begin{subfigure}[b]{0.32\linewidth}
        \centering
        \begin{tikzpicture}
            \node [anchor=south west, inner sep=0] at (0,0) {
                \includegraphics[width=\linewidth]{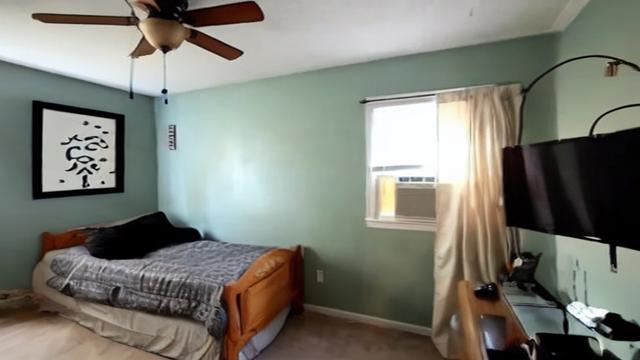}
            };
            \draw[white, thick, dashed] (0.22*\linewidth, 0.01) -- (0.22*\linewidth, 0.5648*\linewidth);
            \draw[white, thick, dashed] (0.78*\linewidth, 0.01) -- (0.78*\linewidth, 0.5648*\linewidth);
        \end{tikzpicture}
        \caption{With object prompt: TV}
        \label{fig:add_tv}
    \end{subfigure}
    \hfill
    \begin{subfigure}[b]{0.32\linewidth}
        \centering
        \begin{tikzpicture}
            \node [anchor=south west, inner sep=0] at (0,0) {
                \includegraphics[width=\linewidth]{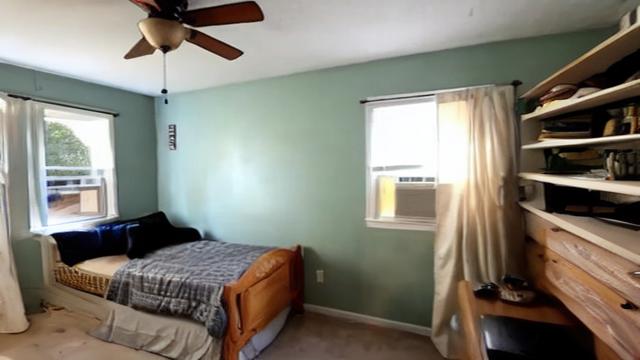}
            };
            \draw[white, thick, dashed] (0.22*\linewidth, 0.01) -- (0.22*\linewidth, 0.5648*\linewidth);
            \draw[white, thick, dashed] (0.78*\linewidth, 0.01) -- (0.78*\linewidth, 0.5648*\linewidth);
        \end{tikzpicture}
        \caption{With object prompt: laptop}
        \label{fig:room_larger}
    \end{subfigure}

    \vspace{1em}

    \begin{subfigure}[b]{0.32\linewidth}
        \centering
        \begin{tikzpicture}
            \node [anchor=south west, inner sep=0] at (0,0) {
                \includegraphics[width=\linewidth]{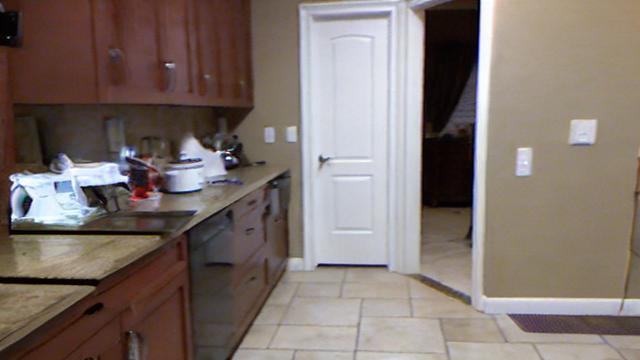}
            };
            \draw[white, thick, dashed] (0.22*\linewidth, 0.01) -- (0.22*\linewidth, 0.5648*\linewidth);
            \draw[white, thick, dashed] (0.78*\linewidth, 0.01) -- (0.78*\linewidth, 0.5648*\linewidth);
        \end{tikzpicture}
        \caption{No object prompt}
        \label{fig:extend_n5}
    \end{subfigure}
    \hfill
    \begin{subfigure}[b]{0.32\linewidth}
        \centering
        \begin{tikzpicture}
            \node [anchor=south west, inner sep=0] at (0,0) {
                \includegraphics[width=\linewidth]{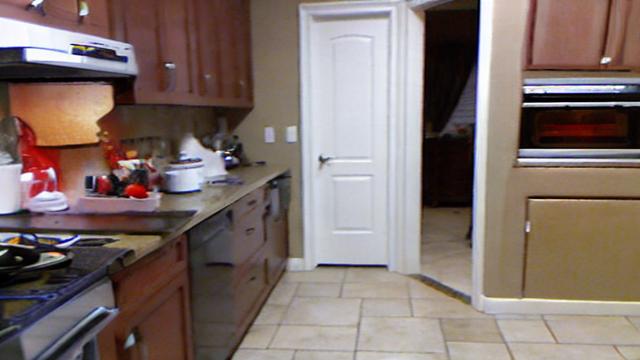}
            };
            \draw[white, thick, dashed] (0.22*\linewidth, 0.01) -- (0.22*\linewidth, 0.5648*\linewidth);
            \draw[white, thick, dashed] (0.78*\linewidth, 0.01) -- (0.78*\linewidth, 0.5648*\linewidth);
        \end{tikzpicture}
        \caption{With object prompt: oven}
        \label{fig:kitchen_oven}
    \end{subfigure}
    \hfill
    \begin{subfigure}[b]{0.32\linewidth}
        \centering
        \begin{tikzpicture}
            \node [anchor=south west, inner sep=0] at (0,0) {
                \includegraphics[width=\linewidth]{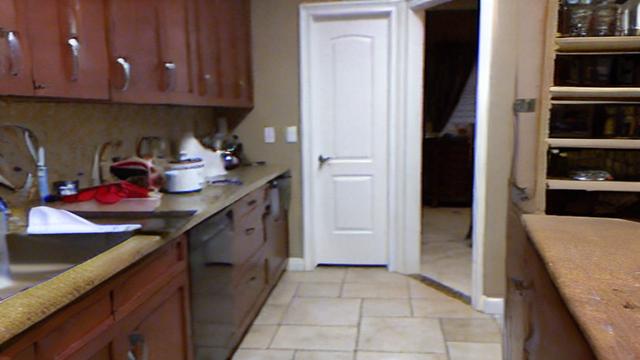}
            };
            \draw[white, thick, dashed] (0.22*\linewidth, 0.01) -- (0.22*\linewidth, 0.5648*\linewidth);
            \draw[white, thick, dashed] (0.78*\linewidth, 0.01) -- (0.78*\linewidth, 0.5648*\linewidth);
        \end{tikzpicture}
        \caption{With object prompt: sink}
        \label{fig:kitchen_sink}
    \end{subfigure}

    \vspace{1em}

    \begin{subfigure}[b]{0.32\linewidth}
        \centering
        \begin{tikzpicture}
            \node [anchor=south west, inner sep=0] at (0,0) {
                \includegraphics[width=\linewidth]{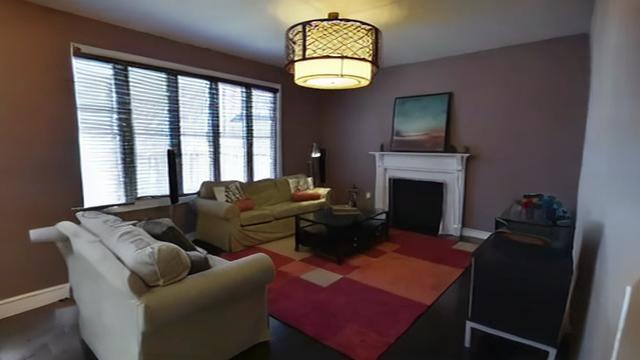}
            };
            \draw[white, thick, dashed] (0.22*\linewidth, 0.01) -- (0.22*\linewidth, 0.5648*\linewidth);
            \draw[white, thick, dashed] (0.78*\linewidth, 0.01) -- (0.78*\linewidth, 0.5648*\linewidth);
        \end{tikzpicture}
        \caption{No object prompt}
        \label{fig:living_room_tv}
    \end{subfigure}
    \hfill
    \begin{subfigure}[b]{0.32\linewidth}
        \centering
        \begin{tikzpicture}
            \node [anchor=south west, inner sep=0] at (0,0) {
                \includegraphics[width=\linewidth]{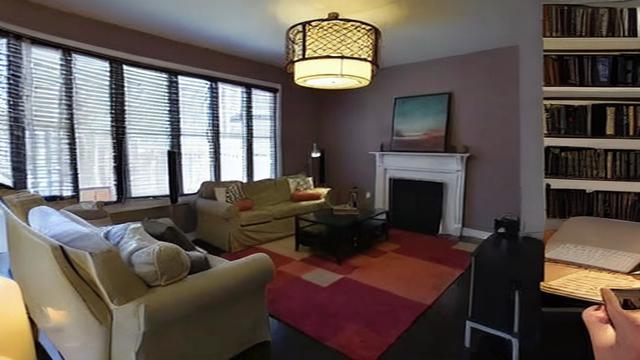}
            };
            \draw[white, thick, dashed] (0.22*\linewidth, 0.01) -- (0.22*\linewidth, 0.5648*\linewidth);
            \draw[white, thick, dashed] (0.78*\linewidth, 0.01) -- (0.78*\linewidth, 0.5648*\linewidth);
        \end{tikzpicture}
        \caption{With object prompt: book}
    \end{subfigure}
    \hfill
    \begin{subfigure}[b]{0.32\linewidth}
        \centering
        \begin{tikzpicture}
            \node [anchor=south west, inner sep=0] at (0,0) {
                \includegraphics[width=\linewidth]{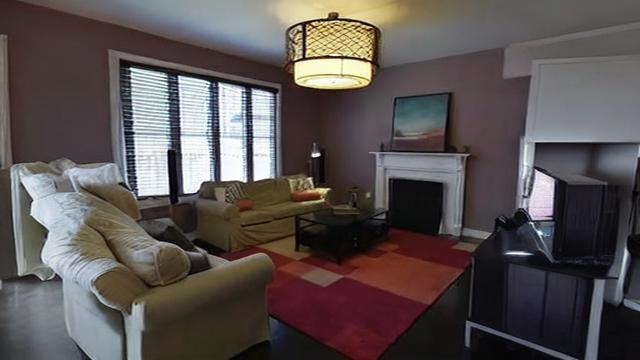}
            };
            \draw[white, thick, dashed] (0.22*\linewidth, 0.01) -- (0.22*\linewidth, 0.5648*\linewidth);
            \draw[white, thick, dashed] (0.78*\linewidth, 0.01) -- (0.78*\linewidth, 0.5648*\linewidth);
        \end{tikzpicture}
        \caption{With object prompt: TV}
        \label{fig:scene_tv}
    \end{subfigure}

    \vspace{1em}

    \begin{subfigure}[b]{0.32\linewidth}
        \centering
        \begin{tikzpicture}
            \node [anchor=south west, inner sep=0] at (0,0) {
                \includegraphics[width=\linewidth]{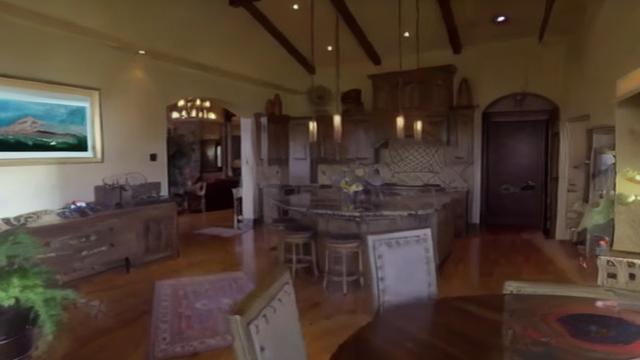}
            };
            \draw[white, thick, dashed] (0.22*\linewidth, 0.01) -- (0.22*\linewidth, 0.5648*\linewidth);
            \draw[white, thick, dashed] (0.78*\linewidth, 0.01) -- (0.78*\linewidth, 0.5648*\linewidth);
        \end{tikzpicture}
        \caption{No object prompt}
    \end{subfigure}
    \hfill
    \begin{subfigure}[b]{0.32\linewidth}
        \centering
        \begin{tikzpicture}
            \node [anchor=south west, inner sep=0] at (0,0) {
                \includegraphics[width=\linewidth]{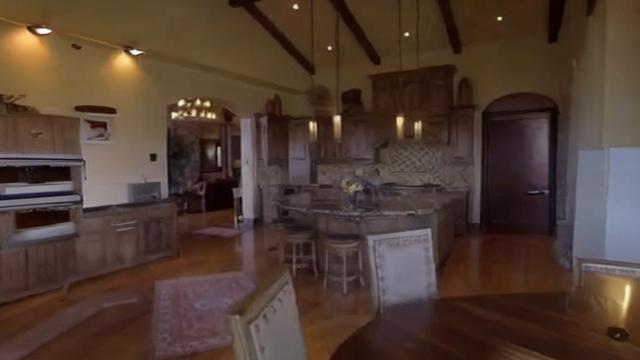}
            };
            \draw[white, thick, dashed] (0.22*\linewidth, 0.01) -- (0.22*\linewidth, 0.5648*\linewidth);
            \draw[white, thick, dashed] (0.78*\linewidth, 0.01) -- (0.78*\linewidth, 0.5648*\linewidth);
        \end{tikzpicture}
        \caption{With object prompt: oven}
        \label{fig:extend_depth}
    \end{subfigure}
    \hfill
    \begin{subfigure}[b]{0.32\linewidth}
        \centering
        \begin{tikzpicture}
            \node [anchor=south west, inner sep=0] at (0,0) {
                \includegraphics[width=\linewidth]{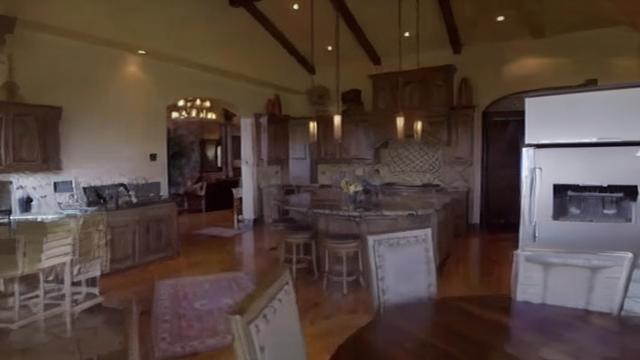}
            };
            \draw[white, thick, dashed] (0.22*\linewidth, 0.01) -- (0.22*\linewidth, 0.5648*\linewidth);
            \draw[white, thick, dashed] (0.78*\linewidth, 0.01) -- (0.78*\linewidth, 0.5648*\linewidth);
        \end{tikzpicture}
        \caption{With object prompt: refrigerator}
    \end{subfigure}
    \caption{\textbf{Samples from SDXL.} We prompted SDXL with and without object cues. The image within the dotted lines is the input image.}
    \label{fig:SDXL_grid}
\end{figure*}

\begin{figure*}[!htbp]
    \centering
    \rotatebox[origin=l]{90}{{\footnotesize TV}}
    \begin{subfigure}[]{0.27\linewidth}
        \centering
        \begin{tikzpicture}
            \node [anchor=south west,inner sep=0] at (0,0) {\includegraphics[width=\linewidth]{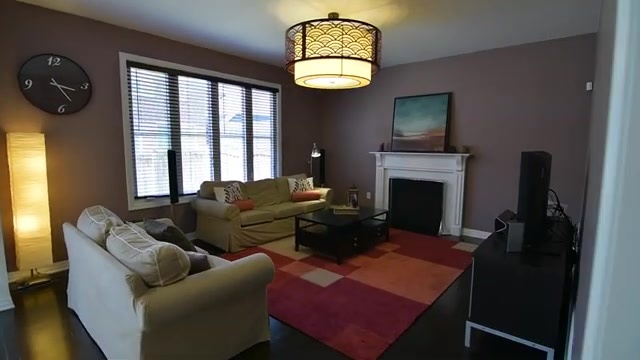}};
            \draw[white, thick, dashed] (0.22*\linewidth, 0.01) -- (0.22*\linewidth, 0.5648*\linewidth);
            \draw[white, thick, dashed] (0.78*\linewidth, 0.01) -- (0.78*\linewidth, 0.5648*\linewidth);
        \end{tikzpicture}
        \label{fig:qual_r1_col1}
    \end{subfigure}\hfill
    \begin{subfigure}[]{0.152\linewidth}
        \centering
        \includegraphics[width=\linewidth]{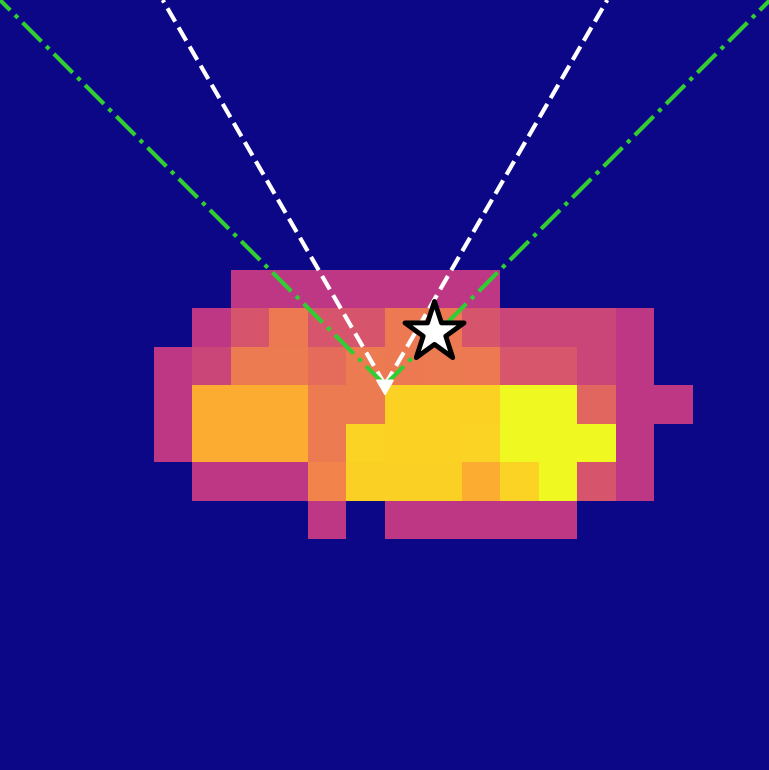}
        \label{fig:qual_r1_col2}
    \end{subfigure}\hfill
    \begin{subfigure}[]{0.27\linewidth}
        \centering
        \includegraphics[width=\linewidth]{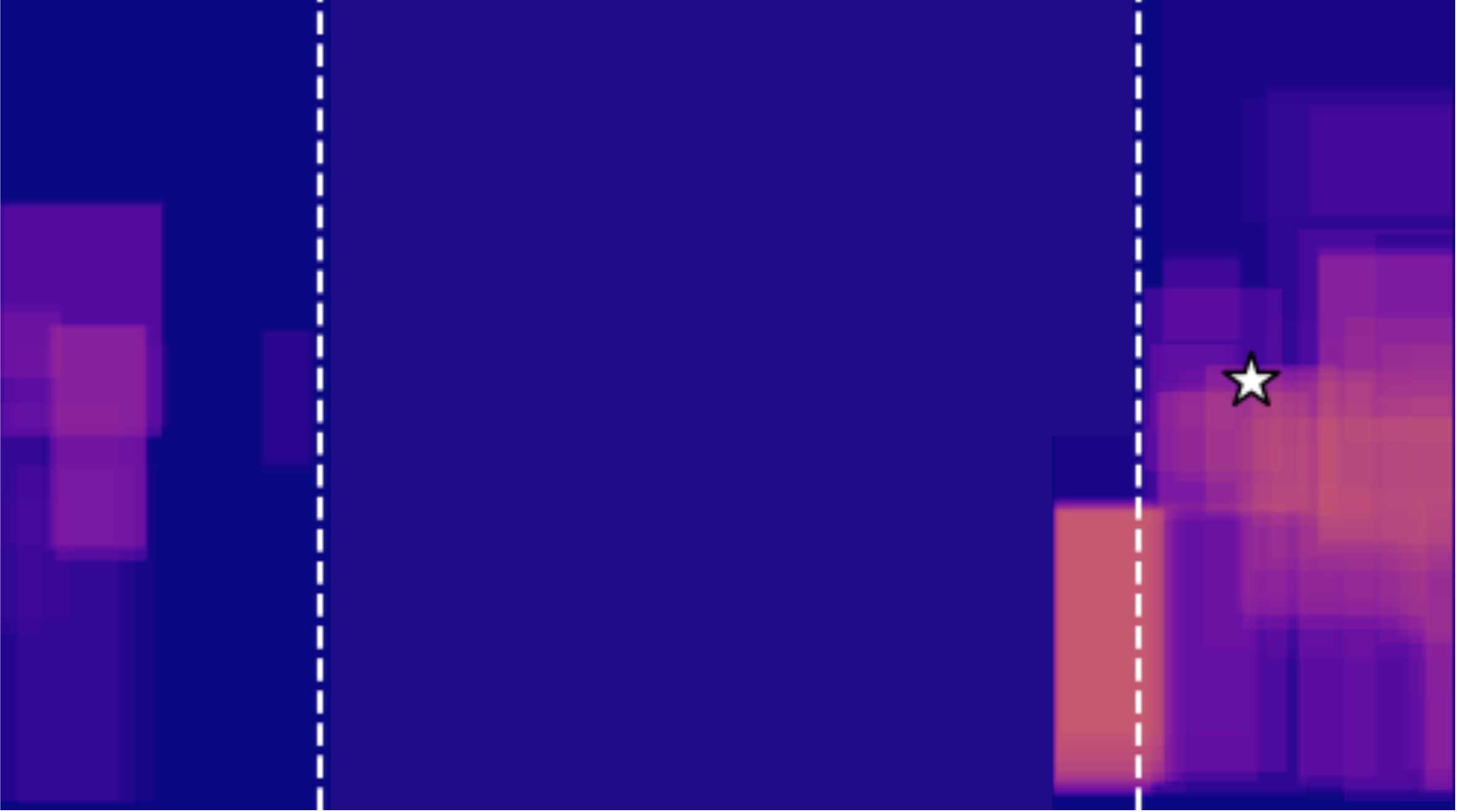}
        \label{fig:qual_r1_col3}
    \end{subfigure}\hfill
    \begin{subfigure}[]{0.27\linewidth}
        \centering
        \includegraphics[width=\linewidth]{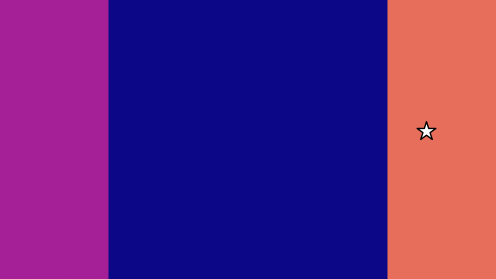}
        \label{fig:qual_r1_col4}
    \end{subfigure}\\
    \vspace{-1em}
    \makebox[5.6pt][l]{\rotatebox[origin=c]{90}{{\footnotesize CHAIR}}}
    \begin{subfigure}[]{0.27\linewidth}
        \centering
        \begin{tikzpicture}
            \node [anchor=south west,inner sep=0] at (0,0) {\includegraphics[width=\linewidth]{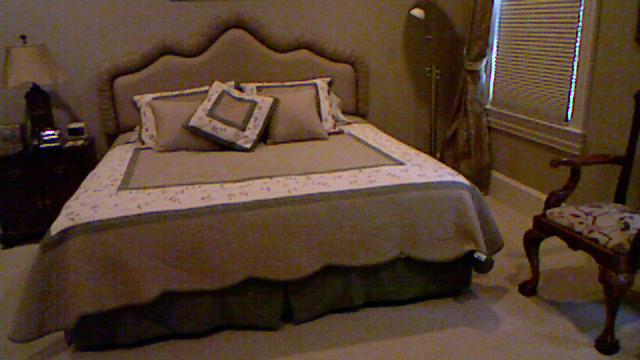}};
            \draw[white, thick, dashed] (0.22*\linewidth, 0.01) -- (0.22*\linewidth, 0.5648*\linewidth);
            \draw[white, thick, dashed] (0.78*\linewidth, 0.01) -- (0.78*\linewidth, 0.5648*\linewidth);
        \end{tikzpicture}
        \label{fig:qual_r2_col1}
    \end{subfigure}\hfill
    \begin{subfigure}[]{0.152\linewidth}
        \centering
        \includegraphics[width=\linewidth]{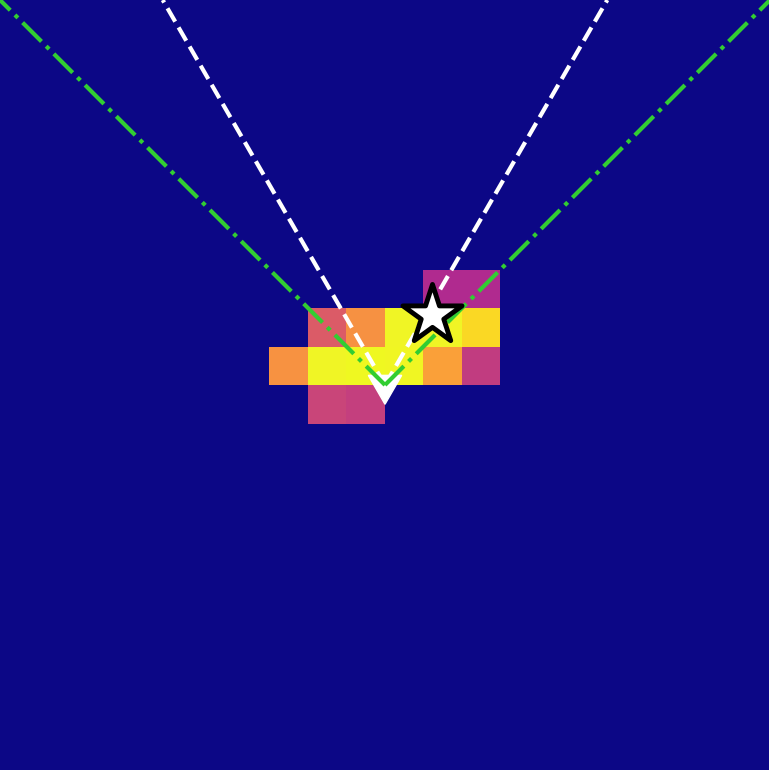}
        \label{fig:qual_r2_col2}
    \end{subfigure}\hfill
    \begin{subfigure}[]{0.27\linewidth}
        \centering
        \includegraphics[width=\linewidth]{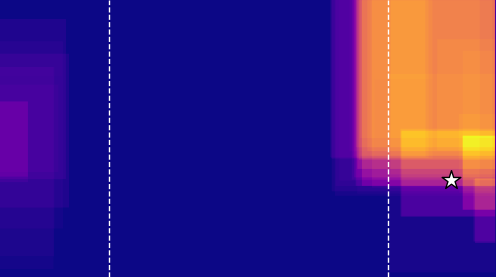}
        \label{fig:qual_r2_col3}
    \end{subfigure}\hfill
    \begin{subfigure}[]{0.27\linewidth}
        \centering
        \includegraphics[width=\linewidth]{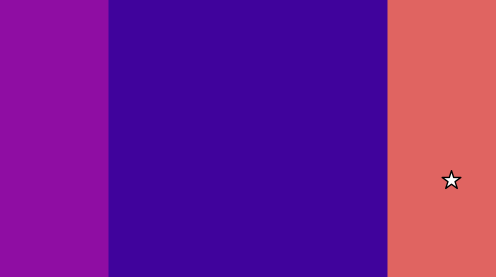}
        \label{fig:qual_r2_col4}
    \end{subfigure}\\
    \vspace{-1em}
    \rotatebox[origin=l]{90}{{\footnotesize BOOK}}
    \begin{subfigure}[]{0.27\linewidth}
        \centering
        \begin{tikzpicture}
            \node [anchor=south west,inner sep=0] at (0,0) {\includegraphics[width=\linewidth]{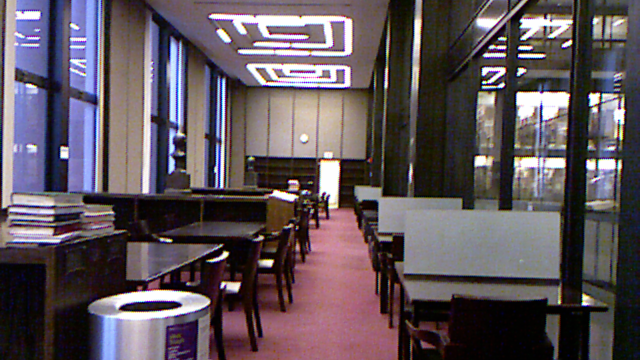}};
            \draw[white, thick, dashed] (0.22*\linewidth, 0.01) -- (0.22*\linewidth, 0.5648*\linewidth);
            \draw[white, thick, dashed] (0.78*\linewidth, 0.01) -- (0.78*\linewidth, 0.5648*\linewidth);
        \end{tikzpicture}
        \label{fig:qual_r3_col1}
    \end{subfigure}\hfill
    \begin{subfigure}[]{0.152\linewidth}
        \centering
        \includegraphics[width=\linewidth]{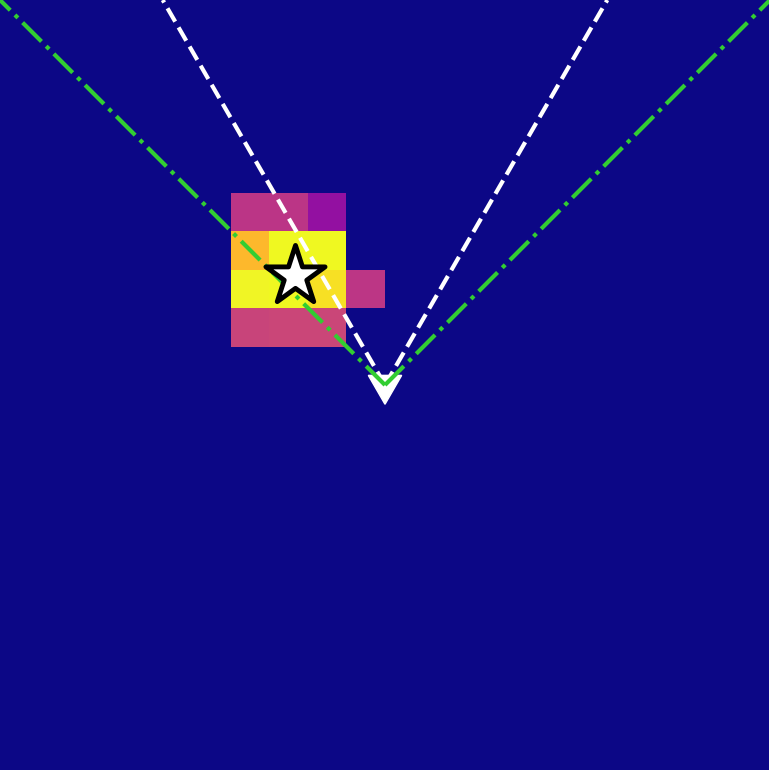}
        \label{fig:qual_r3_col2}
    \end{subfigure}\hfill
    \begin{subfigure}[]{0.27\linewidth}
        \centering
        \includegraphics[width=\linewidth]{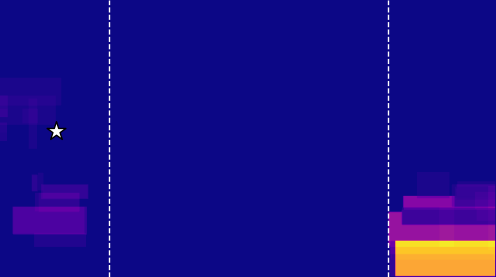}
        \label{fig:qual_r3_col3}
    \end{subfigure}\hfill
    \begin{subfigure}[]{0.27\linewidth}
        \centering
        \includegraphics[width=\linewidth]{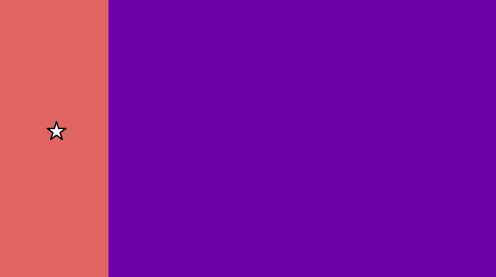}
        \label{fig:qual_r3_col4}
    \end{subfigure}\\
    \vspace{-1em}
    \rotatebox[origin=l]{90}{{\footnotesize SINK}}
    \begin{subfigure}[]{0.27\linewidth}
        \centering
        \begin{tikzpicture}
            \node [anchor=south west,inner sep=0] at (0,0) {\includegraphics[width=\linewidth]{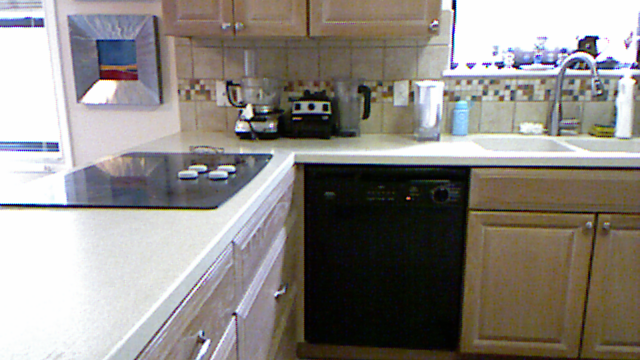}};
            \draw[white, thick, dashed] (0.22*\linewidth, 0.01) -- (0.22*\linewidth, 0.5648*\linewidth);
            \draw[white, thick, dashed] (0.78*\linewidth, 0.01) -- (0.78*\linewidth, 0.5648*\linewidth);
        \end{tikzpicture}
        \label{fig:qual_r4_col1}
    \end{subfigure}\hfill
    \begin{subfigure}[]{0.152\linewidth}
        \centering
        \includegraphics[width=\linewidth]{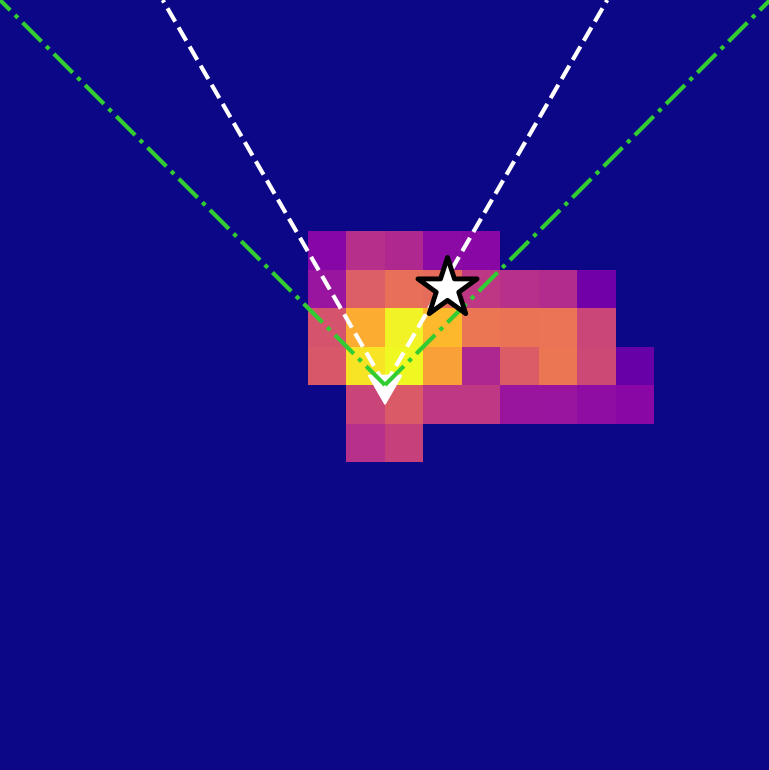}
        \label{fig:qual_r4_col2}
    \end{subfigure}\hfill
    \begin{subfigure}[]{0.27\linewidth}
        \centering
        \includegraphics[width=\linewidth]{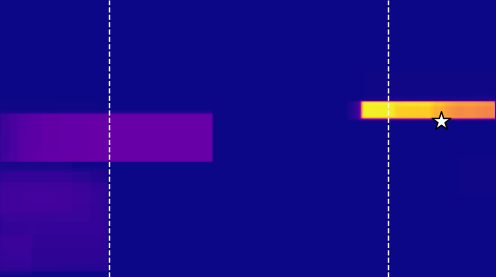}
        \label{fig:qual_r4_col3}
    \end{subfigure}\hfill
    \begin{subfigure}[]{0.27\linewidth}
        \centering
        \includegraphics[width=\linewidth]{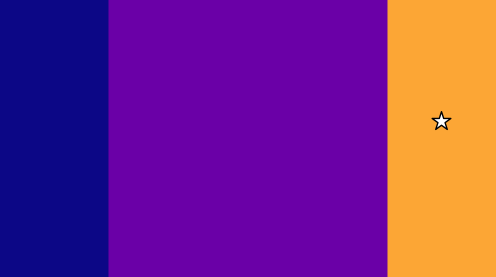}
        \label{fig:qual_r4_col4}
    \end{subfigure}\\
    \vspace{-1em}
    \rotatebox[origin=l]{90}{{\footnotesize OVEN}}
    \begin{subfigure}[]{0.27\linewidth}
        \centering
        \begin{tikzpicture}
            \node [anchor=south west,inner sep=0] at (0,0) {\includegraphics[width=\linewidth]{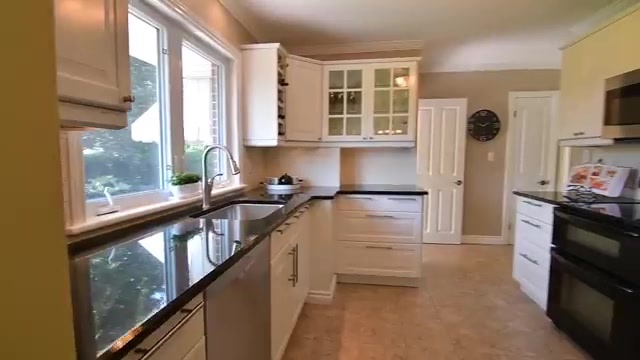}};
            \draw[white, thick, dashed] (0.22*\linewidth, 0.01) -- (0.22*\linewidth, 0.5648*\linewidth);
            \draw[white, thick, dashed] (0.78*\linewidth, 0.01) -- (0.78*\linewidth, 0.5648*\linewidth);
        \end{tikzpicture}
        \label{fig:qual_r5_col1}
    \end{subfigure}\hfill
    \begin{subfigure}[]{0.152\linewidth}
        \centering
        \includegraphics[width=\linewidth]{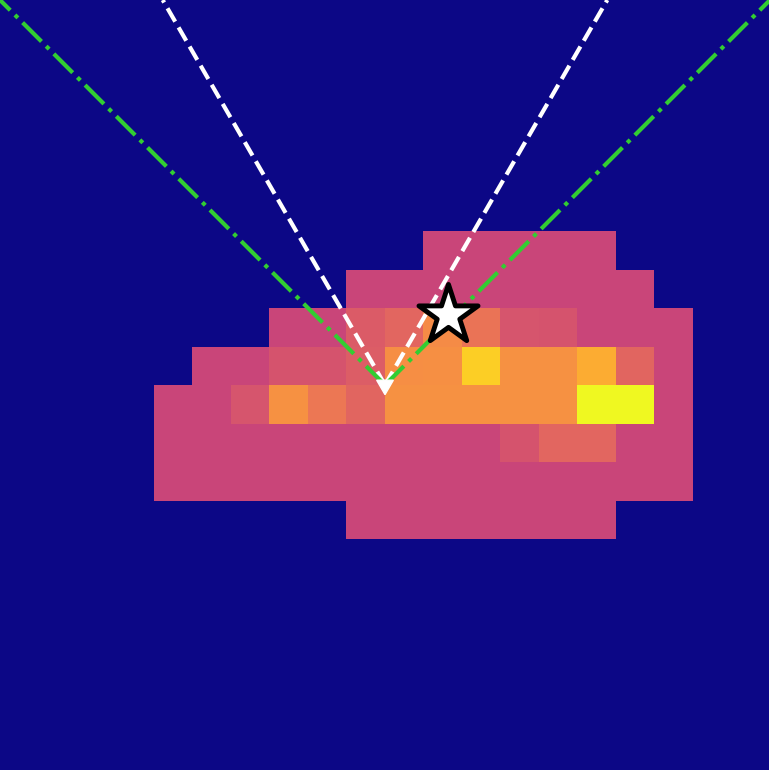}
        \label{fig:qual_r5_col2}
    \end{subfigure}\hfill
    \begin{subfigure}[]{0.27\linewidth}
        \centering
        \includegraphics[width=\linewidth]{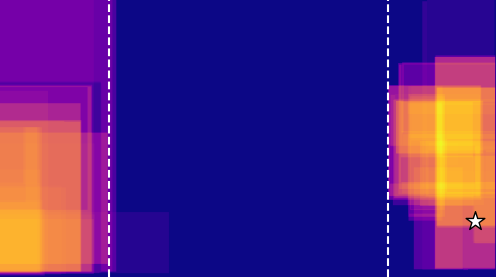}
        \label{fig:qual_r5_col3}
    \end{subfigure}\hfill
    \begin{subfigure}[]{0.27\linewidth}
        \centering
        \includegraphics[width=\linewidth]{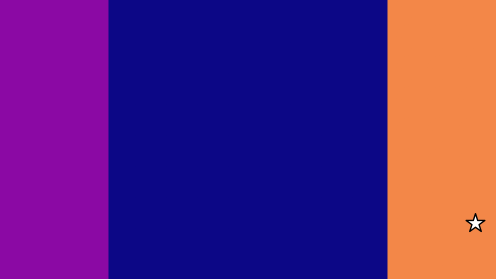}
        \label{fig:qual_r5_col4}
    \end{subfigure}\\
    \vspace{-1em}
    \makebox[5.6pt][l]{\rotatebox[origin=c]{90}{{\footnotesize CHAIR}}}
    \begin{subfigure}[]{0.27\linewidth}
        \centering
        \begin{tikzpicture}
            \node [anchor=south west,inner sep=0] at (0,0) {\includegraphics[width=\linewidth]{heatmaps/scene_4_GT.jpg}};
            \draw[white, thick, dashed] (0.22*\linewidth, 0.01) -- (0.22*\linewidth, 0.5648*\linewidth);
            \draw[white, thick, dashed] (0.78*\linewidth, 0.01) -- (0.78*\linewidth, 0.5648*\linewidth);
        \end{tikzpicture}
        \label{fig:qual_r6_col1}
    \end{subfigure}\hfill
    \begin{subfigure}[]{0.152\linewidth}
        \centering
        \includegraphics[width=\linewidth]{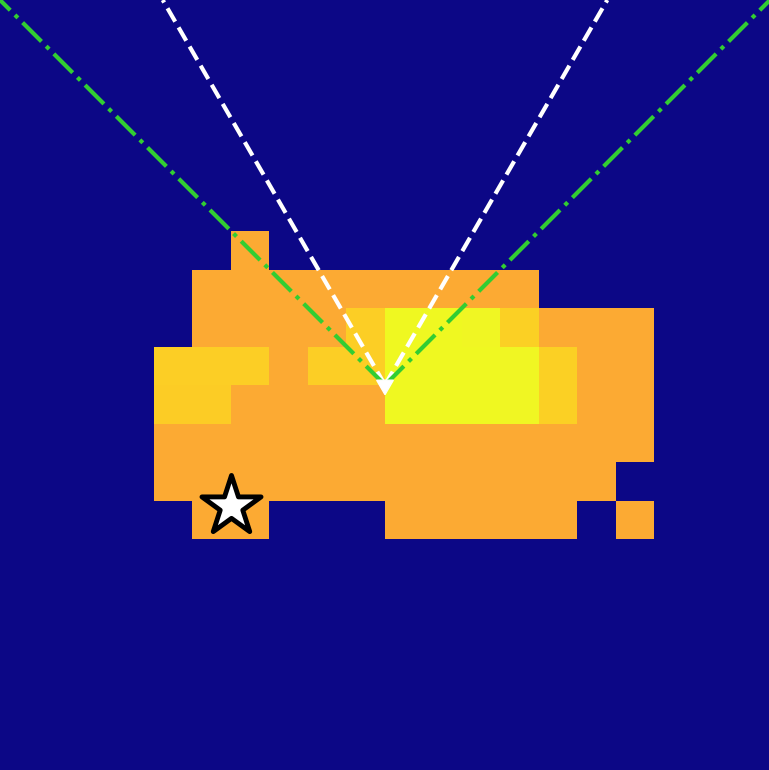}
        \label{fig:qual_r6_col2}
    \end{subfigure}\hfill
    \begin{subfigure}[]{0.27\linewidth}
        \centering
        \includegraphics[width=\linewidth]{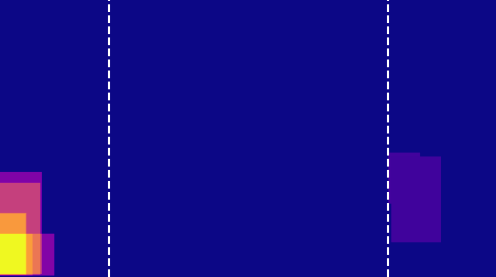}
        \label{fig:qual_r6_col3}
    \end{subfigure}\hfill
    \begin{subfigure}[]{0.27\linewidth}
        \centering
        \includegraphics[width=\linewidth]{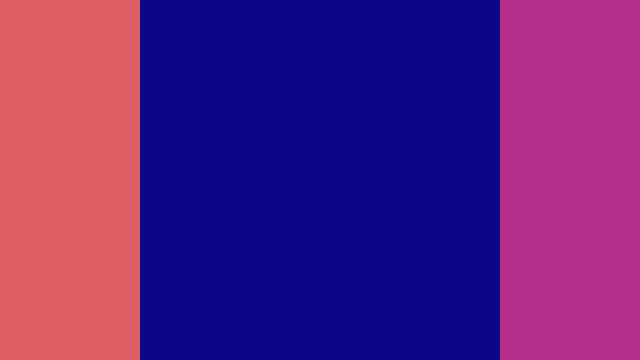}
        \label{fig:qual_r6_col4}
    \end{subfigure}\\
    \vspace{-1em}
    \rotatebox[origin=l]{90}{{\footnotesize BED}}
    \begin{subfigure}[]{0.27\linewidth}
        \centering
        \begin{tikzpicture}
            \node [anchor=south west,inner sep=0] at (0,0) {\includegraphics[width=\linewidth]{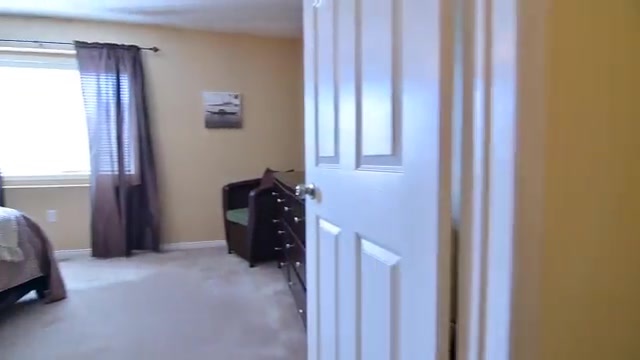}};
            \draw[white, thick, dashed] (0.22*\linewidth, 0.01) -- (0.22*\linewidth, 0.5648*\linewidth);
            \draw[white, thick, dashed] (0.78*\linewidth, 0.01) -- (0.78*\linewidth, 0.5648*\linewidth);
        \end{tikzpicture}
        \caption{$\inim$ (center) and $\outim$ (full)}
        \label{fig:qual_r7_col1}
    \end{subfigure}\hfill
    \begin{subfigure}[]{0.152\linewidth}
        \centering
        \includegraphics[width=\linewidth]{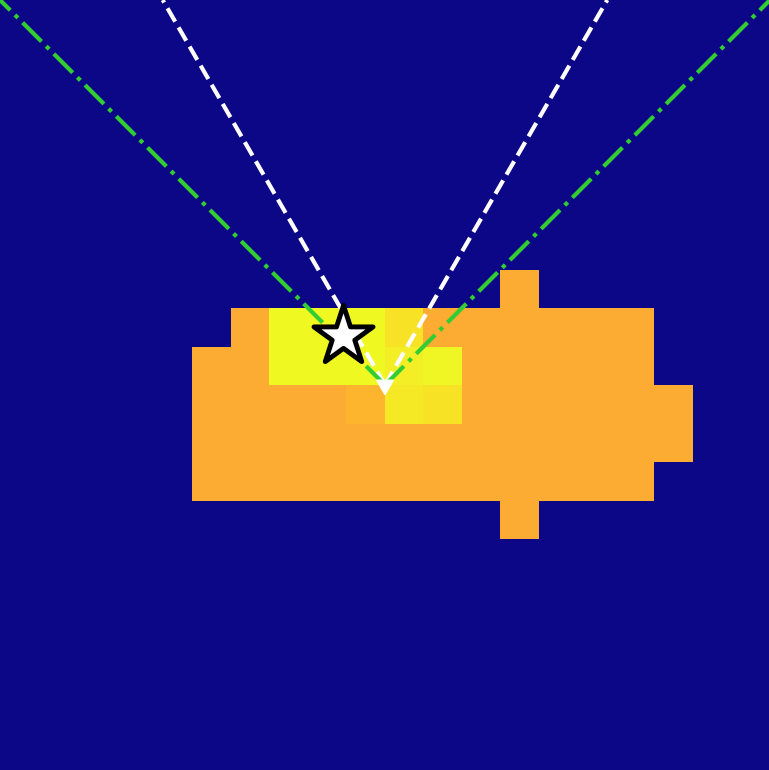}
        \caption{DFM $\sd^\text{3D}$}
        \label{fig:qual_r7_col2}
    \end{subfigure}\hfill
    \begin{subfigure}[]{0.27\linewidth}
        \centering
        \includegraphics[width=\linewidth]{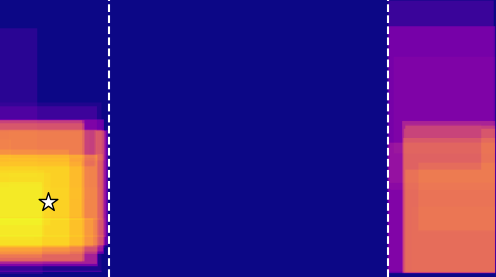}
        \caption{SDXL $\sd^\text{2D}$}
        \label{fig:qual_r7_col3}
    \end{subfigure}\hfill
    \begin{subfigure}[]{0.27\linewidth}
        \centering
        \includegraphics[width=\linewidth]{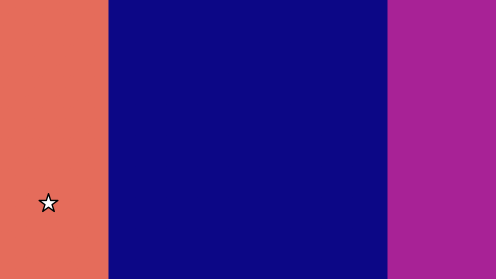}
        \caption{ChatGPT-4o $\sd^\text{2D}$}
        \label{fig:qual_r7_col4}
    \end{subfigure}\\
    \vspace{3pt}
    {\tt 0}\ \includegraphics[width=0.5\linewidth, trim={12pt 8pt 10pt 8pt}, clip]{images/plasma_scale.pdf}\ {\tt 1}
    \caption{\textbf{Additional qualitative results.} We provide additional results for different objects and their corresponding heatmaps. White stars indicate ground truth position if applicable on the corresponding heatmap(s).}
    \label{fig:new}
\end{figure*}

\paragraph{Additional qualitative results.} 
\label{subsec:qual}
In this section, we present additional qualitative results to supplement our analysis. Samples generated using the DFM are illustrated in \cref{fig:DFM_samples}, while those from SDXL are shown in \cref{fig:SDXL_grid}. 
As expected, the DFM-based model exhibits poor qualitative performance on the NYU dataset due to its limited ability to generalize to out-of-distribution data \citep{fwd}. 
The DFM-based model, as depicted in rows 6 and 7 of \cref{fig:new}, frequently predicts flat distributions. While these distributions capture uncertainty and can localize the ground truth position, their higher uncertainty reflects in the normalization process followed by log-scaling, resulting in a wide probability mass spread across the distribution ($\sd^{3D}$).
For top-down heatmaps, we visualize the maximum normalized confidence value along the $+y$-axis for each column of the voxel grid. These visualizations may show spillover into regions without detected objects, which is an artifact of the softmax normalization and voxelization processes. In some 2D heatmaps, ground truth positions are omitted when the object is not visible in the ground truth image but is present in the scene.
For SDXL-generated samples, we observe that semantic quality is significantly influenced by the guidance scale and the text prompt. When object cues are provided, both qualitative and quantitative performance improve noticeably. Conversely, generation without object cues or prompts yields minimal differences in quantitative results, as both scenarios result in limited detections. The qualitative results for SDXL analysis (column 3 in \cref{fig:qual,fig:new}) are based on samples generated using prompts that include object cues.
In certain SDXL-based heatmaps, spillovers into the central region may occur because the detector perceives the entire object, shared between the input image and the outpainted region, as part of the object.

\paragraph{Failure case analysis.}
In our current DFM experiments, 57\% of failures arise from object prediction errors (non-occluded), while 43\% result from detector false negatives in ambiguous contexts (\eg, multiple doors in a corridor). 
No failures stem from the inability to predict occluded objects.
SDXL-based methods with object prompts show no failures.
In VLM experiments, all failures are direct prediction errors of the model.

\paragraph{Outdoor Scenes.}
Our framework is also applicable to outdoor scenes. We demonstrate this with an outdoor scene from the \rk dataset featuring the COCO object `car' (\cref{fig:outdoor}, \cref{tab:1}). The SDXL and VLM-based models perform well as expected while the DFM struggles in outdoor settings due to limited training on large-scale outdoor data, retraining which exceeds our computational resources.

\clearpage
\begin{table*}[!ht]
    \centering
    \caption{Results for the outdoor scene--object pair in \cref{fig:outdoor}.}
    \label{tab:1}
    \scriptsize
    \renewcommand{\arraystretch}{1.2} 
    \setlength{\tabcolsep}{3pt}
    \resizebox{\textwidth}{!}{
    \begin{tabular*}{\textwidth}{@{\extracolsep{\fill}} lcccc|ccc|ccc}
        \toprule
        \multirow{2}{*}{\textbf{Methods}} & 
        \multicolumn{4}{c|}{\textbf{2D Experiment}} & 
        \multicolumn{3}{c|}{\textbf{2.5D Experiment}} & 
        \multicolumn{3}{c}{\textbf{3D Experiment}} \\
        \cmidrule(lr){2-5} \cmidrule(lr){6-8} \cmidrule(lr){9-11}
         & $\cent\downarrow$ & $\ent\downarrow$ & $\dnn (\%)\downarrow$ & $\rwa\uparrow$ 
         & $\cent\downarrow$ & $\ent\downarrow$ & $\dnn (\%)\downarrow$ 
         & $\cent\downarrow$ & $\ent\downarrow$ & $\dnn (\%)\downarrow$ \\
        \midrule
        DFM (25 samples, 4 poses) 
           & 1.000 & 1.000 & $\infty$ & 0.670 
           & 0.998 & 1.000 & $\infty$  
           & 0.998 & 1.000 & $\infty$ \\
        \midrule
        SDXL (w. obj prompt) 
           & 0.891 & 0.771 & 0.000 & 0.670 
           & 0.788 & 0.962 & 0.000 
           & -- & --  & -- \\
        \midrule
        ChatGPT-4o 
           & 0.880 & 0.879 & 0.000 & 0.670 
           & -- & -- & -- 
           & -- & -- & -- \\
        Claude 3.5 Sonnet 
           & 0.880 & 0.879 & 0.000 & 0.670 
           & -- & -- & -- 
           & -- & -- & -- \\
        Gemini 1.5 Pro 
           & 0.994 & 0.993 & 0.000 & 0.670 
           & -- & -- & -- 
           & -- & -- & -- \\
        LLaVa-v1.6-34b 
           & 0.999 & 0.998 & 0.000 & 0.670 
           & -- & -- & -- 
           & -- & -- & -- \\
        \bottomrule
    \end{tabular*}
    }
\end{table*}

\begin{table*}[!htbp]
\centering
\caption{2D Metrics for \rk\ with lower number of poses and samples, used to study the trend.}
\renewcommand{\arraystretch}{1} 
\begin{tabular*}{\textwidth}{@{\extracolsep{\fill}} lccccc}
\toprule
\textbf{Methods} & $\cent\downarrow$ & $\ent\downarrow$ & $\dnn(\%)\downarrow$ & $\falseneg \downarrow$ & $\rwa\uparrow$ \\
\midrule
DFM ($n_s = 25, k=4$) & 1.555 ± 1.460 & 0.774 ± 0.119 & 5.990 ± 39.470 & 0.032 & 0.747 ± 0.139 \\
DFM ($n_{s}=25, k=3$) & 1.877 ± 1.327 & 0.818 ± 0.112 & 18.136 ± 32.482 & 0.121 & 0.626 ± 0.107 \\
DFM ($n_{s}=25, k=2$) & 2.328 ± 1.041 & 0.902 ± 0.067 & 14.037 ± 42.005 & 0.868 & 0.581 ± 0.182 \\
\midrule
DFM ($n_{s}=15, k=4$) & 1.736 ± 0.414 & 0.898 ± 0.117 & 15.334 ± 23.380 & 0.605 & 0.573 ± 0.198  \\
DFM ($n_{s}=10, k=4$) & 2.247 ± 1.206 & 0.923 ± 0.014 & 14.082 ± 29.083 & 0.711 & 0.550 ± 0.337 \\
\bottomrule
\end{tabular*}
\label{table:ablations_Re10k_2D}
\end{table*}

\begin{table*}[!htbp]
\centering
\caption{2D Metrics for \nyu\ with lower number of poses and samples, used to study the trend.}
\renewcommand{\arraystretch}{1}
\begin{tabular*}{\textwidth}{@{\extracolsep{\fill}} lccccc}
\toprule
\textbf{Methods} & $\cent\downarrow$ & $\ent\downarrow$ & $\dnn(\%)\downarrow$ & $\falseneg \downarrow$ & $\rwa\uparrow$ \\
\midrule
DFM ($n_{s}=25, k=4$) & 1.661 ± 1.336 & 0.696 ± 0.128 & 6.217 ± 36.112 & 0.0556 & 0.765 ± 0.236 \\
DFM ($n_{s}=25, k=3$) & 2.053 ± 0.351 & 0.929 ± 0.145 & 20.459 ± 35.216 & 0.173 & 0.589 ± 0.340 \\
DFM ($n_{s}=25, k=2$) & 2.691 ± 1.187 & 0.978 ± 0.072 & 15.871 ± 45.028 & 0.900 & 0.545 ± 0.190 \\
\midrule
DFM ($n_{s}=15, k=4$) & 1.915 ± 0.497 & 0.969 ± 0.030 & 17.242 ± 25.764 & 0.558 & 0.623 ± 0.353 \\
DFM ($n_{s}=10, k=4$) & 2.498 ± 1.354 & 0.998 ± 0.012 & 15.936 ± 31.011 & 0.739 & 0.614 ± 0.356 \\
\bottomrule
\end{tabular*}
\label{table:ablations_nyu_2D}
\end{table*}

\begin{table*}[!htbp]
\centering
\caption{2.5D Metrics for \rk\ with lower number of poses and samples, used to study the trend.}
\renewcommand{\arraystretch}{1} 
\begin{tabular*}{\textwidth}{@{\extracolsep{\fill}} lcccc}
\toprule
\textbf{Methods} & $\cent\downarrow$ & $\ent\downarrow$ & $\dnn(\%)\downarrow$ & $\falseneg \downarrow$ \\
\midrule
DFM ($n_{s}=25, k=4$) & 1.955 ± 1.033 & 0.530 ± 0.441 & 5.128 ± 3.042 & 0.032 \\
DFM ($n_{s}=25, k=3$) & 2.018 ± 0.053 & 0.814 ± 0.080 & 12.889 ± 1.067 & 0.121 \\
DFM ($n_{s}=25, k=2$) & 2.586 ± 2.650 & 0.996 ± 0.003 & 13.461 ± 1.552 & 0.868 \\
\midrule
DFM ($n_{s}=15, k=4$) & 2.113 ± 0.778 & 0.877 ± 0.067 & 9.018 ± 6.332 & 0.605 \\
DFM ($n_{s}=10, k=4$) & 2.498 ± 1.354 & 0.997 ± 0.002 & 18.018 ± 9.607 & 0.711 \\
\bottomrule
\end{tabular*}
\label{table:ablations_Re10k_2.5D}
\end{table*}

\begin{table*}[!htbp]
\centering
\caption{2.5D Metrics for \nyu\ with lower number of poses and samples, used to study the trend.}
\renewcommand{\arraystretch}{1}
\begin{tabular*}{\textwidth}{@{\extracolsep{\fill}} lcccc}
\toprule
\textbf{Methods} & $\cent\downarrow$ & $\ent\downarrow$ & $\dnn(\%)\downarrow$ & $\falseneg \downarrow$ \\
\midrule
DFM ($n_{s}=25, k=4$) & 1.785 ± 2.101 & 0.610 ± 0.207 & 4.214 ± 2.803 & 0.0556 \\
DFM ($n_{s}=25, k=3$) & 2.067 ± 0.093 & 0.798 ± 0.089 & 11.442 ± 1.305 & 0.152 \\
DFM ($n_{s}=25, k=2$) & 2.523 ± 2.301 & 0.990 ± 0.005 & 12.786 ± 1.781 & 0.837 \\
\midrule
DFM ($n_{s}=15, k=4$) & 2.184 ± 0.652 & 0.867 ± 0.079 & 8.336 ± 6.174 & 0.772 \\
DFM ($n_{s}=10, k=4$) & 2.451 ± 1.231 & 0.995 ± 0.003 & 17.124 ± 9.403 & 0.808 \\
\bottomrule
\end{tabular*}
\label{table:ablations_NYU_2.5D}
\end{table*}

\newpage
\begin{table*}[!htbp]
\centering
\caption{3D Ablation study of DFM with different configurations of $n_s$ and $k$ on the \rk dataset.}
\renewcommand{\arraystretch}{1}
\begin{tabular*}{\textwidth}{@{\extracolsep{\fill}} lcccc}
\toprule
\textbf{Methods} & $\cent\downarrow$ & $\ent\downarrow$ & $\dnn(\%)\downarrow$ & $\falseneg \downarrow$ \\
\midrule
DFM ($n_{s}=25, k=4$) & $2.471 \pm 4.415$ & $0.303 \pm 0.202$ & $9.190 \pm 6.223$ & $0.0174$ \\
DFM ($n_{s}=25, k=3$) & $2.774 \pm 7.210$ & $0.412 \pm 0.201$ & $10.344 \pm 7.004$ & $0.0412$ \\
DFM ($n_{s}=25, k=2$) & $4.102 \pm 4.815$ & $0.508 \pm 0.210$ & $12.001 \pm 7.350$ & $0.321$ \\
\midrule
DFM ($n_{s}=15, k=4$) & $3.718 \pm 4.950$ & $0.821 \pm 0.202$ & $11.210 \pm 6.702$ & $0.0645$ \\
DFM ($n_{s}=10, k=4$) & $3.210 \pm 3.550$ & $0.965 \pm 0.050$ & $13.105 \pm 7.219$ & $0.184$ \\
\bottomrule
\end{tabular*}
\label{table:DFM_3D_ablation_Re10k}
\end{table*}

\begin{table*}[!htbp]
\centering
\caption{3D Ablation study of DFM with different configurations of $n_s$ and $k$ on the \nyu dataset.}
\renewcommand{\arraystretch}{1}
\begin{tabular*}{\textwidth}{@{\extracolsep{\fill}} lcccc}
\toprule
\textbf{Methods} & $\cent\downarrow$ & $\ent\downarrow$ & $\dnn(\%)\downarrow$ & $\falseneg \downarrow$ \\
\midrule
DFM ($n_{s}=25, k=4$) & $2.062 \pm 3.915$ & $0.541 \pm 0.374$ & $3.948 \pm 5.801$ & $0.000$ \\
DFM ($n_{s}=25, k=3$) & $3.876 \pm 6.235$ & $0.768 \pm 0.426$ & $6.572 \pm 6.245$ & $0.138$ \\
DFM ($n_{s}=25, k=2$) & $6.125 \pm 7.325$ & $0.983 \pm 0.492$ & $9.815 \pm 7.412$ & $0.453$ \\
\midrule
DFM ($n_{s}=15, k=4$) & $4.521 \pm 5.813$ & $0.912 \pm 0.628$ & $8.210 \pm 6.925$ & $0.084$ \\
DFM ($n_{s}=10, k=4$) & $6.832 \pm 3.750$ & $0.965 \pm 0.702$ & $11.582 \pm 7.625$ & $0.288$ \\
\bottomrule
\end{tabular*}
\label{table:DFM_3D_ablation_NYU}
\end{table*}

\begin{table*}[!htbp]
\centering
\caption{Comparative analysis of metrics across different \textit{SDXL 2D analysis} for the \rk dataset.}
\renewcommand{\arraystretch}{1}
\begin{tabular*}{\textwidth}{@{\extracolsep{\fill}} lccccc}
\toprule
\textbf{Methods} & $\cent\downarrow$ & $\ent\downarrow$ & $\dnn(\%)\downarrow$ & $\falseneg \downarrow$ & $\rwa\uparrow$ \\
\midrule
SDXL w/ object cues & $1.257 \pm 2.033$ & $0.848 \pm 0.150$ & $0.446 \pm 25.490$ & $0.039$ & $0.918 \pm 0.147$ \\
SDXL w/o object cues & $1.888 \pm 1.904$ & $0.882 \pm 0.115$ & $6.274 \pm 14.890$ & $0.554$ & $0.688 \pm 0.272$ \\
SDXL w/o prompts & $2.583 \pm 4.612$ & $0.887 \pm 0.106$ & $6.901 \pm 12.663$ & $0.589$ & $0.773 \pm 0.276$ \\
\bottomrule
\end{tabular*}
\label{table:SDXL_ablations_2D_Re10k}
\end{table*}

\begin{table*}[!htbp]
\centering
\caption{Comparative analysis of metrics across different \textit{SDXL 2D analysis} for the \nyu dataset.}
\renewcommand{\arraystretch}{1}
\begin{tabular*}{\textwidth}{@{\extracolsep{\fill}} lccccc}
\toprule
\textbf{Methods} & $\cent\downarrow$ & $\ent\downarrow$ & $\dnn(\%)\downarrow$ & $\falseneg \downarrow$ & $\rwa\uparrow$ \\
\midrule
SDXL w/ object cues & $1.223 \pm 2.107$ & $0.778 \pm 0.212$ & $0.510 \pm 26.929$ & $0.054$ & $0.944 \pm 0.142$ \\
SDXL w/o object cues & $2.601 \pm 2.906$ & $0.891 \pm 0.011$ & $5.649 \pm 15.050$ & $0.378$ & $0.761 \pm 0.284$ \\
SDXL w/o prompts & $3.555 \pm 2.374$ & $0.988 \pm 0.001$ & $5.798 \pm 15.107$ & $0.374$ & $0.757 \pm 0.285$ \\
\bottomrule
\end{tabular*}
\label{table:SDXL_ablations_2D_NYU}
\end{table*}

\newpage
\begin{table*}[!htbp]
\centering
\caption{Comparative analysis of metrics across different SDXL \textit{2.5D analysis} for the \rk dataset.}
\renewcommand{\arraystretch}{1}
\begin{tabular*}{\textwidth}{@{\extracolsep{\fill}} lcccc}
\toprule
\textbf{Methods} & $\cent\downarrow$ & $\ent\downarrow$ & $\dnn(\%)\downarrow$ & $\falseneg \downarrow$ \\
\midrule
SDXL w/ object cues & $1.752 \pm 2.043$ & $0.617 \pm 0.305$ & $6.277 \pm 2.229$ & $0.000$ \\
SDXL w/o object cues & $1.833 \pm 2.011$ & $0.886 \pm 0.104$ & $18.144 \pm 6.148$ & $0.554$ \\
SDXL w/o prompts & $1.874 \pm 1.988$ & $0.931 \pm 0.016$ & $18.533 \pm 9.178$ & $0.589$ \\
\bottomrule
\end{tabular*}
\label{table:SDXL_Re10k_2.5D}
\end{table*}

\begin{table*}[!htbp]
\centering
\caption{Comparative analysis of metrics across different SDXL \textit{2.5D analysis} for the \nyu dataset.}
\renewcommand{\arraystretch}{1}
\begin{tabular*}{\textwidth}{@{\extracolsep{\fill}} lcccc}
\toprule
\textbf{Methods} & $\cent\downarrow$ & $\ent\downarrow$ & $\dnn(\%)\downarrow$ & $\falseneg \downarrow$ \\
\midrule
SDXL w/ object cues & $1.533 \pm 2.916$ & $0.655 \pm 0.323$ & $5.245 \pm 2.351$ & $0.000$ \\
SDXL w/o object cues & $2.612 \pm 1.398$ & $0.818 \pm 0.102$ & $12.637 \pm 9.184$ & $0.378$ \\
SDXL w/o prompts & $2.807 \pm 1.616$ & $0.987 \pm 0.010$ & $16.998 \pm 7.356$ & $0.374$ \\
\bottomrule
\end{tabular*}
\label{table:SDXL_NYU_2.5D}
\end{table*}

\end{document}